\documentclass[10pt,twocolumn,letterpaper]{article}

\usepackage[pagenumbers]{cvpr}

\usepackage{multicol}
\usepackage{multirow}

\usepackage{pifont}
\newcommand{\cmark}{\ding{51}}
\newcommand{\xmark}{\ding{55}}
\usepackage{makecell}
\usepackage{fontawesome}
\usepackage{titletoc}

\definecolor{fly_green}{RGB}{0, 180, 139}
\definecolor{fly_red}{RGB}{239, 99, 75}
\definecolor{fly_blue}{RGB}{99, 113, 250}

\definecolor{background}{RGB}{0, 150, 255}
\definecolor{building}{RGB}{118, 118, 118}
\definecolor{fence}{RGB}{214, 220, 229}
\definecolor{person}{RGB}{4, 50, 255}
\definecolor{pole}{RGB}{190, 153, 153}
\definecolor{road}{RGB}{155, 55, 255}
\definecolor{sidewalk}{RGB}{102, 102, 156}
\definecolor{vegetation}{RGB}{0, 176, 80}
\definecolor{car}{RGB}{250, 188, 1}
\definecolor{wall}{RGB}{152, 251, 152}
\definecolor{traffic-sign}{RGB}{255, 0, 0}

\definecolor{traffic-light}{RGB}{115, 254, 255}
\definecolor{terrain}{RGB}{152, 82, 0}
\definecolor{rider}{RGB}{255, 47, 147}
\definecolor{truck}{RGB}{0, 145, 147}
\definecolor{bus}{RGB}{147, 144, 0}
\definecolor{train}{RGB}{255, 147, 0}
\definecolor{motorcycle}{RGB}{212, 252, 122}
\definecolor{bicycle}{RGB}{255, 64, 255}

\definecolor{cvprblue}{rgb}{0.21,0.49,0.74}

\usepackage[pagebackref,breaklinks,colorlinks,allcolors=cvprblue]{hyperref}

\def\confName{CVPR}
\def\confYear{2025}

\title{\textit{EventFly}: Event Camera Perception from Ground to the Sky}

\author{
    Lingdong Kong$^{1,2}$, Dongyue Lu$^1$, Xiang Xu$^3$, Lai Xing Ng$^4$, Wei Tsang Ooi$^{1,5}$, Benoit R. Cottereau$^{5,6}$
    \\[0.65ex]
    {\small$^1$National University of Singapore~~ $^2$CNRS@CREATE~~ $^3$Nanjing University of Aeronautics and Astronautics}
    \\
    {\small$^4$Institute for Infocomm Research, A*STAR~~ $^5$IPAL, CNRS IRL 2955, Singapore~~ $^6$CerCo, CNRS UMR 5549, Université Toulouse III}
    \\[0.65ex]
    \faGithub~Project Page: \url{https://event-fly.github.io}
}

\begin{document}

\twocolumn[{
    \renewcommand\twocolumn[1][]{#1}
    \maketitle
    \begin{center}
    \centering
    \captionsetup{type=figure}
    \vspace{-0.6cm}
    \includegraphics[width=\textwidth]{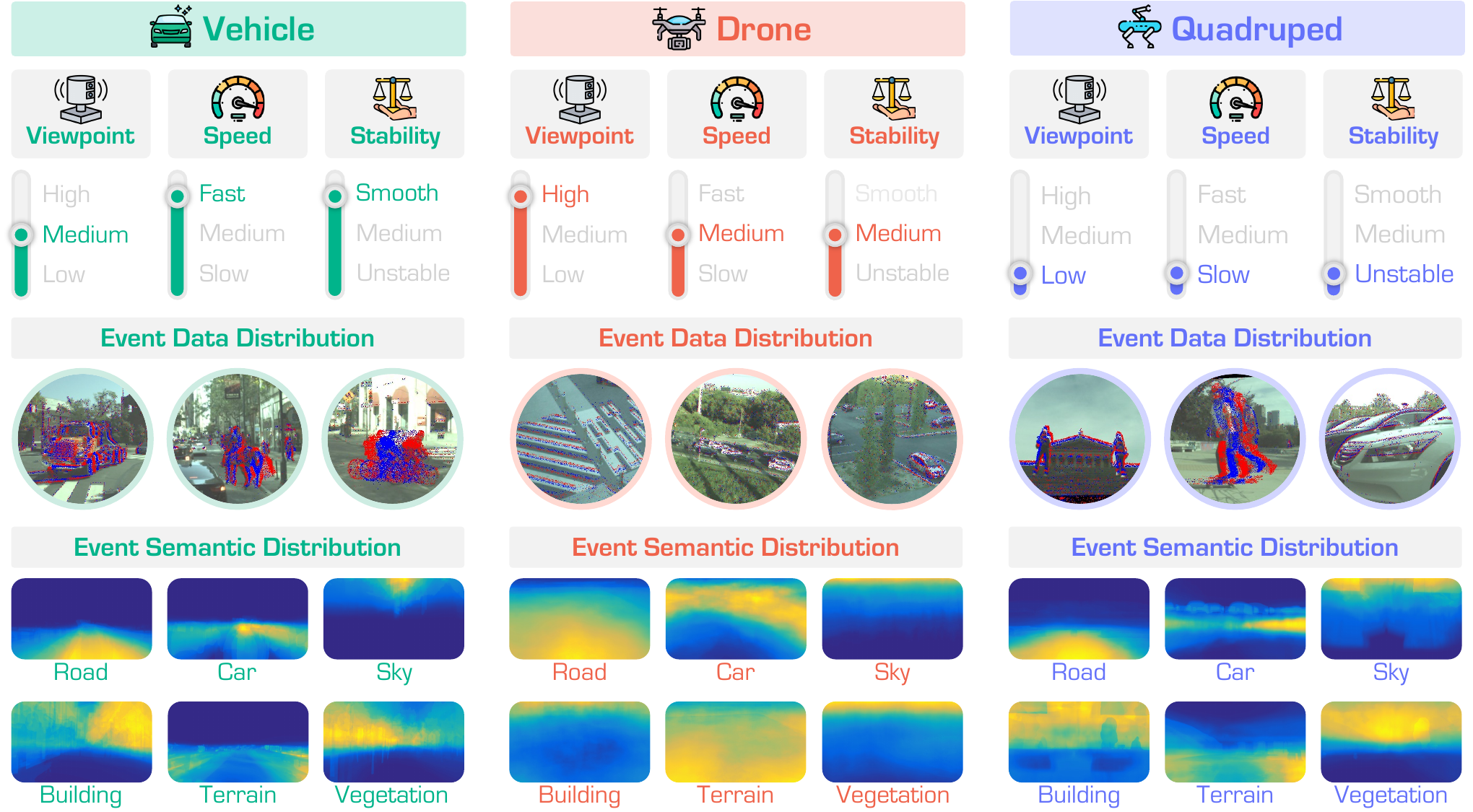}
    \vspace{-0.66cm}
    \caption{Key discrepancies across the \includegraphics[width=0.018\linewidth]{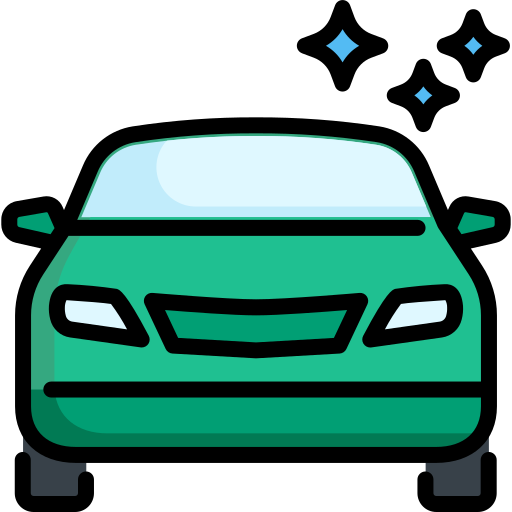}~vehicle ($\mathcal{P}^{\textcolor{fly_green}{\mathbf{v}}}$), \includegraphics[width=0.021\linewidth]{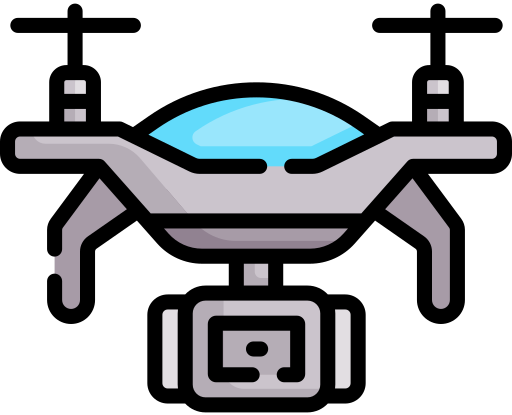}~drone ($\mathcal{P}^{\textcolor{fly_red}{\mathbf{d}}}$), and \includegraphics[width=0.019\linewidth]{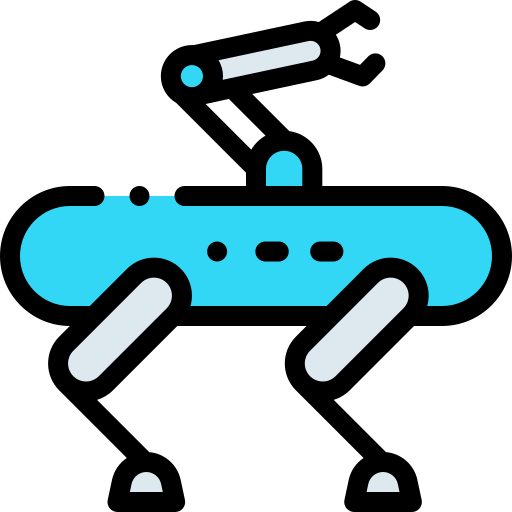}~quadruped ($\mathcal{P}^{\textcolor{fly_blue}{\mathbf{q}}}$) platforms. We compare the core \textbf{attributes} (viewpoint, speed, stability), \textbf{event data distributions}, and \textbf{semantic distributions} across event data acquired by three platforms, highlighting the challenges of adapting event camera perception to diverse operational contexts. These variations motivate the need for a robust \textbf{cross-platform adaptation} framework to harmonize event-based dense perception across distinct environmental setups and conditions.}
    \label{fig:teaser}
    \end{center}
}]

\begin{abstract}
    Cross-platform adaptation in event-based dense perception is crucial for deploying event cameras across diverse settings, such as vehicles, drones, and quadrupeds, each with unique motion dynamics, viewpoints, and class distributions. In this work, we introduce \textbf{\textcolor{fly_green}{Event}\textcolor{fly_red}{Fly}}, a framework for robust cross-platform adaptation in event camera perception. Our approach comprises \textbf{three} key components: \textbf{i)} Event Activation Prior (EAP), which identifies high-activation regions in the target domain to minimize prediction entropy, fostering confident, domain-adaptive predictions; \textbf{ii)} EventBlend, a data-mixing strategy that integrates source and target event voxel grids based on EAP-driven similarity and density maps, enhancing feature alignment; and \textbf{iii)} EventMatch, a dual-discriminator technique that aligns features from source, target, and blended domains for better domain-invariant learning. To holistically assess cross-platform adaptation abilities, we introduce \textbf{\textcolor{fly_green}{E}\textcolor{fly_red}{X}\textcolor{fly_blue}{Po}}, a large-scale benchmark with diverse samples across vehicle, drone, and quadruped platforms. Extensive experiments validate our effectiveness, demonstrating substantial gains over popular adaptation methods. We hope this work can pave the way for more adaptive, high-performing event perception across diverse and complex environments.
\vspace{-0.5cm}
\end{abstract}    
\section{Introduction}
\label{sec:intro}

Event cameras, with their asynchronous operation 
and high temporal resolution, provide great advantages over conventional frame-based sensors by capturing precise pixel intensity changes as they occur \cite{gallego2022survey,shiba2024secrets,gehrig2024dagr,kim2021n-imagenet,kim2024motion}. These attributes make event cameras ideally suited for real-time applications in dynamic environments, such as autonomous driving, aerial navigation, and robotic perception \cite{steffen2019neuromorphic,maqueda2018event}.

Despite this potential, event cameras remain primarily deployed on the vehicle platform \cite{carmichael2024datasets,brodermann2024muses}. Based on a recent survey study \cite{chakravarthi2024survey}, $12$ out of $15$ public event camera perception datasets are collected exclusively from ground vehicles \cite{binas2017ddd17,gehrig2021dsec,perot2020mpx1,zhu2018mvsec,zhao2024subt-mrs,chen2024ecmd}. In contrast, datasets involving alternative robot platforms, such as drones \cite{blosch2010vision,cladera2024evmapper,wang2024eventvot} or mobile robots \cite{zhou2024exact,zhu2024cear,boretti2023pedro}, are limited, making robust event-based perception across diverse domains challenging \cite{shariff2024survey,rancon2022stereospike}. To enable versatile applications, a \textbf{cross-platform approach} is vital for adapting the perception models to the new, real-world, and challenging operational contexts.

Cross-platform adaptation for event-based dense perception presents significant technical challenges. As can be seen from \cref{fig:teaser}, each robot platform -- vehicle, drone, or quadruped -- exhibits unique motion patterns, environmental interactions, and perspective-based dynamics that create distinct \textbf{activation patterns} in the event data \cite{posch2010gen1,son2017gen3,finateu2020gen4,loquercio2021learning}. For example, ground vehicles capture event activity near surfaces \cite{gehrig2024dagr,zhu2018mvsec,gehrig2021dsec}, while drones tend to capture high-altitude landscape features with sparse ground details \cite{chaney2023m3ed,maqueda2018event,loquercio2018dronet,vidal2018ultimate,kaufmann2023champion}. Conventional domain adaptation methods commonly used for frame-based sensors are not well-suited to handle the distinct spatial-temporal nuances of event camera data, which demand tailored alignment strategies \cite{kouw2021survey,oza2024survey,li2024survey,triess2021survey}. A specialized framework for cross-platform adaptation in event-based dense perception is, therefore, crucial for \textbf{reliable, context-sensitive perception} across varied robot sensing conditions.

In this work, we introduce \textit{\textbf{\textcolor{fly_green}{Event}\textcolor{fly_red}{Fly}}}, a robust framework for cross-platform adaptation in event camera dense perception. We bridge domain-specific gaps in event-camera perception by leveraging platform-specific activation patterns and aligning feature distributions across domains. Our design consists of \textbf{three core components}, each addressing a specific challenge in cross-platform event data alignment:

\noindent\textbf{\textcolor{fly_green}{{\footnotesize$\blacksquare$}} Event Activation Prior (EAP).} Event cameras on different platforms reveal distinctive high-activation regions shaped by platform-specific dynamics. EAP leverages these by identifying frequently activated areas in the target domain and minimizing prediction entropy within these regions. This promotes confident, domain-adaptive predictions that align the model to platform-specific event patterns, enhancing performance in target-relevant contexts.

\noindent\textbf{\textcolor{fly_red}{{\footnotesize$\blacksquare$}} Cross-Platform Consistency Regularization.} To bridge the gaps in-between different robot platforms, we propose EventBlend to create hybrid event voxel grids by combining source and target event data in a spatially structured way. Guided by EAP-driven similarity and density maps, this method selectively integrates features based on shared activation patterns, focusing on areas where platform-specific details overlap. This fusion retains platform-specific cues while enhancing cross-domain feature consistency, balancing reliable source information with context-sensitive target features for more robust cross-platform adaptations.

\noindent\textbf{\textcolor{fly_blue}{{\footnotesize$\blacksquare$}} Cross-Platform Adversarial Training.} To further align domain distributions, we propose EventMatch, a dual-discriminator approach across source, target, and blended representations. One discriminator enforces alignment between source and blended domains, while the other softly adapts blended features toward the target, particularly in high-activation regions. This layered approach supports robust, domain-adaptive learning that generalizes well across platforms, balancing domain-invariant features with target-specific adaptations for improved perception accuracy.

To rigorously assess the cross-platform adaptation performance, we establish \textbf{\textit{\textcolor{fly_green}{E}\textcolor{fly_red}{X}\textcolor{fly_blue}{Po}}}, a large-scale \textbf{\textit{\textcolor{fly_green}{E}}}vent-based \textbf{\textcolor{fly_red}{\textit{X}}}ross-\textbf{\textcolor{fly_blue}{\textit{P}}}latf\textbf{\textcolor{fly_blue}{\textit{o}}}rm dense perception benchmark with around $90$k event data from three platforms: \includegraphics[width=0.04\linewidth]{figures/icons/vehicle.png}~\textit{vehicle}, \includegraphics[width=0.045\linewidth]{figures/icons/drone.png}~\textit{drone}, and \includegraphics[width=0.04\linewidth]{figures/icons/quadruped.png}~\textit{quadruped}. From extensive benchmark experiments, our approach consistently demonstrates robust adaptation capabilities, achieving on average $\mathbf{23.8\%}$ \textbf{higher accuracy} and $\mathbf{77.1\%}$ \textbf{better mIoU} across platforms compared to \textit{source-only} training. When evaluated against prior adaptation methods, we outperform by significant margins across almost all semantic classes, highlighting the scalability and effectiveness in diverse, challenging operational contexts.

In summary, our main contributions are as follows:
\begin{itemize}
    \item We propose \textit{\textbf{\textcolor{fly_green}{Event}\textcolor{fly_red}{Fly}}}, a novel framework designed for cross-platform adaptation in event camera perception, facilitating robustness under diverse event data dynamics. To our knowledge, this is the \textit{first} work proposed to address this critical gap in event-based perception tasks.

    \item We introduce Event Activation Prior (EAP), EventBlend, and EventMatch -- a set of tailored techniques that utilize platform-specific activation patterns, spatial data mixing, and dual-domain feature alignment to tackle the unique challenges of event-based cross-platform adaptation.

    \item We establish a large-scale benchmark, \textbf{\textit{\textcolor{fly_green}{E}\textcolor{fly_red}{X}\textcolor{fly_blue}{Po}}}, for cross-platform adaptation in event-based perception, comprising rich collections of samples from vehicle, drone, and quadruped domains, providing a strong foundation for further research in adaptive event camera applications.
\end{itemize}
\section{Related Work}
\label{sec:related_work}

\noindent\textbf{Event Camera Perception.} 
Recent works harnessed event cameras for diverse perception tasks in dynamic settings, such as detection \cite{gehrig2022pushing, gehrig2023rvt, zubic2023from, gehrig2024dagr, lu2024flexevent}, segmentation \cite{alonso2019ev-segnet, hamaguchi2023hmnet, sun2022ess, kong2024openess}, depth estimation \cite{hidalgo2020learning, mostafavi2021event, cho2023learning, zhu2018ev-flownet, zhu2019unsupervised, gehrig2021e-raft}, and visual odometry \cite{censi2014odometry, mueggler2018continuous, cuadrado2023optical, hidalgo2022event}. Object detection with event cameras involves locating objects by leveraging the high temporal resolution of event streams, enabling effective scene parsing even under challenging conditions \cite{zhou2023rgb, peng2024sast, li2023sodformer, zou2023seeing}. Meanwhile, semantic segmentation aims to enhance scene understanding, which is vital for safe navigation in autonomous systems \cite{messikommer2022ev-transfer, wang2021dtl, cho2023learning, klenk2024mem}. Additionally, recent research has focused on integrating event data with complementary modalities to improve accuracy and mitigate the sparse nature of event data \cite{gehrig2020vid2e, wang2021evdistill, biswas2022halsie, chen2024segment, hareb2024evsegsnn}. Our work extends this line by focusing on robust cross-platform adaptation, ensuring that event-based perception models can generalize across different robotic platforms.

\noindent\textbf{Cross-Domain Adaptation.}
Transferring knowledge from a labeled source domain to an unlabeled target domain addresses the challenge of limited annotated data in diverse real-world cases \cite{zhou2023survey, schwonberg2023survey}. Traditional methods in this area leverage strategies like domain adversarial training \cite{tsai2018adaptsegnet}, entropy minimization \cite{vu2019advent, vu2019dada, pan2020intra, saha2021ctrl}, contrastive learning \cite{zhang2021proda, xie2023sepico, hoyer2022hrda}, domain mixing \cite{liu2021bapa-net, zhou2023camix, kong2023conDA, tranheden2021dacs}, and self-training \cite{zou2018cbst, hoyer2023mic, bruggemann2023refign, zhao2024plsr}. However, despite promising performances, these approaches are primarily designed for frame-based data \cite{cordts2016cityscapes,zhang2023cmx,hoyer2022daformer}, which lacks the unique spatiotemporal properties of the stream-based event camera data. Our approach adapts these domain adaptation principles specifically for event-based vision perception, addressing the distinct challenges posed by event camera data.

\noindent\textbf{Domain Adaptation from Event Camera.}
Recent works have begun to address domain shifts for event-based perception. Approaches like Ev-Transfer \cite{messikommer2022ev-transfer}, ESS \cite{sun2022ess}, and HPL-ESS \cite{jing2024hpl-ess} focus on transferring knowledge from RGB frames to event data, enabling effective event-based perception through frame-to-event domain adaptation. Other studies leverage event cameras to aid adaptation in low-light or night-time conditions for conventional RGB sensors \cite{xia2023cmda, jeong2024towards, ercan2024hue, yao2024event}. There are also works that explored efficient learning utilizing shared representations across modalities \cite{ercan2023evreal, yang2023ecdp, kamal2024efficient}. To the best of our knowledge, our work is the first to tackle cross-platform adaptation across multiple event camera platforms, such as vehicles, drones, and quadrupeds, focusing on the unique motion and environmental interactions inherent to each platform.
\section{Methodology}
\label{sec:method}
This work tackles the problem of cross-platform adaptation for event-based dense perception across three distinct platforms: \includegraphics[width=0.04\linewidth]{figures/icons/vehicle.png}~\textit{vehicle}, \includegraphics[width=0.045\linewidth]{figures/icons/drone.png}~\textit{drone}, and \includegraphics[width=0.04\linewidth]{figures/icons/quadruped.png}~\textit{quadruped}, denoted as $\mathcal{P}^{\textcolor{fly_green}{\mathbf{v}}}$, $\mathcal{P}^{\textcolor{fly_red}{\mathbf{d}}}$, and $\mathcal{P}^{\textcolor{fly_blue}{\mathbf{q}}}$, respectively. Each platform exhibits unique spatial and temporal characteristics in the captured event data, influenced by platform-specific perspectives, motion patterns, semantics, and environments.

\begin{figure*}
    \centering
    \includegraphics[width=0.96\linewidth]{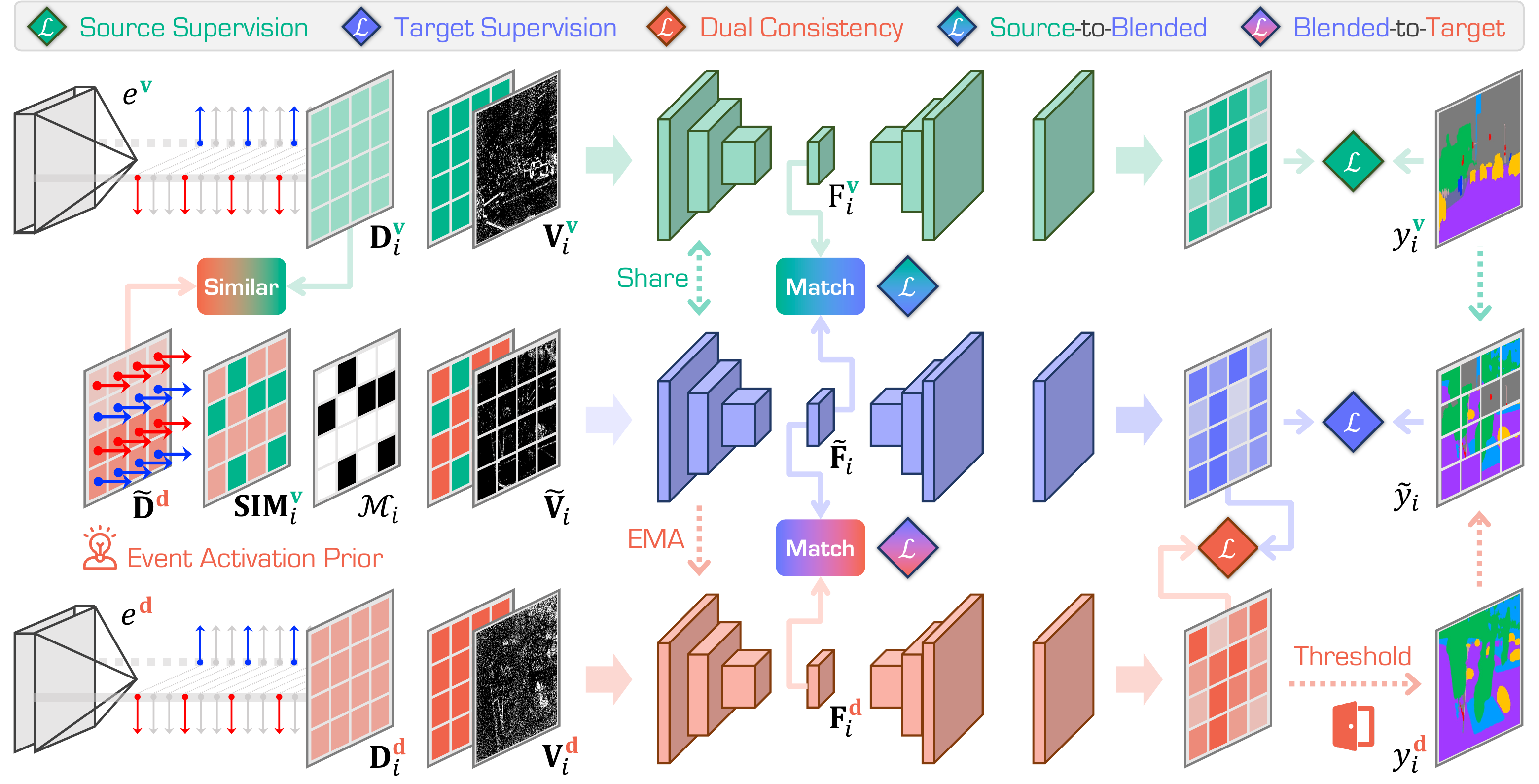}
    \vspace{-0.3cm}
    \caption{Overview of the \textit{\textbf{\textcolor{fly_green}{Event}\textcolor{fly_red}{Fly}}} framework. Guided by the EAP principle (\cref{sec:prior}), the pair of source and target event data $\mathbf{V}^{\textcolor{fly_green}{\mathbf{v}}}_{i}$ and $\mathbf{V}^{\textcolor{fly_red}{\mathbf{d}}}_{i}$ are mixed via the EventBlend operation (\cref{sec:eventblend}), where the blending mask $\mathcal{M}_i$ is obtained by measuring the similarities between density maps $\mathbf{D}^{\textcolor{fly_green}{\mathbf{v}}}_{i}$ and $\Tilde{\mathbf{D}}^{\textcolor{fly_red}{\mathbf{d}}}$. The features $\mathbf{F}^{\textcolor{fly_green}{\mathbf{v}}}$, $\mathbf{F}^{\textcolor{fly_red}{\mathbf{d}}}$, and $\Tilde{\mathbf{F}}$ from the source, target, and blended domains are then used for EventMatch (\cref{sec:eventmatch}). This facilitates learning an intermediary representation that is both robust (source-aligned) and adaptable (target-sensitive).}
    \label{fig:framework}
    \vspace{-0.2cm}
\end{figure*}

\subsection{Preliminaries}
\label{sec:preliminary}

Event cameras capture continuous, asynchronous streams \cite{brandli2014davis}, where each event $\mathbf{e}_i = (e^{x}_{i}, e^{y}_{i}, e^{t}_{i}, e^{p}_{i})$ consists of pixel coordinates $(e^{x}_{i}, e^{y}_{i})$, timestamp $e^{t}_{i}$, and polarity $e^{p}_{i} \in \{-1, +1\}$, indicating either a decrease or increase in brightness. This format allows event cameras to capture high-speed motion and scene changes in dynamic environments.

\noindent\textbf{Event Data Representation.}
To efficiently process event data, we convert raw events into voxel grid representations, denoted as $\mathbf{V}$. This structured format regularizes the event data, making it compatible with learning-based methods and consistent across different domains. To construct each voxel grid $\mathbf{V}_i\in \mathbb{R}^{T \times H \times W}$, where $T$ is the number of temporal bins, and $H$ and $W$ are the spatial dimensions of the event sensor, we divide the event stream into fixed time intervals, with positive and negative events accumulated in separate bins along the temporal dimension. This polarity-sensitive encoding captures spatial and temporal dynamics of events that reflect both motion directionality and density:
\begin{equation}
   \sum_{\mathbf{e}_j \in \varepsilon_i} e^{p}_{j} \delta(e^{x}_{j} - e^{x}) \delta(e^{y}_{j} - e^{y}) \max \{1 - |\hat{e}^{t}_j - e^{t}|, 0\}~,
   \vspace{-0.2cm}
\end{equation}
where $\delta$ is the Kronecker delta function; $\hat{e}^{t}_j = (C-1)\frac{e^{t}_{j} - e^{t}_{0}}{\Delta T}$ normalizes the event timestamps. Here, $\Delta T$ is the duration of each time window, and $e^{t}_{0}$ is the initial timestamp.

\noindent\textbf{Cross-Platform Adaptation.}
We aim to address the challenge of adapting across three distinct domains, represented by $\mathcal{D} = \{\mathcal{P}^{\textcolor{fly_green}{\mathbf{v}}}, \mathcal{P}^{\textcolor{fly_red}{\mathbf{d}}}, \mathcal{P}^{\textcolor{fly_blue}{\mathbf{q}}}\}$, respectively. Each domain consists of event voxel grids $\mathbf{V}$ and associated labels $y \in \mathbb{R}^{H \times W}$, where each pixel label belongs to one of $C$ semantic classes.
In our cross-platform adaptation setting, we assume access to fully labeled data from the source platform while the target platform has only unlabeled data. The goal is to leverage the labeled source data and unlabeled target data to improve event-based perception performance on the target domain.\footnote{Without loss of generality, our subsequent explanations focus on the adaptation from \textit{vehicle} domain $\mathcal{P}^{\textcolor{fly_green}{\mathbf{v}}}$ to \textit{drone} domain $\mathcal{P}^{\textcolor{fly_red}{\mathbf{d}}}$, although the proposed framework generalizes to any pair of domains in $\mathcal{D}$.}

\subsection{Event Activation Prior}
\label{sec:prior}
Cross-platform adaptation in event-based perception hinges on aligning platform-specific activation patterns, which are influenced by domain-unique motions, and environments. We introduce \textbf{Event Activation Prior (EAP)}, a regularization technique that encourages confident, low-entropy predictions in target-domain regions with high event activation.

In each platform, event data often exhibits recurrent activation patterns in specific regions, driven by platform-specific motion and scene dynamics (see~\cref{fig:teaser}). For instance, lower regions in a \textit{vehicle} domain capture details like road surfaces and obstacles, while upper regions in a \textit{drone} domain capture landscape features and environmental context. Such regions, denoted as $\mathbf{S} \subset \{0, 1, \dots, H-1\} \times \{0, 1, \dots, W-1\}$ and along with the conditional entropy  $H(y_{\mathbf{S}} | \mathbf{V}_{\mathbf{S}}, \mathbf{S})$, typically have consistent semantic patterns, allowing for more confident predictions. By leveraging these high-activation areas, we aim to minimize the entropy of predictions in such regions, thereby guiding the model to produce more confident outputs aligned with the unique characteristics of the target platform domain.

\noindent\textbf{Formulating EAP.}
To incorporate EAP in training, we define it as a regularization that minimizes prediction uncertainty in these high-activation event areas. For each subregion $\mathbf{S}$, EAP aims to  minimize the conditional entropy $H(y_{\mathbf{S}} | \mathbf{V}_{\mathbf{S}}, \mathbf{S})$, focusing on domain-specific high-activation regions. We enforce the expectation constraint over the model parameters $\theta$ as $\mathbb{E}_{\theta}[ H( \mathbf{V}_{\mathbf{S}}, y_{\mathbf{S}} | \mathbf{S})] \leq c$, where $c$ is a small constant that encourages high-confidence predictions. This is transformed into a \textbf{prior} on parameters $\theta$ through the maximum entropy principles \cite{grandvalet2004entropy-minimization}, yielding:
\begin{equation}
    P(\theta) \propto \exp\left( -\lambda H( \mathbf{V}_{\mathbf{S}}, y_{\mathbf{S}} | \mathbf{S}) \right) \propto \exp\left( -\lambda H(y_{\mathbf{S}} | \mathbf{V}_{\mathbf{S}} , \mathbf{S}) \right),
    \label{eq:eap}
\end{equation}
where $\lambda > 0$ is the Lagrange multiplier corresponding to constant $c$, which balances the effect of EAP on training.

\noindent\textbf{Empirical Estimation of EAP.}
To apply EAP, we empirically estimate the entropy in high-activation event regions by aggregating over activations in the target platform domain, where activation patterns are highly informative. Using an empirical estimator, we approximate $H(y | \mathbf{V}, \mathbf{S})$ as:
\begin{equation}
\vspace{-0.1cm}
    H_{\mathrm{emp}}(y | \mathbf{V}, \mathbf{S}) = \mathbb{E}_{\mathbf{V}, y, \mathbf{S}}\left[\hat{P}(y | \mathbf{V}, \mathbf{S}) \log \hat{P}(y | \mathbf{V}, \mathbf{S})\right],
\end{equation}
where $\hat{P}(y | \mathbf{V}, \mathbf{S})$ is the empirical prediction probability conditioned on the voxel grid $\mathbf{V}$ and restricted to region $\mathbf{S}$. By minimizing $H_{\mathrm{emp}}(y | \mathbf{V}, \mathbf{S})$, we encourage confident predictions within these regions, aligning the model’s predictions with the target domain activation patterns.

\noindent\textbf{Integrating EAP into Training.}
To incorporate the EAP in \cref{eq:eap} into cross-platform adaptation, we define the overall objective as a maximum-a-posteriori (MAP) estimation:
\begin{equation}
    C(\theta) = \mathcal{L}(\theta) - \lambda H_{\mathrm{emp}}(y | \mathbf{V}, \mathbf{S})~,
    \label{eq:max-a-posteriori}
\end{equation}
where $\mathcal{L}(\theta)$ represents the supervised loss on source data. $H_{\mathrm{emp}}(y | \mathbf{V}, \mathbf{S})$ minimizes uncertainty in the target domain by leveraging EAP over high-activation regions. By focusing on these regions, the event camera perception model tends to effectively learn to adapt its predictions to align with platform-specific activation cues in the target domain, achieving robust adaptation across varied platforms.

\subsection{EventBlend}
\label{sec:eventblend}
Cross-platform adaptation in event-based perception requires a selective integration of data from both source and target domains. Here, we introduce \textbf{EventBlend}, a data-mixing strategy designed to construct hybrid voxel grids by blending event data from the source and target domains, guided by high-activation regions identified through EAP. As shown in \cref{fig:framework}, by constructing these blended voxel grids, we aim to enhance the generalization ability across domain shifts, enabling the model to adapt effectively to the target distribution while leveraging reliable source labels.

\noindent\textbf{Event Density Maps.}
To identify regions where the source and target domains exhibit similar or divergent activation patterns, we employ density maps, which highlight areas of frequent event activity. For each source event data with voxel grid $\mathbf{V}^{\textcolor{fly_green}{\mathbf{v}}}_{i} \in \mathbb{R}^{T \times H \times W}$, we calculate a corresponding source density map, \(\mathbf{D}^{\textcolor{fly_green}{\mathbf{v}}}_{i} \in \mathbb{R}^{H \times W}\), by summing activations along the temporal axis $T$ for each spatial location $(\mu, \nu)$:
\begin{equation}
\vspace{-0.1cm}
    \mathbf{D}^{\textcolor{fly_green}{\mathbf{v}}}_{i}(\mu, \nu) = \sum\nolimits_{t=1}^{T} |\mathbf{V}^{\textcolor{fly_green}{\mathbf{v}}}_{i}(t, \mu, \nu)|~.
    \label{eq:density_source}
\end{equation}
For the target domain, we pre-compute an aggregated density map, $\Tilde{\mathbf{D}}^{\textcolor{fly_red}{\mathbf{d}}} \in \mathbb{R}^{H \times W}$, by summing activations over all target samples \(\mathbf{V}^{\textcolor{fly_red}{\mathbf{d}}}_{j}\), where $j$ indexes each target sample:
\begin{equation}
\vspace{-0.1cm}
    \Tilde{\mathbf{D}}^{\textcolor{fly_red}{\mathbf{d}}}(\mu, \nu) = \frac{1}{N^{\textcolor{fly_red}{\mathbf{d}}}} \sum\nolimits_{j} \sum\nolimits_{t=1}^{T} \left| \mathbf{V}^{\textcolor{fly_red}{\mathbf{d}}}_{j}(t, \mu, \nu) \right|~,
    \label{eq:density_target}
\end{equation}
where $N^{\textcolor{fly_red}{\mathbf{d}}}$ denotes the total number of samples in the target domain.
These density maps highlight regions with significant activity, which aids in identifying areas where the domains may share similar activation patterns, consistent with the high-activation regions established by EAP.

\noindent\textbf{Similarity Map \& Binary Mask.}
To guide selective blending, for each source sample, we construct a similarity map $\mathbf{SIM}_i \in \mathbb{R}^{H \times W}$ to quantify spatial overlap in activation patterns between two domains. This map is defined as:
\begin{equation}
\vspace{-0.1cm}
    \mathbf{SIM}_i(\mu, \nu) = 1 - | \mathbf{D}^{\textcolor{fly_green}{\mathbf{v}}}_{i}(\mu, \nu) - \Tilde{\mathbf{D}}^{\textcolor{fly_red}{\mathbf{d}}}(\mu, \nu) |~.
\end{equation}
Here, high values in $\mathbf{SIM}_i$ correspond to regions where source and target domains exhibit similar activation intensities, aligning with the high-activation areas specified by EAP in \cref{eq:eap}. Next, we derive a binary mask $\mathcal{M}_i \in \{0, 1\}^{H \times W}$ by applying a threshold $\tau$ to the similarity map:
\begin{align}
\vspace{-0.2cm}
    \mathcal{M}_i(\mu, \nu) =
    \begin{cases}
    \mathbf{1}, & \mathrm{if}~~ \mathbf{SIM}_i(\mu, \nu) \geq \tau~, \\
    \mathbf{0}, & \mathrm{otherwise}~.
    \end{cases}
\vspace{-0.1cm}
\end{align}
This mask directs the blending process by specifying which regions to retain from the source and which to adapt from the target. A mask value of $\mathbf{1}$ indicates that activations in $(\mu, \nu)$ from the source domain should be retained, while a value of $\mathbf{0}$ specifies a shift to the target domain activations.
It is worth mentioning that for the selections of $\mathbf{D}^{\textcolor{fly_green}{\mathbf{v}}}_{i}$ and $\Tilde{\mathbf{D}}^{\textcolor{fly_red}{\mathbf{d}}}$ in \cref{eq:density_source} and \cref{eq:density_target}, we only consider \textbf{activated} pixel locations. That is to say, the similarity score does not count cases where both source and target are \textbf{non-activated}.

\noindent\textbf{Constructing Blended Event Voxel Grids.}
For a pair of source and target samples $\mathbf{V}^{\textcolor{fly_green}{\mathbf{v}}}_{i}$ and $\mathbf{V}^{\textcolor{fly_red}{\mathbf{d}}}_{i}$, we create a new event voxel grid $\Tilde{\mathbf{V}}_i \in \mathbb{R}^{T \times H \times W}$, using $\mathcal{M}_i$. For each spatial location $(\mu, \nu)$, temporal sequences are selectively copied from either source and target based on $\mathcal{M}_i(\mu, \nu)$:
\begin{align}
\vspace{-0.2cm}
    \Tilde{\mathbf{V}}_{i}(:, \mu, \nu) =
    \begin{cases}
    \mathbf{V}^{\textcolor{fly_green}{\mathbf{v}}}_{i}(:, \mu, \nu), & \mathrm{if }~~ \mathcal{M}(\mu, \nu) = \mathbf{1}~, \\
    \mathbf{V}^{\textcolor{fly_red}{\mathbf{d}}}_{i}(:, \mu, \nu), & \mathrm{if }~~ \mathcal{M}(\mu, \nu) = \mathbf{0}~.
    \end{cases}
\end{align}
This selective copying approach allows $\Tilde{\mathbf{V}}_{i}$ to incorporate source-domain stability (\ie, human annotations) in high-similarity regions while adapting to target-specific patterns in low-similarity areas. Finally, we generate the label map $\Tilde{y}_{i} \in \mathbb{R}^{H \times W}$ for $\Tilde{\mathbf{V}}_{i}$ by combining ground truth labels from the source and pseudo-labels from the target. Here, the pseudo-labels can be obtained offline from a source-pretrained model or generated online using a mean teacher framework \cite{tarvainen2017mean-teacher}.
During adaptation, we use $\Tilde{\mathbf{V}}_{i}$ and $\Tilde{y}_{i}$ for supervised learning. The blended data contains activations and labels from both domains. We leverage EAP-aligned high-activation areas, \ie, \cref{eq:max-a-posteriori}, where source-target consistency is high, facilitating robust cross-domain adaptation.

\subsection{EventMatch}
\label{sec:eventmatch}
In cross-platform adaptation, domain shifts manifest as discrepancies in spatial activation patterns, motion characteristics, and perspectives unique to each platform. The \textbf{EAP} (\cf~\cref{sec:prior}) identifies high-activation areas for stable adaptation, while \textbf{EventBlend} (\cf~\cref{sec:eventblend}) generates hybrid voxel grids that capture shared spatial structures across domains. However, without explicit feature alignment, the model might still struggle to fully adapt to platform-specific idiosyncrasies, especially in target-specific contexts.

To address this, we propose the \textbf{EventMatch} technique to align the blended domain as an intermediary, enabling robust domain-invariant feature learning by maintaining source reliability while selectively adapting to target-specific characteristics. We introduce two fully convolutional discriminators, $\sigma_1$ and $\sigma_2$, each responsible for a distinct aspect of feature alignment:
\begin{itemize}
    \item \textbf{Source \& Blended} (via $\sigma_1$): We align the blended voxel grids $\Tilde{\mathbf{V}}$ with the source domain, encouraging them to retain reliable source-domain characteristics. This alignment preserves robustness and ensures that the blended features maintain patterns learned from the ground truth supervised labels, \ie, $y$, from the source domain.
    \item \textbf{Blended \& Target} (via $\sigma_2$): Rather than the strict alignment, we regularize the blended features toward a soft alignment with the target, selectively adapting target-relevant characteristics. This alignment is focused on high-activation regions identified by EAP, allowing the blended features to bridge the domain gap in target-specific contexts without sacrificing source consistency.
\end{itemize}
This dual-discriminator setup allows the blended features to act as an intermediary representation that is both robust (source-aligned) and adaptable (target-sensitive).

\begin{table*}[t]
    \centering
    \caption{
        \textbf{Benchmark results of platform adaptation} from \includegraphics[width=0.018\linewidth]{figures/icons/vehicle.png}~vehicle ($\mathcal{P}^{\textcolor{fly_green}{\mathbf{v}}}$) to \includegraphics[width=0.021\linewidth]{figures/icons/drone.png}~drone ($\mathcal{P}^{\textcolor{fly_red}{\mathbf{d}}}$). \textcolor{gray}{Target} is trained with ground truth from the target domain. All scores are given in percentage (\%). The \colorbox{fly_red!12}{\textcolor{fly_red}{second best}} and \colorbox{fly_green!12}{\textcolor{fly_green}{best}} scores under each metric are highlighted in colors.}
    \vspace{-0.25cm}
    \resizebox{\linewidth}{!}{
    \begin{tabular}{r|p{26pt}<{\centering}p{26pt}<{\centering}p{26pt}<{\centering}p{26pt}<{\centering}|p{26pt}<{\centering}p{26pt}<{\centering}p{26pt}<{\centering}p{26pt}<{\centering}p{26pt}<{\centering}p{26pt}<{\centering}p{26pt}<{\centering}p{26pt}<{\centering}p{26pt}<{\centering}p{26pt}<{\centering}p{26pt}<{\centering}}
    \toprule
    \textbf{Method} & \textbf{Acc} & \textbf{mAcc} & \textbf{mIoU} & \textbf{fIoU} & \rotatebox{90}{\textcolor{background}{$\blacksquare$}~ground~} & \rotatebox{90}{\textcolor{building}{$\blacksquare$}~build} & \rotatebox{90}{\textcolor{fence}{$\blacksquare$}~fence} & \rotatebox{90}{\textcolor{person}{$\blacksquare$}~person} & \rotatebox{90}{\textcolor{pole}{$\blacksquare$}~pole} & \rotatebox{90}{\textcolor{road}{$\blacksquare$}~road} & \rotatebox{90}{\textcolor{sidewalk}{$\blacksquare$}~walk} & \rotatebox{90}{\textcolor{vegetation}{$\blacksquare$}~veg} & \rotatebox{90}{\textcolor{car}{$\blacksquare$}~car} & \rotatebox{90}{\textcolor{wall}{$\blacksquare$}~wall} & \rotatebox{90}{\textcolor{traffic-sign}{$\blacksquare$}~sign}
    \\\midrule\midrule
    \textcolor{gray}{Source-Only}~\textcolor{gray}{$\circ$} & \textcolor{gray}{$43.69$} & \textcolor{gray}{$33.81$} & \textcolor{gray}{$15.04$} & \textcolor{gray}{$11.81$} & \textcolor{gray}{$48.71$} & \textcolor{gray}{$11.57$} & \textcolor{gray}{$0.92$} & \textcolor{gray}{$8.42$} & \textcolor{gray}{$13.33$} & \textcolor{gray}{$25.48$} & \textcolor{gray}{$8.18$} & \textcolor{gray}{$31.51$} & \textcolor{gray}{$14.88$} & \textcolor{gray}{$0.04$} & \textcolor{gray}{$2.41$}
    \\\midrule
    AdaptSegNet \cite{tsai2018adaptsegnet} & $49.14$ & $35.38$ & $21.16$ & $12.15$ & $29.37$ & $23.57$ & $0.17$ & $0.48$ & $13.45$ & $38.23$ & $17.85$ & $48.73$ & $29.42$ & $35.55$ & $0.40$ 
    \\
    CBST \cite{zou2018cbst} & $57.95$ & $41.18$ & $24.31$ & $16.02$ & $33.05$ & $24.43$ & $0.00$ & $3.08$ & $18.24$ & $56.32$ & $16.84$ & $56.15$ & $23.61$ & $35.65$ & $0.00$
    \\
    IntraDA \cite{pan2020intra} & $57.37$ & $38.85$ & $23.58$ & $15.91$ & $32.31$ & $23.17$ & $0.00$ & $4.90$ & $14.91$ & $56.70$ & $18.67$ & $54.94$ & $20.71$ & $33.08$ & $0.00$
    \\
    DACS \cite{tranheden2021dacs} & $59.81$ & $42.01$ & $27.07$ & $16.14$ & $35.16$ & $26.12$ & \cellcolor{fly_red!12}$0.18$ & $4.11$ & $18.49$ & $55.64$ & \cellcolor{fly_red!12}$21.74$ & $56.81$ & \cellcolor{fly_red!12}$34.69$ & \cellcolor{fly_green!12}$44.73$ & $0.05$
    \\
    MIC \cite{hoyer2023mic} & $63.11$ & $45.60$ & $28.87$ & $17.46$ & $41.40$ & $25.19$ & $0.01$ & \cellcolor{fly_red!12}$10.11$ & $22.86$ & $59.25$ & $20.84$ & $58.86$ & $33.95$ & $44.18$ & $0.90$
    \\
    PLSR \cite{zhao2024plsr} & \cellcolor{fly_red!12}$64.61$ & \cellcolor{fly_red!12}$45.93$ & \cellcolor{fly_red!12}$29.69$ & \cellcolor{fly_red!12}$17.99$ & \cellcolor{fly_red!12}$42.09$ & \cellcolor{fly_red!12}$30.06$ & $0.00$ & $9.75$ & \cellcolor{fly_red!12}$23.32$ & \cellcolor{fly_red!12}$62.48$ & $20.65$ & \cellcolor{fly_red!12}$60.15$ & $31.69$ & $44.27$ & \cellcolor{fly_red!12}$2.06$ 
    \\
    \textbf{\textcolor{fly_green}{Event}\textcolor{fly_red}{Fly}} \textbf{(Ours)} & \cellcolor{fly_green!12}$69.17$ & \cellcolor{fly_green!12}$48.20$ & \cellcolor{fly_green!12}$32.67$ & \cellcolor{fly_green!12}$20.01$ & \cellcolor{fly_green!12}$46.64$ & \cellcolor{fly_green!12}$30.55$ & \cellcolor{fly_green!12}$1.27$ & \cellcolor{fly_green!12}$10.91$ & \cellcolor{fly_green!12}$25.50$ & \cellcolor{fly_green!12}$67.17$ & \cellcolor{fly_green!12}$24.21$ & \cellcolor{fly_green!12}$61.01$ & \cellcolor{fly_green!12}$41.30$ & \cellcolor{fly_red!12}$44.54$ & \cellcolor{fly_green!12}$6.21$
    \\\midrule
    \cellcolor{gray!10}\textcolor{gray}{Target}~\textcolor{gray}{$\bullet$} & \cellcolor{gray!10}\textcolor{gray}{$79.57$} & \cellcolor{gray!10}\textcolor{gray}{$52.25$} & \cellcolor{gray!10}\textcolor{gray}{$42.90$} & \cellcolor{gray!10}\textcolor{gray}{$23.30$} & \cellcolor{gray!10}\textcolor{gray}{$74.48$} & \cellcolor{gray!10}\textcolor{gray}{$39.40$} & \cellcolor{gray!10}\textcolor{gray}{$7.10$} & \cellcolor{gray!10}\textcolor{gray}{$0.33$} & \cellcolor{gray!10}\textcolor{gray}{$31.67$} & \cellcolor{gray!10}\textcolor{gray}{$71.96$} & \cellcolor{gray!10}\textcolor{gray}{$31.64$} & \cellcolor{gray!10}\textcolor{gray}{$67.87$} & \cellcolor{gray!10}\textcolor{gray}{$57.51$} & \cellcolor{gray!10}\textcolor{gray}{$66.14$} & \cellcolor{gray!10}\textcolor{gray}{$23.79$}
    \\\bottomrule
    \end{tabular}}
    \vspace{-0.2cm}
\label{tab:vehicle2drone}
\end{table*}

\noindent\textbf{Domain Adversarial Training.}
Each discriminator learns to distinguish between the feature representations from its respective domains, guiding the model to produce domain-agnostic features. Given feature outputs $\mathbf{F}^{\textcolor{fly_green}{\mathbf{v}}}$, $\mathbf{F}^{\textcolor{fly_red}{\mathbf{d}}}$, and $\Tilde{\mathbf{F}}$ for the source, target, and blended domains, respectively:
\begin{itemize}
    \item \textbf{Source-Blended Alignment:} Discriminator $\sigma_1$ classifies \( \mathbf{F}^{\textcolor{fly_green}{\mathbf{v}}} \) as source and \( \Tilde{\mathbf{F}} \) as blended, maximizing $\mathcal{L}_{\sigma_1}$ as:
    \begin{equation}
    - \sum\nolimits_{\mu, \nu} \left[(1 - z_1) \log \sigma_1(\Tilde{\mathbf{F}}) + z_1 \log \sigma_1(\mathbf{F}^{\textcolor{fly_green}{\mathbf{v}}})\right],
    \end{equation}
    where $z_1 = 1$ for source and $z_1 = 0$ for blended. This alignment ensures that the blended features preserve the supervised stability from the source.
    \item \textbf{Blended-Target Adaptation:} Discriminator $\sigma_2$ adapts $\Tilde{\mathbf{F}}$ toward high-activation regions in the target, where target-relevant properties are critical. It classifies $\Tilde{\mathbf{F}}$ as blended and $\mathbf{F}^{\textcolor{fly_red}{\mathbf{d}}}$ as target, maximizing $\mathcal{L}_{\sigma_2}$ as:
    \begin{equation}
     - \sum\nolimits_{\mu, \nu} \left[(1 - z_2) \log \sigma_2(\mathbf{F}^{\textcolor{fly_red}{\mathbf{d}}}) + z_2 \log \sigma_2(\Tilde{\mathbf{F}})\right],
    \end{equation}
    where $z_2 = 1$ for target and $z_2 = 0$ for blended. Here, $\sigma_2$ does not force complete target alignment but rather emphasizes adaptation in target-relevant areas, enabling $\Tilde{\mathbf{F}}$ to maintain a balanced representation across domains.
\end{itemize}

\noindent\textbf{Optimization.}
To achieve consistent cross-domain alignment, the backbone network is trained adversarially against both discriminators. The goal is to produce $\Tilde{\mathbf{F}}$ that each discriminator finds challenging to classify distinctly, thereby promoting a domain-agnostic representation:
\begin{equation}
    \mathcal{L}_{\mathrm{adv}} = - \sum\nolimits_{\mu, \nu} \left[\log \sigma_1(\Tilde{\mathbf{F}}) + \log \sigma_2(\Tilde{\mathbf{F}})\right].
\end{equation}
This objective ensures that the backbone network produces features for blended data that capture both source robustness and target adaptability, guided by high-activation areas where source-target consistency is most beneficial. To achieve this goal, we set two weight coefficients, $\phi_1$ and $\phi_2$, to balance between these two objectives.

During training, we employ a min-max optimization framework to iteratively update the backbone, $\sigma_1$, and $\sigma_2$:
\begin{itemize}
    \item \textbf{Step 1:} Update $\sigma_1$ and $\sigma_2$ to improve the classification of the source, blended, and target features.
    \item \textbf{Step 2:} Update the backbone network to minimize  $\mathcal{L}_{\mathrm{adv}}$, encouraging it to generate intermediary features that support robust cross-domain generalization.
\end{itemize}
Through this adversarial process, EventMatch allows the model to bridge the domain gap effectively, with blended features serving as a reliable intermediary that balances source stability with target-specific adaptability.

\noindent\textbf{Overall Framework.} As shown in \cref{fig:framework}, the source and blended domains share the same network, while the network from the target domain is updated via exponential-moving average (EMA), encouraging consistency among domains.

\subsection{The EXPo Benchmark}
\label{sec:expo_benchmark}
To facilitate robust event camera perception across diverse platforms and environments, we establish a large-scale event-based cross-platform adaptation benchmark: \textbf{\textit{\textcolor{fly_green}{E}\textcolor{fly_red}{X}\textcolor{fly_blue}{Po}}}. Our benchmark is based on \cite{chaney2023m3ed}. We include $89,228$ frames captured from three distinct platforms -- vehicle, drone, and quadruped -- spanning $21$ sequences: $6$ from the vehicle, $7$ from the drone, and $8$ from the quadruped. Following \textit{DSEC-Semantic} \cite{sun2022ess}, we support $11$ semantic classes relevant to real-world event-based perception and cover varied environments including city, urban, suburban, and rural scenes. 
Due to space limits, more details are in \textbf{Appendix}.
\section{Experiments}
\label{sec:experiments}

\begin{table*}[t]
    \centering
    \caption{\textbf{Benchmark results of platform adaptation} from \includegraphics[width=0.018\linewidth]{figures/icons/vehicle.png}~vehicle ($\mathcal{P}^{\textcolor{fly_green}{\mathbf{v}}}$) to \includegraphics[width=0.019\linewidth]{figures/icons/quadruped.png}~quadruped ($\mathcal{P}^{\textcolor{fly_blue}{\mathbf{q}}}$). \textcolor{gray}{Target} is trained with ground truth from the target domain. All scores are given in percentage (\%). The \colorbox{fly_red!12}{\textcolor{fly_red}{second best}} and \colorbox{fly_green!12}{\textcolor{fly_green}{best}} scores under each metric are in colors.}
    \vspace{-0.25cm}
    \resizebox{\linewidth}{!}{
    \begin{tabular}{r|p{26pt}<{\centering}p{26pt}<{\centering}p{26pt}<{\centering}p{26pt}<{\centering}|p{26pt}<{\centering}p{26pt}<{\centering}p{26pt}<{\centering}p{26pt}<{\centering}p{26pt}<{\centering}p{26pt}<{\centering}p{26pt}<{\centering}p{26pt}<{\centering}p{26pt}<{\centering}p{26pt}<{\centering}p{26pt}<{\centering}}
    \toprule
    \textbf{Method} & \textbf{Acc} & \textbf{mAcc} & \textbf{mIoU} & \textbf{fIoU} & \rotatebox{90}{\textcolor{background}{$\blacksquare$}~ground~} & \rotatebox{90}{\textcolor{building}{$\blacksquare$}~build} & \rotatebox{90}{\textcolor{fence}{$\blacksquare$}~fence} & \rotatebox{90}{\textcolor{person}{$\blacksquare$}~person} & \rotatebox{90}{\textcolor{pole}{$\blacksquare$}~pole} & \rotatebox{90}{\textcolor{road}{$\blacksquare$}~road} & \rotatebox{90}{\textcolor{sidewalk}{$\blacksquare$}~walk} & \rotatebox{90}{\textcolor{vegetation}{$\blacksquare$}~veg} & \rotatebox{90}{\textcolor{car}{$\blacksquare$}~car} & \rotatebox{90}{\textcolor{wall}{$\blacksquare$}~wall} & \rotatebox{90}{\textcolor{traffic-sign}{$\blacksquare$}~sign}
    \\\midrule\midrule
    \textcolor{gray}{Source-Only}~\textcolor{gray}{$\circ$} & \textcolor{gray}{$66.59$} & \textcolor{gray}{$39.73$} & \textcolor{gray}{$25.15$} & \textcolor{gray}{$16.52$} & \textcolor{gray}{$63.01$} & \textcolor{gray}{$39.26$} & \textcolor{gray}{$3.88$} & \textcolor{gray}{$17.88$} & \textcolor{gray}{$10.12$} & \textcolor{gray}{$51.67$} & \textcolor{gray}{$9.27$} & \textcolor{gray}{$68.02$} & \textcolor{gray}{$12.35$} & \textcolor{gray}{$0.24$} & \textcolor{gray}{$0.99$}
    \\\midrule
    AdaptSegNet \cite{tsai2018adaptsegnet} & $67.25$ & $48.73$ & $32.79$ & $14.89$ & $45.00$ & $45.88$ & $30.00$ & $34.92$ & $12.22$ & $55.50$ & $15.85$ & $73.84$ & $16.07$ & $31.35$ & $0.00$
    \\
    CBST \cite{zou2018cbst} & $69.25$ & $49.58$ & $35.06$ & $14.95$ & $47.39$ & $54.68$ & $34.27$ & $36.83$ & \cellcolor{fly_red!12}$13.78$ & $56.15$ & $18.13$ & $74.23$ & $16.18$ & $34.06$ & $0.00$
    \\
    IntraDA \cite{pan2020intra} & $68.29$ & $48.91$ & $34.25$ & $14.82$ & $43.75$ & $55.36$ & $32.64$ & $33.39$ & $11.60$ & $55.31$ & $17.00$ & $76.00$ & $20.30$ & $31.40$ & $0.00$
    \\
    DACS \cite{tranheden2021dacs} & $69.55$ & \cellcolor{fly_red!12}$53.88$ & $36.51$ & $14.66$ & $43.72$ & \cellcolor{fly_red!12}$57.27$ & \cellcolor{fly_red!12}$38.43$ & $35.42$ & \cellcolor{fly_green!12}$14.02$ & $57.10$ & $18.43$ & \cellcolor{fly_red!12}$76.16$ & $24.79$ & $36.21$ & $0.00$
    \\
    MIC \cite{hoyer2023mic} & $70.78$ & $49.22$ & $36.93$ & \cellcolor{fly_red!12}$15.60$ & \cellcolor{fly_green!12}$51.71$ & $51.73$ & $33.54$ & $38.10$ & $9.44$ & $54.27$ & $20.74$ & $74.40$ & \cellcolor{fly_green!12}$29.79$ & \cellcolor{fly_red!12}$41.78$ & \cellcolor{fly_red!12}$0.70$
    \\
    PLSR \cite{zhao2024plsr} & \cellcolor{fly_red!12}$70.91$ & $53.65$ & \cellcolor{fly_red!12}$37.57$ & $15.25$ & $49.04$ & $53.28$ & $37.54$ & $36.64$ & $12.91$ & \cellcolor{fly_red!12}$57.60$ & \cellcolor{fly_green!12}$25.29$ & $75.92$ & $24.92$ & $39.85$ & $0.24$
    \\
    \textbf{\textcolor{fly_green}{Event}\textcolor{fly_red}{Fly}} \textbf{(Ours)} & \cellcolor{fly_green!12}$73.42$ & \cellcolor{fly_green!12}$54.14$ & \cellcolor{fly_green!12}$40.05$ & \cellcolor{fly_green!12}$15.78$ & \cellcolor{fly_red!12}$50.07$ & \cellcolor{fly_green!12}$61.33$ & \cellcolor{fly_green!12}$39.17$ & \cellcolor{fly_green!12}$41.97$ & $12.83$ & \cellcolor{fly_green!12}$59.14$ & \cellcolor{fly_red!12}$23.51$ & \cellcolor{fly_green!12}$79.80$ & \cellcolor{fly_red!12}$27.26$ & \cellcolor{fly_green!12}$42.65$ & \cellcolor{fly_green!12}$2.86$
    \\\midrule
    \cellcolor{gray!10}\textcolor{gray}{Target}~\textcolor{gray}{$\bullet$} & \cellcolor{gray!10}\textcolor{gray}{$80.02$} & \cellcolor{gray!10}\textcolor{gray}{$60.55$} & \cellcolor{gray!10}\textcolor{gray}{$49.84$} & \cellcolor{gray!10}\textcolor{gray}{$19.58$} & \cellcolor{gray!10}\textcolor{gray}{$74.80$} & \cellcolor{gray!10}\textcolor{gray}{$56.23$} & \cellcolor{gray!10}\textcolor{gray}{$46.08$} & \cellcolor{gray!10}\textcolor{gray}{$55.28$} & \cellcolor{gray!10}\textcolor{gray}{$21.79$} & \cellcolor{gray!10}\textcolor{gray}{$59.90$} & \cellcolor{gray!10}\textcolor{gray}{$30.31$} & \cellcolor{gray!10}\textcolor{gray}{$77.24$} & \cellcolor{gray!10}\textcolor{gray}{$58.38$} & \cellcolor{gray!10}\textcolor{gray}{$62.47$} & \cellcolor{gray!10}\textcolor{gray}{$5.81$}
    \\\bottomrule
    \end{tabular}}
    \vspace{-0.15cm}
\label{tab:vehicle2quadruped}
\end{table*}
\begin{table*}[t]
    \centering
    \caption{
        \textbf{Benchmark results of platform adaptation} among the \includegraphics[width=0.018\linewidth]{figures/icons/vehicle.png}~vehicle ($\mathcal{P}^{\textcolor{fly_green}{\mathbf{v}}}$), \includegraphics[width=0.021\linewidth]{figures/icons/drone.png}~drone ($\mathcal{P}^{\textcolor{fly_red}{\mathbf{d}}}$), and \includegraphics[width=0.019\linewidth]{figures/icons/quadruped.png}~quadruped ($\mathcal{P}^{\textcolor{fly_blue}{\mathbf{q}}}$) platforms, respectively. A total of \textbf{six} cross-platform adaptation settings are considered in our benchmark. \textcolor{gray}{Target} is trained with ground truth from the target domain. All scores are given in percentage (\%). The \colorbox{fly_red!12}{\textcolor{fly_red}{second best}} and \colorbox{fly_green!12}{\textcolor{fly_green}{best}} scores under each metric are highlighted in colors.}
    \vspace{-0.25cm}
    \resizebox{\linewidth}{!}{
    \begin{tabular}{r|p{27pt}<{\centering}p{27pt}<{\centering}p{27pt}<{\centering}p{27pt}<{\centering}|p{27pt}<{\centering}p{27pt}<{\centering}p{27pt}<{\centering}p{27pt}<{\centering}|p{27pt}<{\centering}p{27pt}<{\centering}p{27pt}<{\centering}p{27pt}<{\centering}|p{27pt}<{\centering}p{27pt}<{\centering}}
    \toprule
    \multirow{2}{*}{\textbf{Method}} & \multicolumn{2}{c}{$\mathcal{P}^{\textcolor{fly_green}{\mathbf{v}}}$ $\rightarrow$ $\mathcal{P}^{\textcolor{fly_red}{\mathbf{d}}}$} & \multicolumn{2}{c|}{$\mathcal{P}^{\textcolor{fly_green}{\mathbf{v}}}$ $\rightarrow$ $\mathcal{P}^{\textcolor{fly_blue}{\mathbf{q}}}$} & \multicolumn{2}{c}{$\mathcal{P}^{\textcolor{fly_red}{\mathbf{d}}}$ $\rightarrow$ $\mathcal{P}^{\textcolor{fly_green}{\mathbf{v}}}$} & \multicolumn{2}{c|}{$\mathcal{P}^{\textcolor{fly_red}{\mathbf{d}}}$ $\rightarrow$ $\mathcal{P}^{\textcolor{fly_blue}{\mathbf{q}}}$} & \multicolumn{2}{c}{$\mathcal{P}^{\textcolor{fly_blue}{\mathbf{q}}}$ $\rightarrow$ $\mathcal{P}^{\textcolor{fly_green}{\mathbf{v}}}$} & \multicolumn{2}{c|}{$\mathcal{P}^{\textcolor{fly_blue}{\mathbf{q}}}$ $\rightarrow$ $\mathcal{P}^{\textcolor{fly_red}{\mathbf{d}}}$} & \multicolumn{2}{c}{\textbf{Average}}
    \\
    & \textbf{Acc} & \textbf{mIoU} & \textbf{Acc} & \textbf{mIoU} & \textbf{Acc} & \textbf{mIoU} & \textbf{Acc} & \textbf{mIoU} & \textbf{Acc} & \textbf{mIoU} & \textbf{Acc} & \textbf{mIoU} & \textbf{Acc} & \textbf{mIoU}
    \\\midrule\midrule
    \textcolor{gray}{Source-Only}~\textcolor{gray}{$\circ$} & \textcolor{gray}{$43.69$} & \textcolor{gray}{$15.04$} & \textcolor{gray}{$66.59$} & \textcolor{gray}{$25.15$} & \textcolor{gray}{$57.91$} & \textcolor{gray}{$20.79$} & \textcolor{gray}{$66.83$} & \textcolor{gray}{$23.06$} & \textcolor{gray}{$57.49$} & \textcolor{gray}{$21.30$} & \textcolor{gray}{$52.62$} & \textcolor{gray}{$16.85$} & \textcolor{gray}{$57.52$} & \textcolor{gray}{$20.37$}
    \\\midrule
    AdaptSegNet \cite{tsai2018adaptsegnet} & $49.14$ & $21.16$ & $67.25$ & $32.79$ & $68.29$ & $29.55$ & $67.57$ & $33.99$ & $66.74$ & $30.65$ & $57.07$ & $20.96$ & $62.88$ & $28.18$
    \\
    DACS \cite{tranheden2021dacs} & $59.81$ & $27.07$ & $69.55$ & $36.51$ & $71.78$ & $36.10$ & $67.73$ & $36.11$ & $71.20$ & $34.78$ & $60.74$ & $24.50$ & $66.80$ & $32.51$
    \\
    MIC \cite{hoyer2023mic} & $63.11$ & $28.87$ & $70.78$ & $36.93$ & \cellcolor{fly_red!12}$72.46$ & $36.88$ & $67.29$ & \cellcolor{fly_red!12}$36.27$ & $72.46$ & $35.22$ & \cellcolor{fly_red!12}$64.49$ & $26.11$ & $68.43$ & $33.38$
    \\
    PLSR \cite{zhao2024plsr} & \cellcolor{fly_red!12}$64.61$ & \cellcolor{fly_red!12}$29.69$ & \cellcolor{fly_red!12}$70.91$ & \cellcolor{fly_red!12}$37.57$ & \cellcolor{fly_red!12}$72.46$ & \cellcolor{fly_red!12}$37.18$ & \cellcolor{fly_red!12}$67.83$ & $36.21$ & \cellcolor{fly_red!12}$72.93$ & \cellcolor{fly_red!12}$36.38$ & $63.57$ & \cellcolor{fly_red!12}$27.34$ & \cellcolor{fly_red!12}$68.72$ & \cellcolor{fly_red!12}$34.06$
    \\
    \textbf{\textcolor{fly_green}{Event}\textcolor{fly_red}{Fly}} \textbf{(Ours)} & \cellcolor{fly_green!12}$69.17$ & \cellcolor{fly_green!12}$32.67$ & \cellcolor{fly_green!12}$73.42$ & \cellcolor{fly_green!12}$40.05$ & \cellcolor{fly_green!12}$75.50$ & \cellcolor{fly_green!12}$39.92$ & \cellcolor{fly_green!12}$69.68$ & \cellcolor{fly_green!12}$37.37$ & \cellcolor{fly_green!12}$73.93$ & \cellcolor{fly_green!12}$37.70$ & \cellcolor{fly_green!12}$65.78$ & \cellcolor{fly_green!12}$28.79$ & \cellcolor{fly_green!12}$71.25$ & \cellcolor{fly_green!12}$36.08$
    \\\midrule
    \cellcolor{gray!10}\textcolor{gray}{Target}~\textcolor{gray}{$\bullet$} & \cellcolor{gray!10}\textcolor{gray}{$79.57$} & \cellcolor{gray!10}\textcolor{gray}{$42.90$} & \cellcolor{gray!10}\textcolor{gray}{$80.02$} & \cellcolor{gray!10}\textcolor{gray}{$49.84$} & \cellcolor{gray!10}\textcolor{gray}{$86.12$} & \cellcolor{gray!10}\textcolor{gray}{$55.93$} & \cellcolor{gray!10}\textcolor{gray}{$80.02$} & \cellcolor{gray!10}\textcolor{gray}{$49.84$} & \cellcolor{gray!10}\textcolor{gray}{$86.12$} & \cellcolor{gray!10}\textcolor{gray}{$55.93$} & \cellcolor{gray!10}\textcolor{gray}{$79.57$} & \cellcolor{gray!10}\textcolor{gray}{$42.90$} & \cellcolor{gray!10}\textcolor{gray}{$81.90$} & \cellcolor{gray!10}\textcolor{gray}{$49.56$}
    \\\bottomrule
\end{tabular}}
\vspace{-0.2cm}
\label{tab:benchmark}
\end{table*}

\subsection{Settings}
\noindent\textbf{Implementation Details.}
Our framework is implemented using PyTorch \cite{paszke2019pytorch}. The backbone network is from E2VID \cite{rebecq2019e2vid}, while the event segmentation head is from ESS \cite{sun2022ess}. The hyperparameters $\tau$, $\phi_1$ and $\phi_2$ are set to $0.4$, $1$e$-3$, and $2$e$-3$, respectively. The model is optimized using AdamW \cite{loshchilov2019adamw} and OneCycle learning rate scheduler \cite{onecycle} for $100$k iterations. The learning rate and batch size are set to $1$e$-3$ and $8$. Due to the lack of baselines, we reproduce popular frame-based adaptation methods for adversarial training, contrastive learning, and self-training \cite{tsai2018adaptsegnet,zou2018cbst,pan2020intra,tranheden2021dacs,hoyer2023mic,zhao2024plsr}. To ensure fair comparisons, all methods adopt the same backbone and training iterations. The models are trained using NVIDIA RTX 4090 GPUs.

\noindent\textbf{Evaluation Metrics.}
Following conventions, we use Accuracy (Acc) and Intersection-over-Union (IoU) as the main metrics for evaluation. We also report mean Acc (mAcc), mean IoU (mIoU), and frequency-weighted IoU (fIoU) for a holistic comparison among different adaptation methods.

\begin{figure*}
    \centering
    \includegraphics[width=\linewidth]{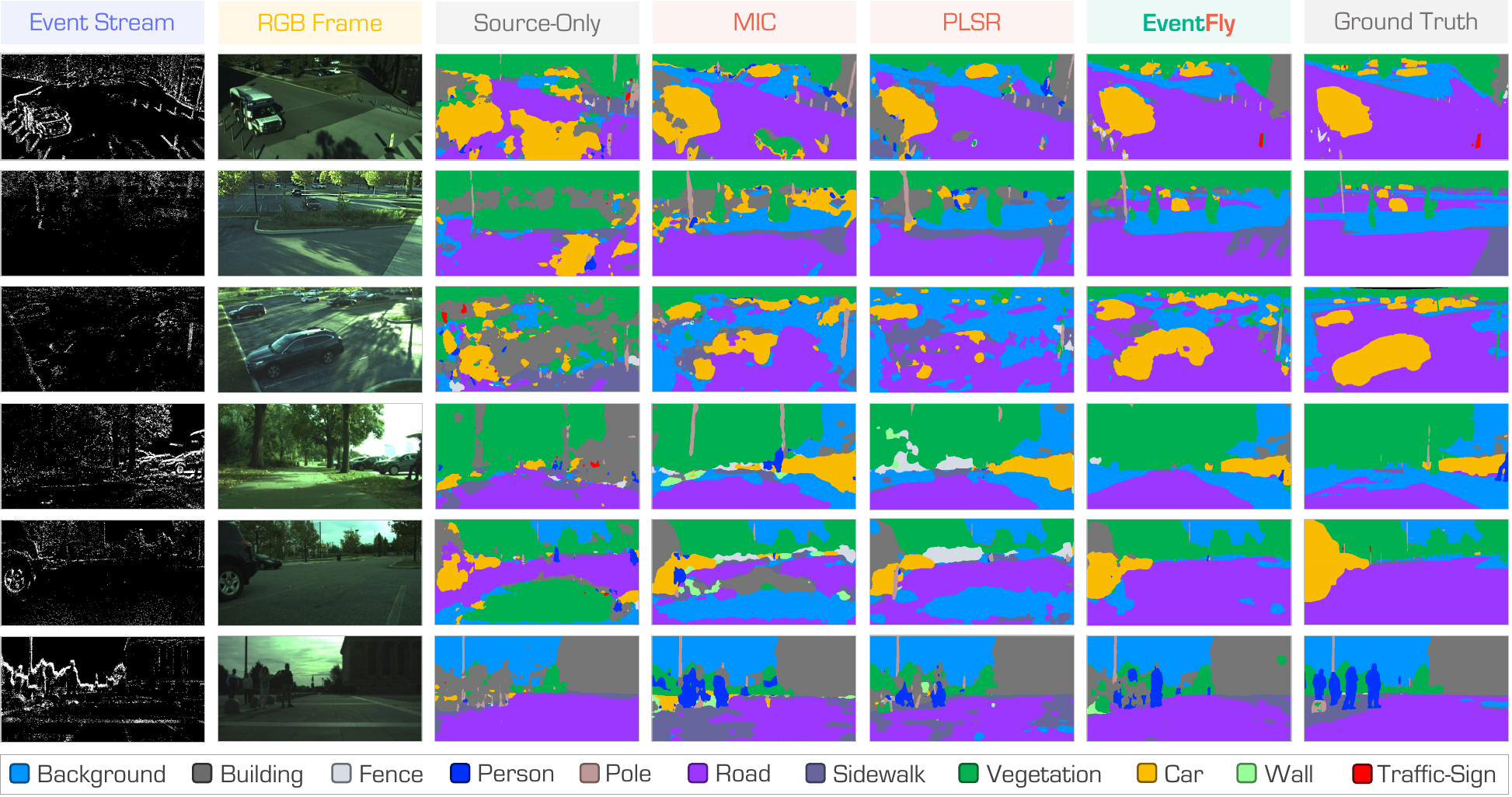}
    \vspace{-0.7cm}
    \caption{\textbf{Qualitative assessments of adaptation} from \includegraphics[width=0.018\linewidth]{figures/icons/vehicle.png}~vehicle to \includegraphics[width=0.021\linewidth]{figures/icons/drone.png}~drone (the first $3$ rows), and from \includegraphics[width=0.018\linewidth]{figures/icons/vehicle.png}~vehicle to \includegraphics[width=0.019\linewidth]{figures/icons/quadruped.png}~quadruped (the last $3$ rows). We use grayscaled event images for better visibility. The RGB frames are for reference purposes only. Best viewed in colors.}
    \label{fig:qualitative}
    \vspace{-0.2cm}
\end{figure*}

\subsection{Comparative Study}
\noindent\textbf{Adapt from Vehicle to Drone.}
\cref{tab:vehicle2drone} presents the results of transferring \textit{vehicle} to \textit{drone}, where there is a huge domain gap in between. The \textit{source-only} ($15.04\%$ mIoU) and supervised \textit{target} ($42.90\%$ mIoU) models validate this gap. Prior adaptation methods show good results in closing the domain gap, while our \textbf{\textit{\textcolor{fly_green}{Event}\textcolor{fly_red}{Fly}}} surpasses all competitors for almost all semantic classes, demonstrating strong robustness enhancements for event-based cross-platform adaptation.

\noindent\textbf{Adapt from Vehicle to Quadruped.}
We compare the adaptation methods from \textit{vehicle} to \textit{quadruped} in \cref{tab:vehicle2quadruped}. Since the data and semantic distributions of these two platforms are closer (compared to \textit{drone}), the adaptation methods show larger performance gains. We achieve a promising $40.05\%$ mIoU under this setting, which corresponds to a $59.2\%$ relative improvement over the \textit{source-only} baseline.

\noindent\textbf{Qualitative Assessments.}
\cref{fig:qualitative} visually compares the abilities of different cross-platform adaptation methods. As can be seen, the large domain gap between source and target platforms causes huge performance degradation, resulting in messy semantic predictions (\textit{source-only}). While state-of-the-art adaptation methods \cite{hoyer2023mic,zhao2024plsr} restore the performance to some extent, their predictions still suffer from ambiguous object and background contours. Differently, our \textbf{\textit{\textcolor{fly_green}{Event}\textcolor{fly_red}{Fly}}} generates more accurate scene semantics, credited to the superior robustness from the EAP-driven designs.

\noindent\textbf{Cross-Platform Adaptations.}
Since our \textbf{\textit{\textcolor{fly_green}{E}\textcolor{fly_red}{X}\textcolor{fly_blue}{Po}}} benchmark consists of three platforms, we benchmark state-of-the-art methods across a total of six adaptation settings, where each platform interchangeably serve as the source/target domains. As shown in \cref{tab:benchmark}, we observe discrepancies in adaptation difficulties. Specifically, the \textit{drone} domain is the most difficult to adapt to, mainly due to its unique motion patterns and perspective-based dynamic. Our approach exhibits better robustness than prior arts, achieving the best metrics across all six settings. Such a strong generalization ability is crucial for the deployment of event camera perception algorithms in real-world environments.

\begin{table}[t]
    \centering
    \caption{\textbf{Ablation results of each component} in cross-platform adaptation from \includegraphics[width=0.036\linewidth]{figures/icons/vehicle.png}~vehicle ($\mathcal{P}^{\textcolor{fly_green}{\mathbf{v}}}$) to \includegraphics[width=0.042\linewidth]{figures/icons/drone.png}~drone ($\mathcal{P}^{\textcolor{fly_red}{\mathbf{d}}}$) and \includegraphics[width=0.036\linewidth]{figures/icons/vehicle.png}~vehicle ($\mathcal{P}^{\textcolor{fly_green}{\mathbf{v}}}$) to \includegraphics[width=0.038\linewidth]{figures/icons/quadruped.png}~quadruped ($\mathcal{P}^{\textcolor{fly_blue}{\mathbf{q}}}$), respectively. \textbf{Dual:} The dual-branch network. \textbf{Match:} The EventMatch technique. \textbf{Blend:} The EventBlend operation. All scores are given in percentage (\%). We compare settings from the $(a)$ source-only, $(b)$ single-component effects, $(c)$ double-component effects, $(d)$ full configuration, and $(e)$ supervised target results, respectively, in this ablation study.}
    \vspace{-0.25cm}
    \resizebox{\linewidth}{!}{
    \begin{tabular}{c|p{24pt}<{\centering}p{27pt}<{\centering}p{27pt}<{\centering}|p{28pt}<{\centering}p{28pt}<{\centering}|p{28pt}<{\centering}p{28pt}<{\centering}}
        \toprule
        \multirow{2}{*}{\textbf{\#}} & \multirow{2}{*}{\textbf{Dual}} & \multirow{2}{*}{\textbf{Match}} & \multirow{2}{*}{\textbf{Blend}} & \multicolumn{2}{c|}{$\mathcal{P}^{\textcolor{fly_green}{\mathbf{v}}}$ $\rightarrow$ $\mathcal{P}^{\textcolor{fly_red}{\mathbf{d}}}$} & \multicolumn{2}{c}{$\mathcal{P}^{\textcolor{fly_green}{\mathbf{v}}}$ $\rightarrow$ $\mathcal{P}^{\textcolor{fly_blue}{\mathbf{q}}}$}
        \\
        & & & & \textbf{Acc} & \textbf{mIoU} & \textbf{Acc} & \textbf{mIoU}
        \\\midrule\midrule
        $(a)$ & \textcolor{fly_red}{\xmark} & \textcolor{fly_red}{\xmark} & \textcolor{fly_red}{\xmark} & \textcolor{gray}{$43.69$} & \textcolor{gray}{$15.04$} & \textcolor{gray}{$66.59$} & \textcolor{gray}{$25.15$}
        \\\midrule
        \multirow{2}{*}{$(b)$} & \textcolor{fly_red}{\xmark} & \textcolor{fly_green}{\cmark} & \textcolor{fly_red}{\xmark} & $59.86$ & $25.28$ & $67.74$ & $33.66$
        \\
        & \textcolor{fly_red}{\xmark} & \textcolor{fly_red}{\xmark} & \textcolor{fly_green}{\cmark} & $61.85$ & $27.91$ & $69.55$ & $36.82$
        \\\midrule
        \multirow{2}{*}{$(c)$} & \textcolor{fly_green}{\cmark} & \textcolor{fly_green}{\cmark} & \textcolor{fly_red}{\xmark} & $66.50$ & $30.24$ & $70.90$ & $38.10$
        \\
        & \textcolor{fly_green}{\cmark} & \textcolor{fly_red}{\xmark} & \textcolor{fly_green}{\cmark} & $68.43$ & $31.96$ & $72.67$ & $39.11$
        \\\midrule
        $(d)$ & \textcolor{fly_green}{\cmark} & \textcolor{fly_green}{\cmark} & \textcolor{fly_green}{\cmark} & \cellcolor{fly_green!12}$69.17$ & \cellcolor{fly_green!12}$32.67$ & \cellcolor{fly_green!12}$73.42$ & \cellcolor{fly_green!12}$40.05$
        \\\midrule
        $(e)$ & \cellcolor{gray!10} & \cellcolor{gray!10}{\textcolor{gray}{Target}} & \cellcolor{gray!10} & \cellcolor{gray!10}\textcolor{gray}{$79.57$} & \cellcolor{gray!10}\textcolor{gray}{$42.90$} & \cellcolor{gray!10}\textcolor{gray}{$80.02$} & \cellcolor{gray!10}\textcolor{gray}{$49.84$}
        \\\bottomrule
    \end{tabular}}
    \vspace{-0.4cm}
\label{tab:ablation}
\end{table}

\begin{figure}
    \centering
    \includegraphics[width=\linewidth]{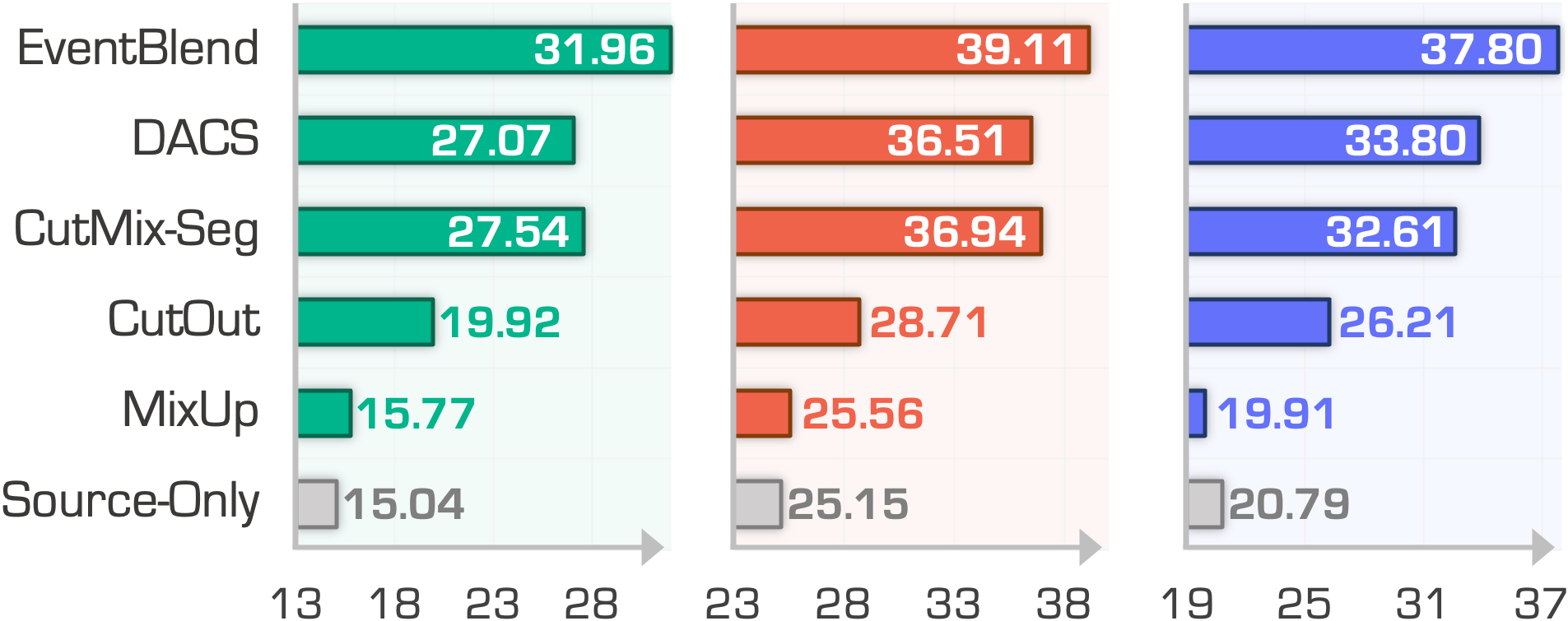}
    \vspace{-0.6cm}
    \caption{\textbf{Ablation study} (mIoU scores in \%) on different domain-mixing techniques under the (left) $\mathcal{P}^{\textcolor{fly_green}{\mathbf{v}}}$ $\rightarrow$ $\mathcal{P}^{\textcolor{fly_red}{\mathbf{d}}}$, (middle) $\mathcal{P}^{\textcolor{fly_green}{\mathbf{v}}}$ $\rightarrow$ $\mathcal{P}^{\textcolor{fly_blue}{\mathbf{q}}}$, and (right) $\mathcal{P}^{\textcolor{fly_red}{\mathbf{d}}}$ $\rightarrow$ $\mathcal{P}^{\textcolor{fly_green}{\mathbf{v}}}$ cross-platform adaptation settings.}
    \label{fig:blend}
    \vspace{-0.29cm}
\end{figure}

\subsection{Ablation Study}

\noindent\textbf{Component Analysis.}
Our \textbf{\textit{\textcolor{fly_green}{Event}\textcolor{fly_red}{Fly}}} framework compromises three key components: EventBlend, EventMatch, and the dual-backbone network. As shown in \cref{tab:ablation}, each contribution provides distinct performance improvements. Since our designs are well-motivated by EAP, the combinations of any two (row $\#c$) and all (row $\#d$) of them further yield better cross-platform adaptation performance. 

\noindent\textbf{Domain Blending Techniques.}
To validate that our EAP-driven EventBlend operation indeed encourages domain regularization effect, we compare the heuristic data mixing techniques \cite{zhang2018mixup,devries2017cutout} and recent domain mixing methods \cite{french2020cutmix-seg,tranheden2021dacs}. As shown in \cref{fig:blend}, conventional methods, which neglect the event activation discrepancies, show sub-par performance in bridging the domains. Guided by EAP, our domain blending technique facilitates entropy minimization in high-activation regions, which in turn brings robust feature adaptation across heterogeneous platforms.


\noindent\textbf{Similarity Threshold.}
The strength of blending source and target platforms is vital in our framework. Therefore, we study the effect of the threshold parameter $\tau$. Notably, the $0$ (\ie, \textit{source-only}) or high ($>0.7$) value of $\tau$ brings limited gains, as the domain consistency regularization is weak. We found that the best possible trade-offs appear between $0.3$ and $0.5$, further validating the importance of bridging domains for cross-platform adaptation.
\vspace{0.1cm}
\section{Conclusion}
\label{sec:conclusion}
In this work, we presented \textbf{\textit{\textcolor{fly_green}{Event}\textcolor{fly_red}{Fly}}}, a framework for robust cross-platform adaptation in event-based dense perception, enabling robust deployment across the \textit{vehicle}, \textit{drone}, and \textit{quadruped} platforms. Leveraging the proposed Event Activation Prior (EAP), EventBlend, and EventMatch, we addressed the unique adaptation challenges of event stream data collected from different robot platforms. We also introduced \textbf{\textit{\textcolor{fly_green}{E}\textcolor{fly_red}{X}\textcolor{fly_blue}{Po}}}, a large-scale benchmark designed to evaluate cross-platform event perception capabilities. Our approach demonstrated notable improvements over existing methods. We hope this work lays the foundation for more adaptive event-based perception across real-world environments.

\section*{Acknowledgments}
This work is under the programme DesCartes and is supported by the National Research Foundation, Prime Minister’s Office, Singapore, under its Campus for Research Excellence and Technological Enterprise (CREATE) programme. This work is also supported by the Apple Scholars in AI/ML Ph.D. Fellowship program.

The authors would like to sincerely thank the Program Chairs, Area Chairs, and Reviewers for the time and efforts devoted during the review process.

\section*{Table of Contents}
\startcontents[appendices]
\printcontents[appendices]{l}{1}{\setcounter{tocdepth}{3}}

\section{The EXPo Benchmark}
\label{sec_supp:benchmark}

In this section, we elaborate on the data structure, definitions, configurations, statistics, and visual examples of the proposed \textbf{\textit{\textcolor{fly_green}{E}\textcolor{fly_red}{X}\textcolor{fly_blue}{Po}}} (\textbf{\textit{\textcolor{fly_green}{E}}}vent-based \textbf{\textcolor{fly_red}{\textit{X}}}ross-\textbf{\textcolor{fly_blue}{\textit{P}}}latf\textbf{\textcolor{fly_blue}{\textit{o}}}rm perception) benchmark.

\begin{table}[t]
    \centering
    \caption{The summary of platform-level statistics in \textbf{\textit{\textcolor{fly_green}{E}\textcolor{fly_red}{X}\textcolor{fly_blue}{Po}}}.}
    \vspace{-0.2cm}
    \resizebox{\linewidth}{!}{
    \begin{tabular}{l|p{62pt}<{\centering}|p{62pt}<{\centering}|p{62pt}<{\centering}}
    \toprule
    \textbf{Platform} & \includegraphics[width=0.16\linewidth]{figures/icons/vehicle.png} \textcolor{fly_green}{\textbf{Vehicle}} & \includegraphics[width=0.18\linewidth]{figures/icons/drone.png} \textcolor{fly_red}{\textbf{Drone}} & \includegraphics[width=0.16\linewidth]{figures/icons/quadruped.png} \textcolor{fly_blue}{\textbf{Quadruped}}
    \\\midrule\midrule
    Frame (train) & $30,321$ & $13,458$ & $17,302$
    \\
    Frame (val) & $12,998$ & $5,772$ & $7,421$
    \\
    Res. ($H$, $W$) & $360\times 640$ & $360\times 640$ & $360\times 640$
    \\
    Res. ($T$) & $20$ & $20$ & $20$ 
    \\
    Duration & $5,000,000$ & $5,000,000$ & $5,000,000$
    \\\midrule
    Semantics & \multicolumn{3}{c}{$19$ Classes / $11$ Classes}
    \\\midrule
    Environment & \multicolumn{3}{c}{City, Urban, Suburban, Rural}
    \\\bottomrule
    \end{tabular}}
    \label{tab:expo_benchmark}
\end{table}

\subsection{Benchmark Overview}
\label{sec_supp:bench_overview}
Our \textbf{\textit{\textcolor{fly_green}{E}\textcolor{fly_red}{X}\textcolor{fly_blue}{Po}}} benchmark serves as the first comprehensive effort to tackle the challenging task of cross-platform adaptation for event camera perception. Building upon the newly launched M3ED dataset \cite{chaney2023m3ed}, our benchmark focuses on enabling robust, domain-adaptive perception across diverse robotic platforms. By incorporating a rich variety of event data and semantic labels, we aim to highlight key discrepancies among platforms and provide a robust testbed for evaluating cross-platform performance.

\cref{tab:expo_benchmark} provides the platform-level statistics of each platform. The overall benchmark consists of $89,228$ frames collected from three distinct platforms -- \textit{vehicle}, \textit{drone}, and \textit{quadruped} -- across $21$ sequences: $6$ from the vehicle, $7$ from the drone, and $8$ from the quadruped. The sequences capture a wide range of dynamic real-world scenarios and span diverse environments, including city, urban, suburban, and rural scenes. This diversity ensures that the benchmark covers both structured and unstructured environments, replicating real-world challenges faced by event cameras deployed across different robotic platforms.

\subsection{Cross-Platform Configurations}
\label{sec_supp:bench_config}

The \textbf{\textit{\textcolor{fly_green}{E}\textcolor{fly_red}{X}\textcolor{fly_blue}{Po}}} benchmark aims to highlight platform-specific discrepancies, such as motion dynamics, perspectives, and environmental interactions. Specifically, ground vehicles capture low-altitude perspectives with dense surface-level details, such as roads, curbs, and obstacles. Drones provide high-altitude views with sparse ground-level features, focusing on landscapes, buildings, and environmental structures. Quadrupeds, on the other hand, operate closer to human eye levels, capturing mixed indoor-outdoor dynamics and a wider range of semantic elements. These platform-specific variations make this benchmark a holistic resource for studying domain-specific adaptation and developing robust models capable of generalizing across diverse operational settings.

The event camera data in our benchmark is collected using the Prophesee Gen 4 (EVKv4) event camera \cite{finateu2020gen4}, a state-of-the-art sensor known for its high temporal resolution and dynamic range. This sensor offers a spatial resolution of $720 \times 1280$ pixels and a field of view of $63^{\circ} \times 38^{\circ}$. This consistent sensor setup is employed across all three platforms, ensuring that the observed domain gaps arise purely from platform-specific differences, such as variations in motion patterns, viewpoint dynamics, and environmental interactions, rather than discrepancies in sensor specifications.
By eliminating sensor-level variations, the benchmark ensures that the adaptation challenge remains focused on the core differences between the platforms. This configuration not only strengthens our validity for cross-platform adaptation but also facilitates meaningful comparisons of model performance across varied operational contexts.

\subsection{Benchmark Structure}
\label{sec_supp:bench_data_structure}

The \textbf{\textit{\textcolor{fly_green}{E}\textcolor{fly_red}{X}\textcolor{fly_blue}{Po}}} benchmark comprises $21$ sequences distributed across three platforms: $6$ sequences for the \textit{vehicle}, $7$ sequences for the \textit{drone}, and $8$ sequences for the \textit{quadruped}. \cref{tab:dataset_structure} provides a detailed breakdown of the dataset structure and sequence information for each platform.

Specifically, the benchmark includes $43,766$ frames from the \textit{vehicle} platform, $19,899$ frames from the \textit{drone} platform, and $25,563$ frames from the \textit{quadruped} platform, resulting in a total of $89,228$ frames. The detailed information for each sequence across the three platforms is shown in \cref{tab:dataset_structure}. This extensive collection makes \textbf{\textit{\textcolor{fly_green}{E}\textcolor{fly_red}{X}\textcolor{fly_blue}{Po}}} the largest benchmark for event camera perception.

As shown in \cref{tab:expo_benchmark}, we split each platform into two subsets: training set and validation set. We sample for each sequence in each platform the last $40\%$ of frames for validation, and use the remaining data for training. In total, there are $61,081$ frames for training and $26,191$ frames for validation. Since the original spatial resolution is high, we subsample it from $720 \times 1280$ pixels to $360 \times 640$ pixels, \ie, resize both the height and width to half of the original values. Following the setting of DSEC-Semantic \cite{sun2022ess}, the temporal resolution is set to $20$ (bins). Additionally, the duration $\Delta T$ is set to $5,000,000$.

\begin{table}[t]
    \centering
    \caption{\textbf{The dataset structure and sequence information} among the \includegraphics[width=0.036\linewidth]{figures/icons/vehicle.png}~vehicle ($\mathcal{P}^{\textcolor{fly_green}{\mathbf{v}}}$), \includegraphics[width=0.042\linewidth]{figures/icons/drone.png}~drone ($\mathcal{P}^{\textcolor{fly_red}{\mathbf{d}}}$), and \includegraphics[width=0.038\linewidth]{figures/icons/quadruped.png}~quadruped ($\mathcal{P}^{\textcolor{fly_blue}{\mathbf{q}}}$) platforms, respectively, in the proposed \textbf{\textit{\textcolor{fly_green}{E}\textcolor{fly_red}{X}\textcolor{fly_blue}{Po}}} benchmark.}
    \vspace{-0.2cm}
    \resizebox{\linewidth}{!}{
    \begin{tabular}{c|l|c|c}
        \toprule
        \textbf{Platform} & \textbf{Sequence Name} & \textbf{\# Frames} & \textbf{Total}
        \\\midrule\midrule
        \multirow{6}{*}{\textbf{Vehicle}} & \texttt{horse} & $714$ & \multirow{6}{*}{\textcolor{fly_green}{$\mathbf{43,766}$}}
        \\
        & \texttt{penno\_small\_loop} & $1,102$ & 
        \\
        & \texttt{rittenhouse} & $9,752$ & 
        \\
        & \texttt{ucity\_small\_loop} & $16,867$ & 
        \\
        & \texttt{city\_hall} & $7,453$ & 
        \\
        & \texttt{penno\_big\_loop} & $7,878$ &
        \\\midrule
        \multirow{7}{*}{\textbf{Drone}} & \texttt{fast\_flight\_1} & $2,229$ & \multirow{7}{*}{\textcolor{fly_red}{$\mathbf{19,899}$}}
        \\
        & \texttt{fast\_flight\_2} & $4,077$ & 
        \\
        & \texttt{penno\_parking\_1} & $2,810$ & 
        \\
        & \texttt{penno\_parking\_2} & $2,713$ & 
        \\
        & \texttt{penno\_plaza} & $1,694$ &
        \\
        & \texttt{penno\_cars} & $3,073$ &
        \\
        & \texttt{penno\_trees} & $3,303$ &
        \\\midrule
        \multirow{8}{*}{\textbf{Quadruped}} & \texttt{penno\_short\_loop} & $2,942$ & \multirow{8}{*}{\textcolor{fly_blue}{$\mathbf{25,563}$}}
        \\
        & \texttt{skatepark\_1} & $2,305$ & 
        \\
        & \texttt{skatepark\_2} & $1,652$ &
        \\
        & \texttt{srt\_green\_loop} & $1,597$ & 
        \\
        & \texttt{srt\_under\_bridge\_1} & $5,083$ &
        \\
        & \texttt{srt\_under\_bridge\_2} & $4,533$ & 
        \\
        & \texttt{art\_plaza\_loop} & $3,615$ & 
        \\
        & \texttt{rocky\_steps} & $3,836$ &
        \\\bottomrule
    \end{tabular}}
\label{tab:dataset_structure}
\end{table}

\subsection{Semantic Definitions}
\label{sec_supp:bench_definition}

The \textbf{\textit{\textcolor{fly_green}{E}\textcolor{fly_red}{X}\textcolor{fly_blue}{Po}}} benchmark consists of a total of $19$ semantic classes, which ensure a holistic dense perception for the event camera scenes acquired by the three platforms. The specific definition of each class is listed as follows:
\begin{itemize}
    \item \textcolor{road}{$\blacksquare$}~\texttt{road} (ID: $0$): The drivable surface designed for vehicle travel, typically marked by lanes and boundaries.

    \item \textcolor{sidewalk}{$\blacksquare$}~\texttt{sidewalk} (ID: $1$): Elevated pathways adjacent to roads, designated for pedestrian use.

    \item \textcolor{building}{$\blacksquare$}~\texttt{building} (ID: $2$): Permanent structures designed for residential, commercial, or industrial purposes.
    
    \item \textcolor{wall}{$\blacksquare$}~\texttt{wall} (ID: $3$): Vertical structures that enclose or divide areas, often used for security or boundary delineation.
    
    \item \textcolor{fence}{$\blacksquare$}~\texttt{fence} (ID: $4$): Lightweight barriers, usually made of wood or metal, marking boundaries or containing areas.
    
    \item \textcolor{pole}{$\blacksquare$}~\texttt{pole} (ID: $5$): Vertical cylindrical objects, such as lamp posts or utility poles, used for lighting, signage, or power distribution.
    
    \item \textcolor{traffic-light}{$\blacksquare$}~\texttt{traffic-light} (ID: $6$): Signal devices positioned at road intersections to manage traffic flow and ensure the safety of traffic participants.
    
    \item \textcolor{traffic-sign}{$\blacksquare$}~\texttt{traffic-sign} (ID: $7$): Informational or regulatory signs placed along roads to guide and control traffic behavior.
    
    \item \textcolor{vegetation}{$\blacksquare$}~\texttt{vegetation} (ID: $8$): Plant life, including trees, shrubs, and grass, typically forming natural surroundings in outdoor environments.
    
    \item \textcolor{terrain}{$\blacksquare$}~\texttt{terrain} (ID: $9$): Unpaved ground surfaces such as dirt paths, grassy fields, or rocky areas.
    
    \item \textcolor{background}{$\blacksquare$}~\texttt{sky} (ID: $10$): The open expanse above the ground, often capturing atmospheric and weather conditions.
    
    \item \textcolor{person}{$\blacksquare$}~\texttt{person} (ID: $11$): Human individuals present in the scene, either stationary or in motion.
    
    \item \textcolor{rider}{$\blacksquare$}~\texttt{rider} (ID: $12$): Individuals on moving devices such as bicycles, motorcycles, or scooters, distinct from pedestrians.
    
    \item \textcolor{car}{$\blacksquare$}~\texttt{car} (ID: $13$): Small to medium-sized motorized vehicles used for personal or commercial transport.
    
    \item \textcolor{truck}{$\blacksquare$}~\texttt{truck} (ID: $14$): Larger motorized vehicles designed for transporting goods or heavy materials.
    
    \item \textcolor{bus}{$\blacksquare$}~\texttt{bus} (ID: $15$): Large motorized vehicles used for mass public transportation of passengers.
    
    \item \textcolor{train}{$\blacksquare$}~\texttt{train} (ID: $16$): Rail-based vehicles, including locomotives and wagons, used for transporting passengers or freight.
    
    \item \textcolor{motorcycle}{$\blacksquare$}~\texttt{motorcycle} (ID: $17$): Two-wheeled motorized vehicles, often used for individual transport or recreation.
    
    \item \textcolor{bicycle}{$\blacksquare$}~\texttt{bicycle} (ID: $18$): Non-motorized two-wheeled vehicles powered by pedaling, used for transport or leisure.
\end{itemize}

Our benchmark supports two versions of label mappings, \ie, the \textbf{19-class setting} and the \textbf{11-class setting}, where the latter is consistent with the seminar event-based semantic segmentation work ESS \cite{sun2022ess}. \cref{tab:label_mapping} summarizes the relationship between these two label mappings. In our benchmark experiments, we adopt the 11-class setting for comparing different adaptation methods across platforms.

\begin{table}[t]
    \centering
    \caption{\textbf{The definitions of the semantic classes} in the \textbf{\textit{\textcolor{fly_green}{E}\textcolor{fly_red}{X}\textcolor{fly_blue}{Po}}} benchmark. We provide two versions of label mappings, \ie, the \textbf{19-class} setting and the \textbf{11-class} setting, to ensure a holistic dense perception of the scenes acquired by the event camera.}
    \vspace{-0.2cm}
    \resizebox{\linewidth}{!}{
    \begin{tabular}{c|l|c|l}
        \toprule
        \multicolumn{2}{c|}{\textbf{19-Class}} & \multicolumn{2}{c}{\textbf{11-Class}}
        \\\midrule
        \textbf{ID} & \textbf{Class Name} & \textbf{ID} & \textbf{Class Name}
        \\\midrule\midrule
        $0$ & \textcolor{road}{$\blacksquare$}~\texttt{road} & $5$ & \textcolor{road}{$\blacksquare$}~\texttt{road}
        \\\midrule
        $1$ & \textcolor{sidewalk}{$\blacksquare$}~\texttt{sidewalk} & $6$ & \textcolor{sidewalk}{$\blacksquare$}~\texttt{sidewalk}
        \\\midrule
        $2$ & \textcolor{building}{$\blacksquare$}~\texttt{building} & $1$ & \textcolor{building}{$\blacksquare$}~\texttt{building}
        \\\midrule
        $3$ & \textcolor{wall}{$\blacksquare$}~\texttt{wall} & $9$ & \textcolor{wall}{$\blacksquare$}~\texttt{wall}
        \\\midrule
        $4$ & \textcolor{fence}{$\blacksquare$}~\texttt{fence} & $2$ & \textcolor{fence}{$\blacksquare$}~\texttt{fence}
        \\\midrule
        $5$ & \textcolor{pole}{$\blacksquare$}~\texttt{pole} & $4$ & \textcolor{pole}{$\blacksquare$}~\texttt{pole}
        \\\midrule
        $6$ & \textcolor{traffic-light}{$\blacksquare$}~\texttt{traffic-light}~ & \multirow{2}{*}{$10$} & \multirow{2}{*}{\textcolor{traffic-sign}{$\blacksquare$}~\texttt{traffic-sign}~}
        \\
        $7$ & \textcolor{traffic-sign}{$\blacksquare$}~\texttt{traffic-sign}~ & 
        \\\midrule
        $8$ & \textcolor{vegetation}{$\blacksquare$}~\texttt{vegetation} & $7$ & \textcolor{vegetation}{$\blacksquare$}~\texttt{vegetation}
        \\\midrule
        $9$ & \textcolor{terrain}{$\blacksquare$}~\texttt{terrain} & \multirow{2}{*}{$0$} & \multirow{2}{*}{\textcolor{background}{$\blacksquare$}~\texttt{background}}
        \\
        $10$ & \textcolor{background}{$\blacksquare$}~\texttt{sky} & 
        \\\midrule
        $11$ & \textcolor{person}{$\blacksquare$}~\texttt{person} & \multirow{2}{*}{$3$} & \multirow{2}{*}{\textcolor{person}{$\blacksquare$}~\texttt{person}}
        \\
        $12$ & \textcolor{rider}{$\blacksquare$}~\texttt{rider} & 
        \\\midrule
        $13$ & \textcolor{car}{$\blacksquare$}~\texttt{car} & \multirow{6}{*}{$8$} & \multirow{6}{*}{\textcolor{car}{$\blacksquare$}~\texttt{car}}
        \\
        $14$ & \textcolor{truck}{$\blacksquare$}~\texttt{truck} & 
        \\
        $15$ & \textcolor{bus}{$\blacksquare$}~\texttt{bus} & 
        \\
        $16$ & \textcolor{train}{$\blacksquare$}~\texttt{train} & 
        \\
        $17$ & \textcolor{motorcycle}{$\blacksquare$}~\texttt{motorcycle} & 
        \\
        $18$ & \textcolor{bicycle}{$\blacksquare$}~\texttt{bicycle} &
        \\\bottomrule
    \end{tabular}}
\label{tab:label_mapping}
\end{table}

\subsection{Platform-Specific Statistics}
\label{sec_supp:platform_statisctis}

Each of the three platforms in the \textbf{\textit{\textcolor{fly_green}{E}\textcolor{fly_red}{X}\textcolor{fly_blue}{Po}}} benchmark represents a unique collection of event camera data. To better understand the domain gaps among these platforms, we calculate the following platform-specific statistics.

\begin{table*}[t]
    \centering
    \caption{
        \textbf{The platform-specific semantic distributions} among the \includegraphics[width=0.018\linewidth]{figures/icons/vehicle.png}~vehicle ($\mathcal{P}^{\textcolor{fly_green}{\mathbf{v}}}$), \includegraphics[width=0.021\linewidth]{figures/icons/drone.png}~drone ($\mathcal{P}^{\textcolor{fly_red}{\mathbf{d}}}$), and \includegraphics[width=0.019\linewidth]{figures/icons/quadruped.png}~quadruped ($\mathcal{P}^{\textcolor{fly_blue}{\mathbf{q}}}$) platforms, respectively, in the proposed \textbf{\textit{\textcolor{fly_green}{E}\textcolor{fly_red}{X}\textcolor{fly_blue}{Po}}} benchmark. We compare the relative proportions (normalized to $1$) of each semantic class from three platforms. The distributions of \textit{vehicle}, \textit{drone}, and \textit{quadruped} are denoted by the \textcolor{fly_green}{$\bullet$~\textbf{green}}, \textcolor{fly_red}{$\bullet$~\textbf{red}}, and \textcolor{fly_blue}{$\bullet$~\textbf{blue}} colors, respectively.}
    \vspace{-0.2cm}
    \resizebox{\linewidth}{!}{
    \begin{tabular}{c|c|c|c}
    \toprule
    \texttt{road} & \texttt{sidewalk} & \texttt{building} & \texttt{wall}
    \\
    \begin{minipage}[b]{0.5\columnwidth}\centering\raisebox{-.5\height}{\includegraphics[width=\linewidth]{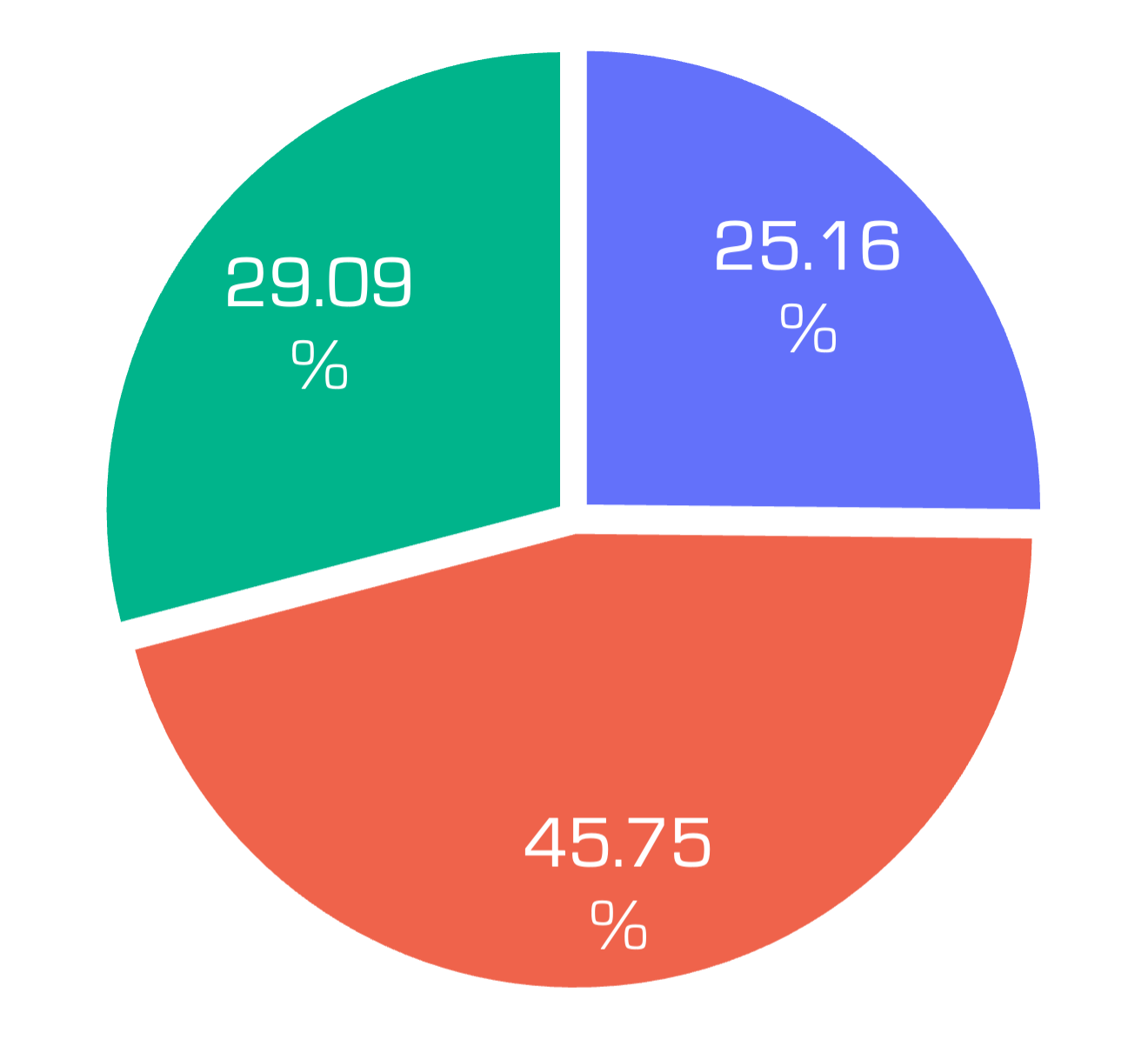}}\end{minipage} & \begin{minipage}[b]{0.5\columnwidth}\centering\raisebox{-.5\height}{\includegraphics[width=\linewidth]{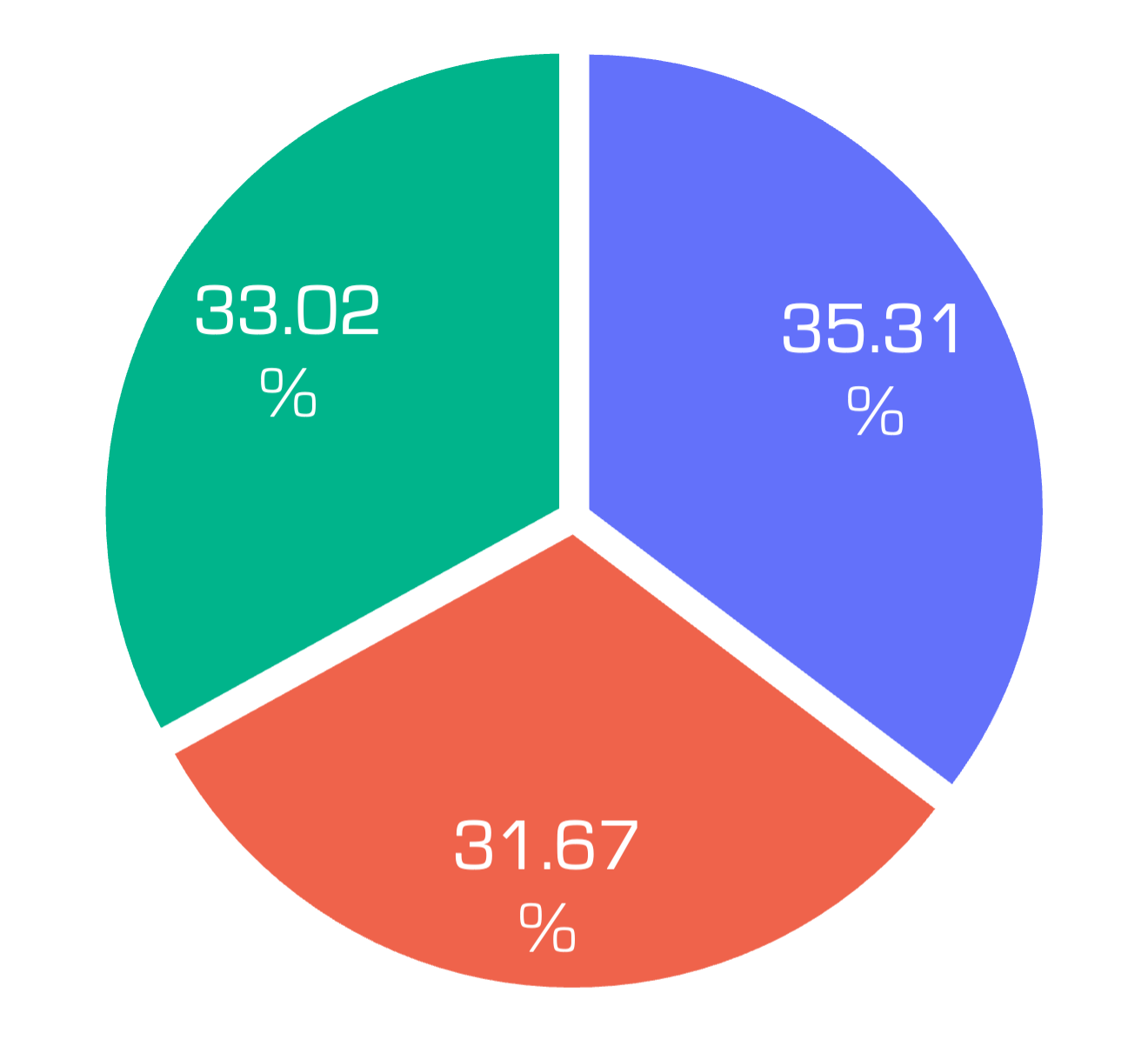}}\end{minipage} & \begin{minipage}[b]{0.5\columnwidth}\centering\raisebox{-.5\height}{\includegraphics[width=\linewidth]{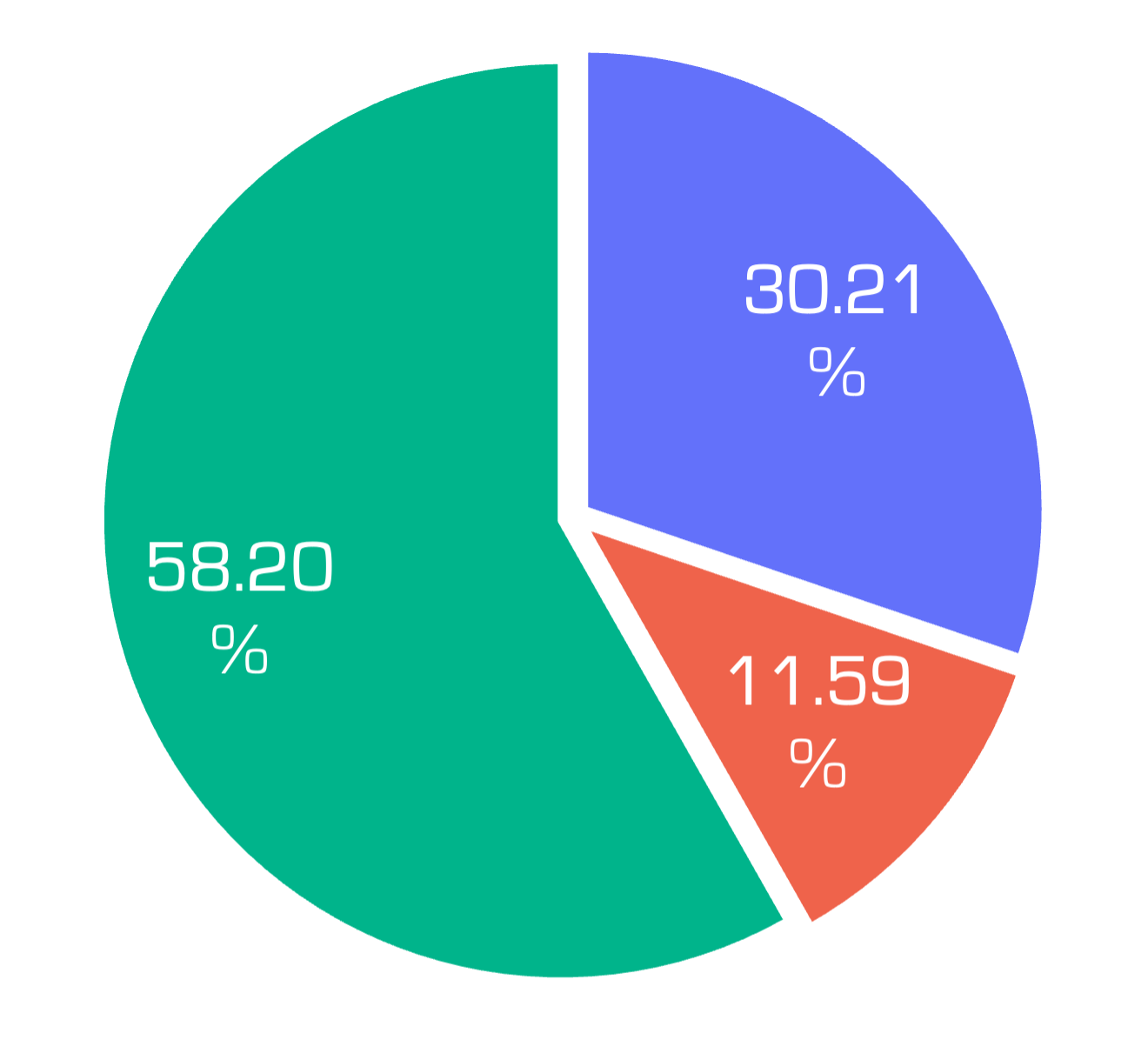}}\end{minipage} & \begin{minipage}[b]{0.5\columnwidth}\centering\raisebox{-.5\height}{\includegraphics[width=\linewidth]{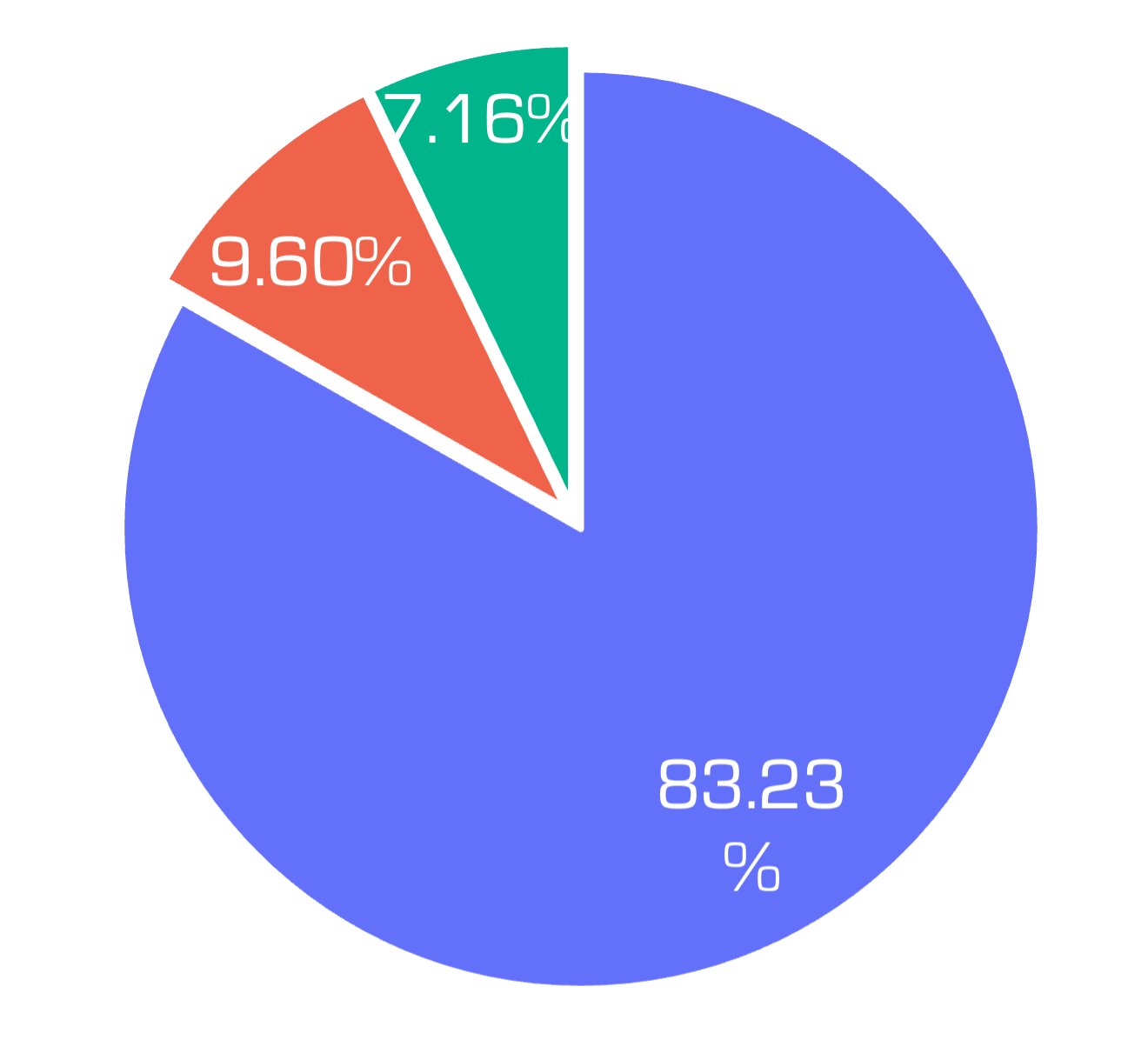}}\end{minipage}
    \\\midrule
    \texttt{fence} & \texttt{pole} & \texttt{traffic-light} & \texttt{traffic-sign}
    \\
    \begin{minipage}[b]{0.5\columnwidth}\centering\raisebox{-.5\height}{\includegraphics[width=\linewidth]{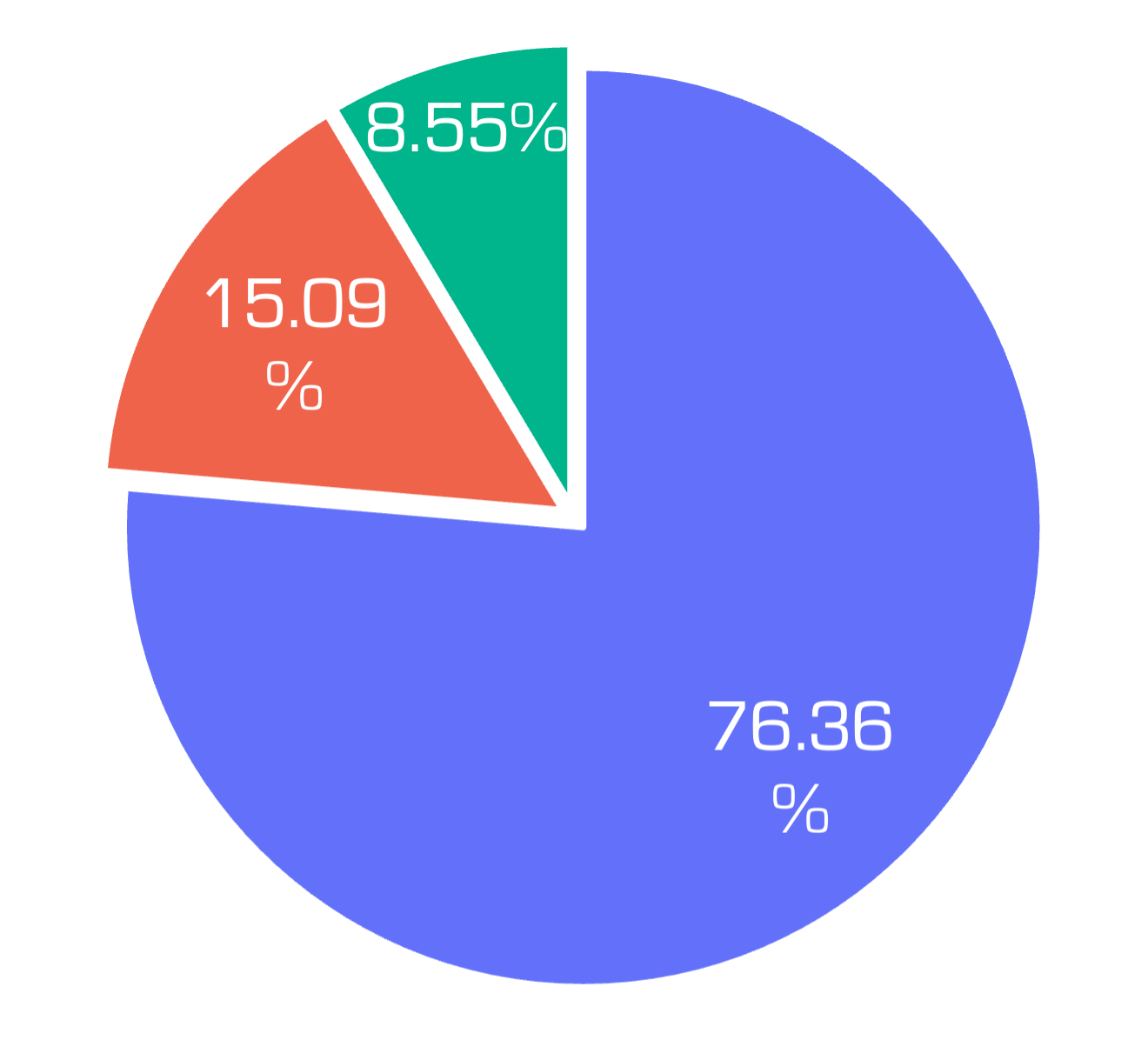}}\end{minipage} & \begin{minipage}[b]{0.5\columnwidth}\centering\raisebox{-.5\height}{\includegraphics[width=\linewidth]{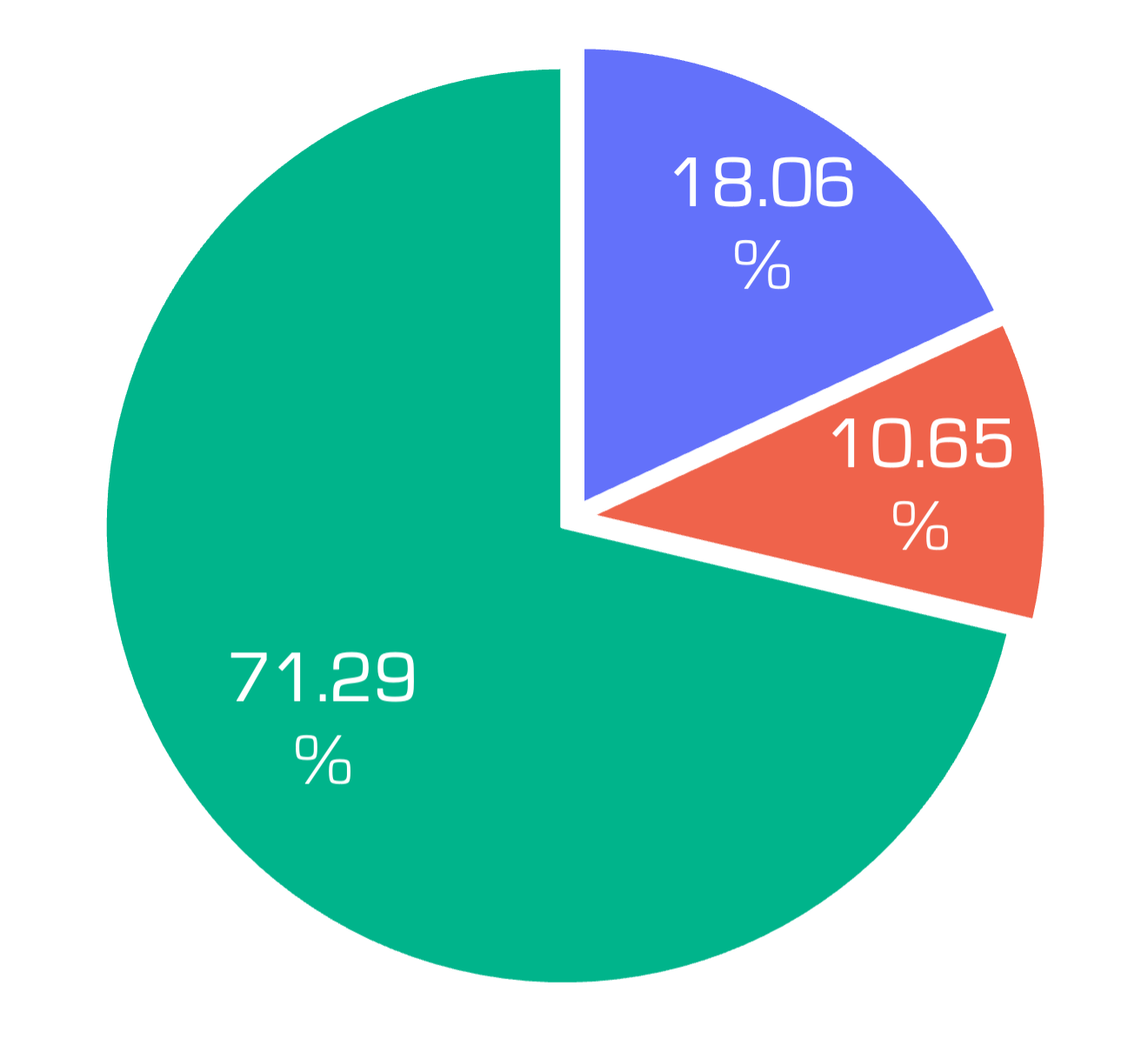}}\end{minipage} & \begin{minipage}[b]{0.5\columnwidth}\centering\raisebox{-.5\height}{\includegraphics[width=\linewidth]{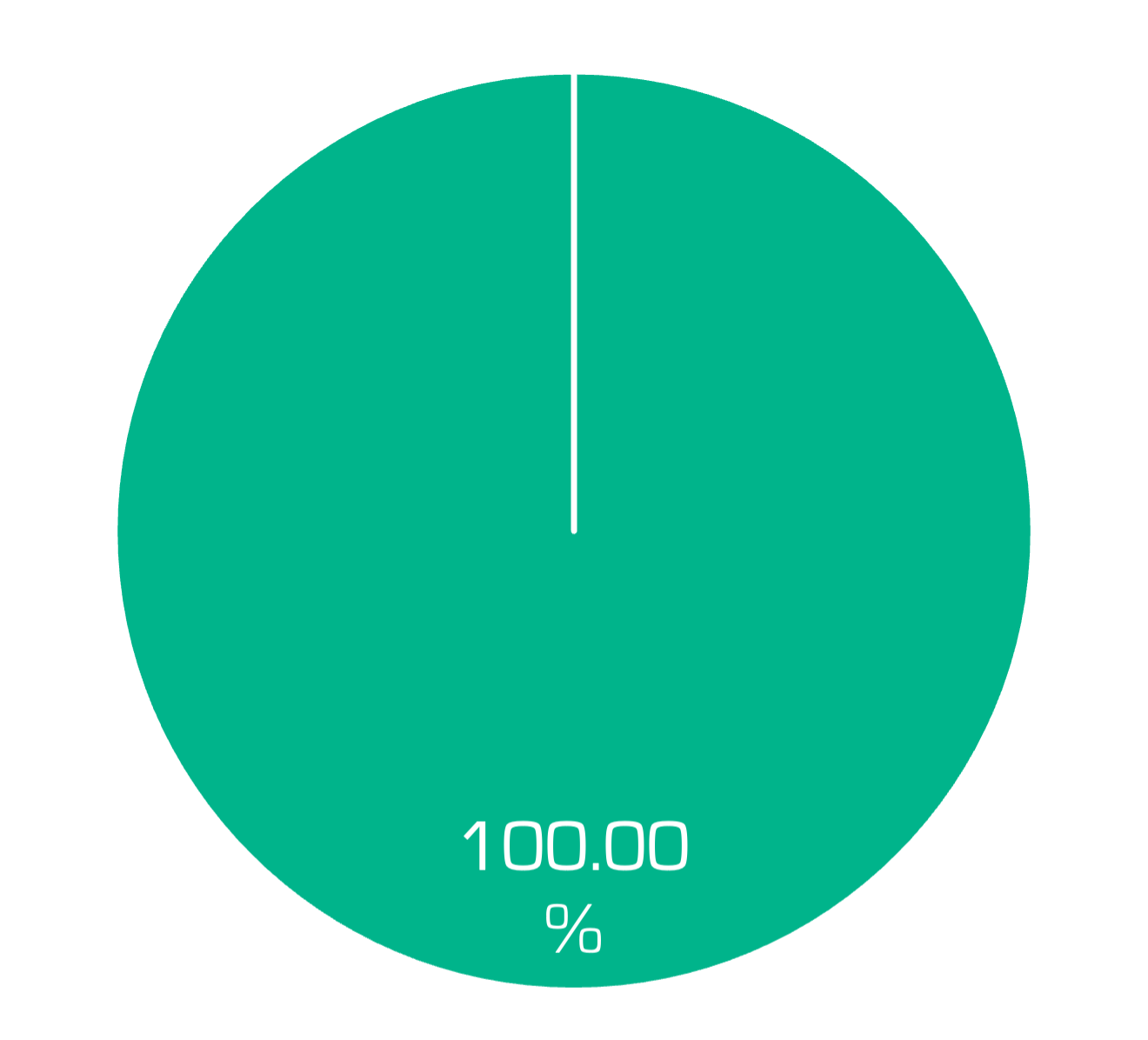}}\end{minipage} & \begin{minipage}[b]{0.5\columnwidth}\centering\raisebox{-.5\height}{\includegraphics[width=\linewidth]{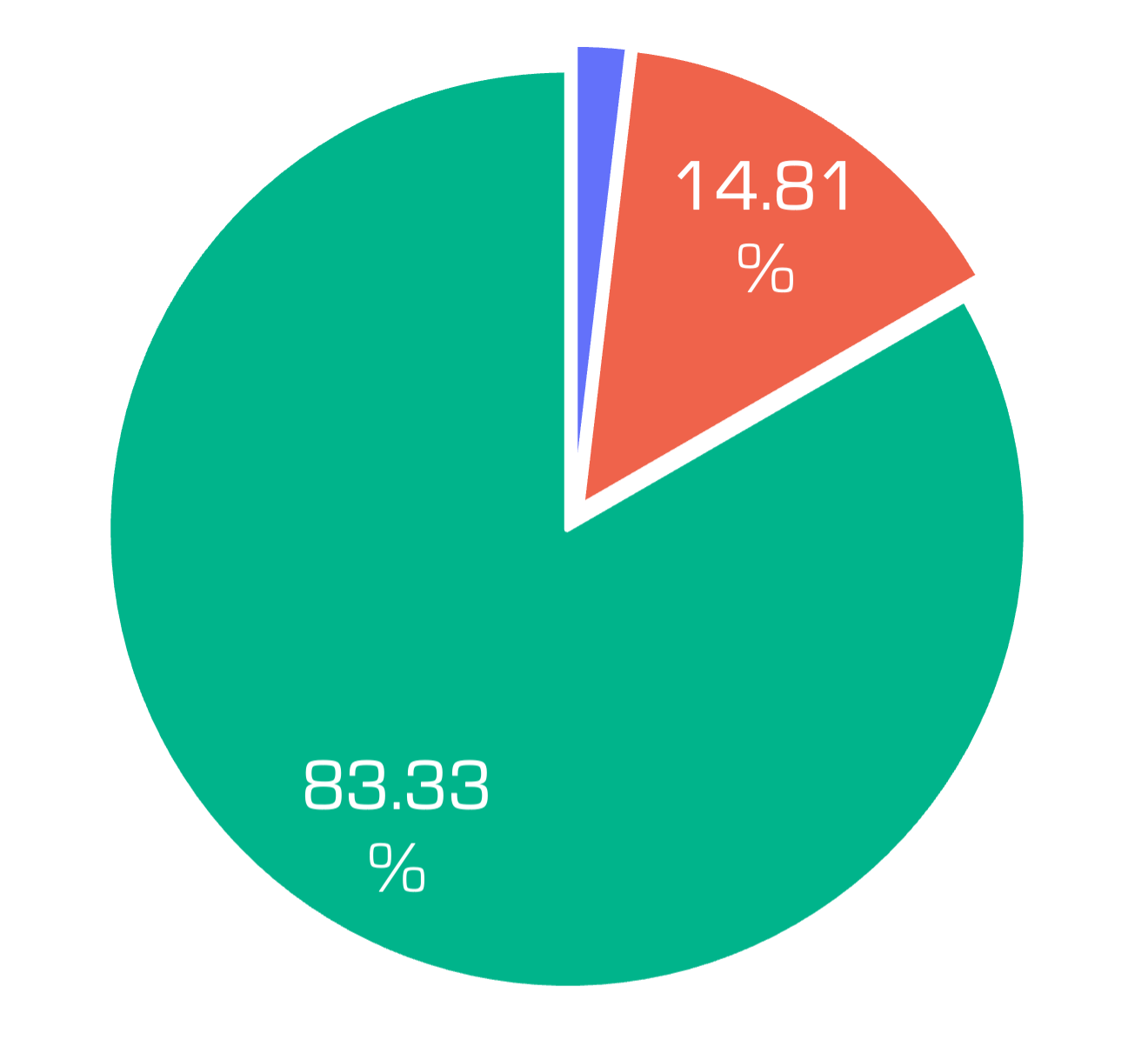}}\end{minipage}
    \\\midrule
    \texttt{vegetation} & \texttt{terrain} & \texttt{sky} & \texttt{person}
    \\
    \begin{minipage}[b]{0.5\columnwidth}\centering\raisebox{-.5\height}{\includegraphics[width=\linewidth]{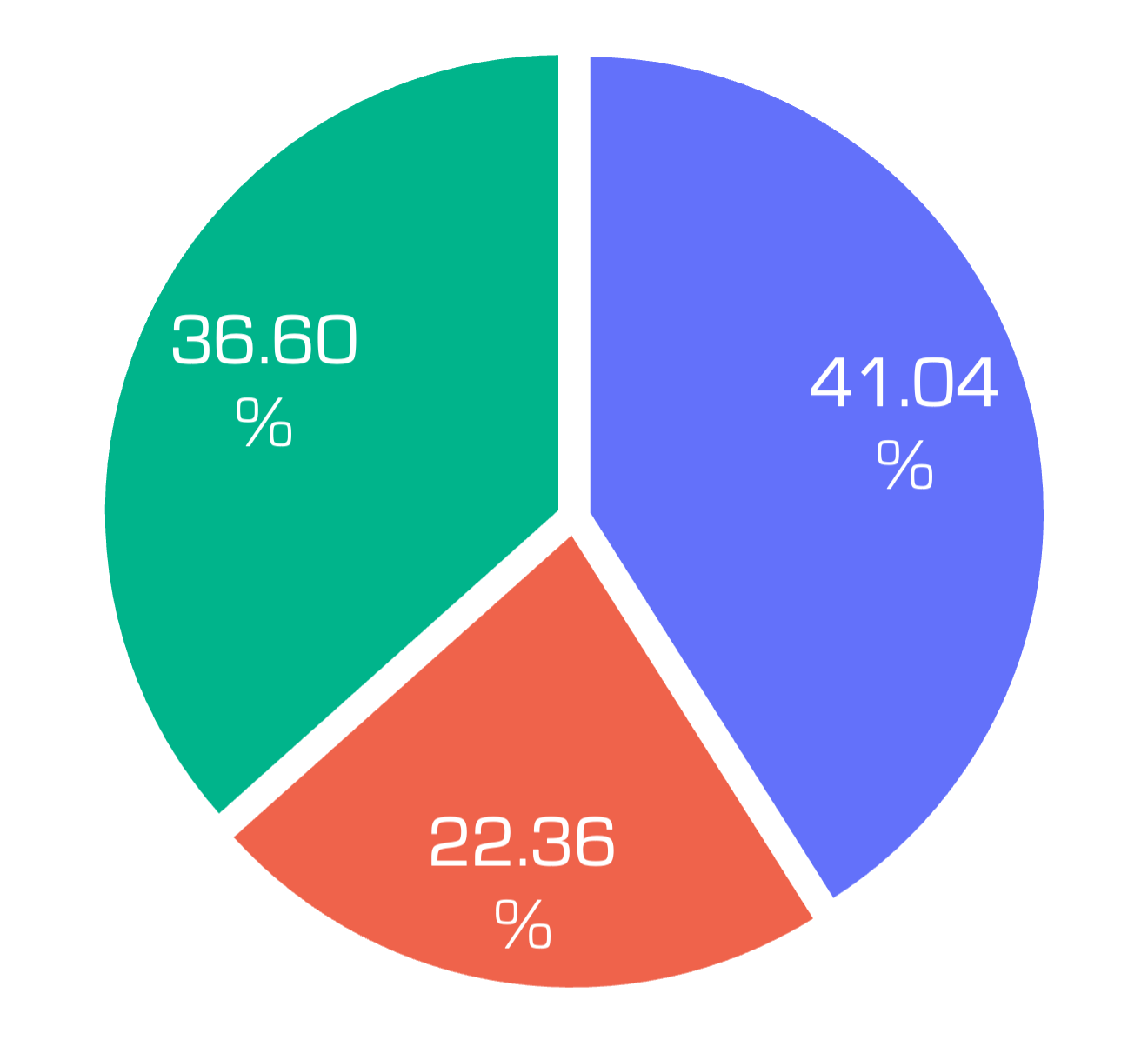}}\end{minipage} & \begin{minipage}[b]{0.5\columnwidth}\centering\raisebox{-.5\height}{\includegraphics[width=\linewidth]{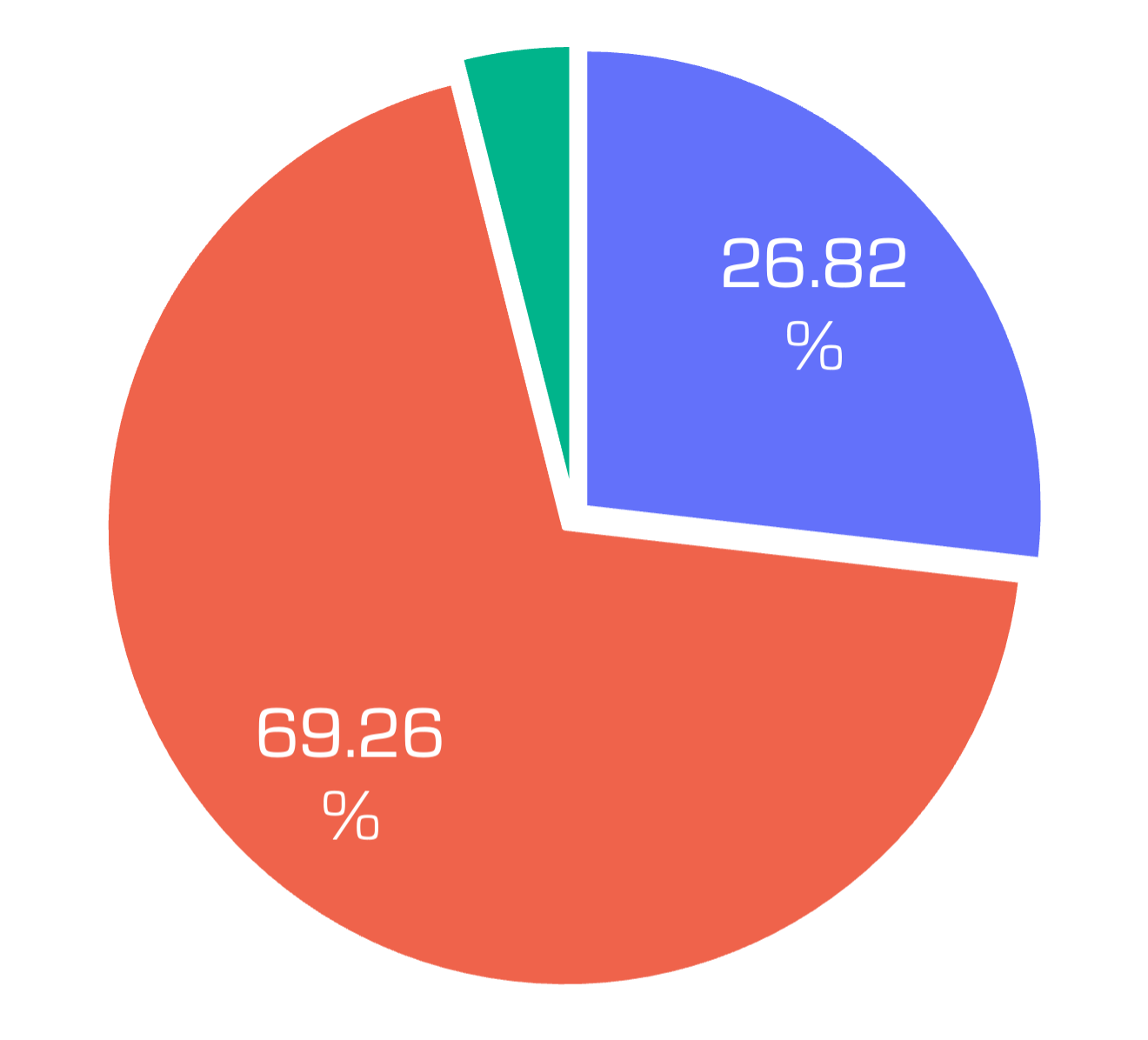}}\end{minipage} & \begin{minipage}[b]{0.5\columnwidth}\centering\raisebox{-.5\height}{\includegraphics[width=\linewidth]{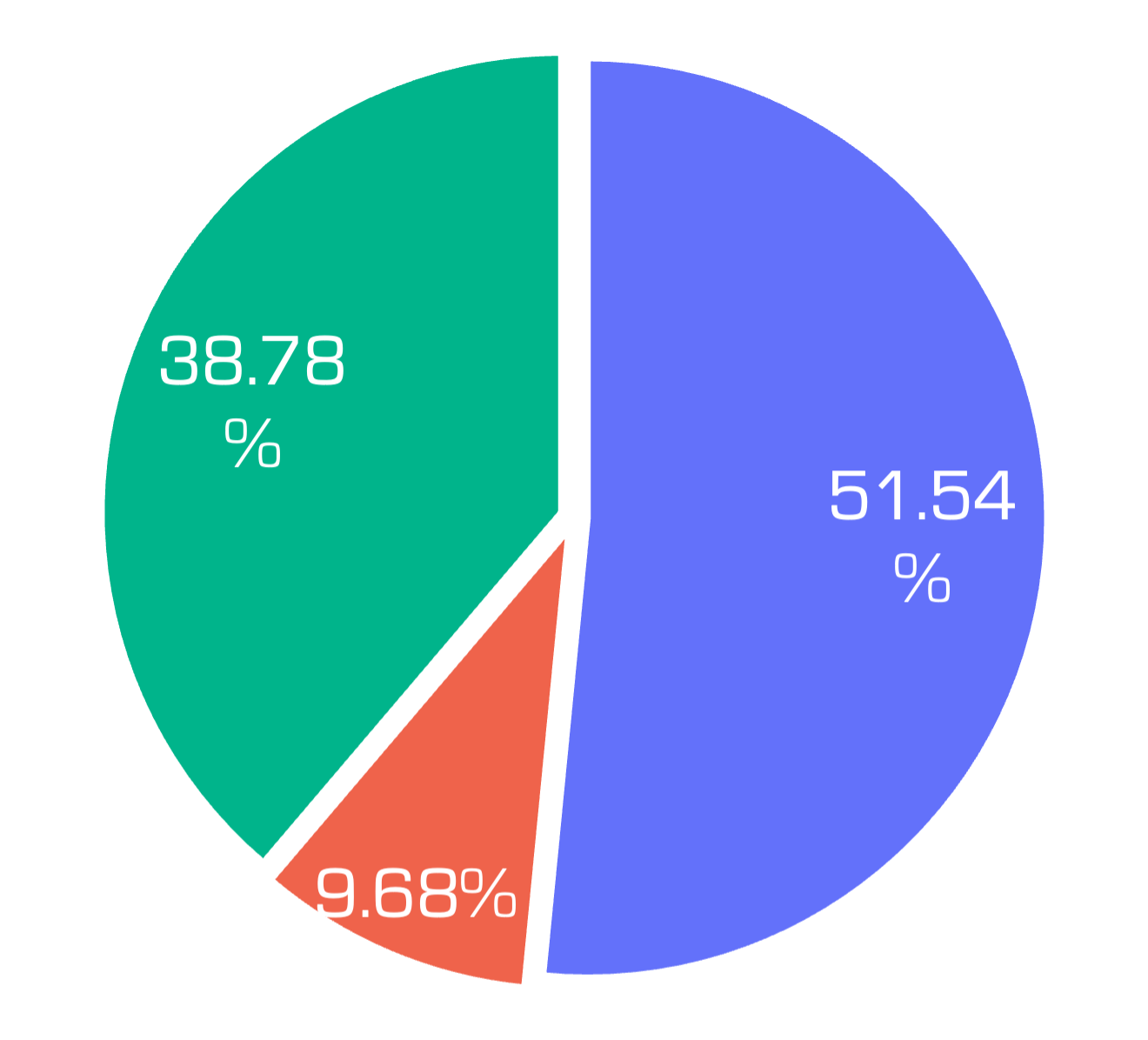}}\end{minipage} & \begin{minipage}[b]{0.5\columnwidth}\centering\raisebox{-.5\height}{\includegraphics[width=\linewidth]{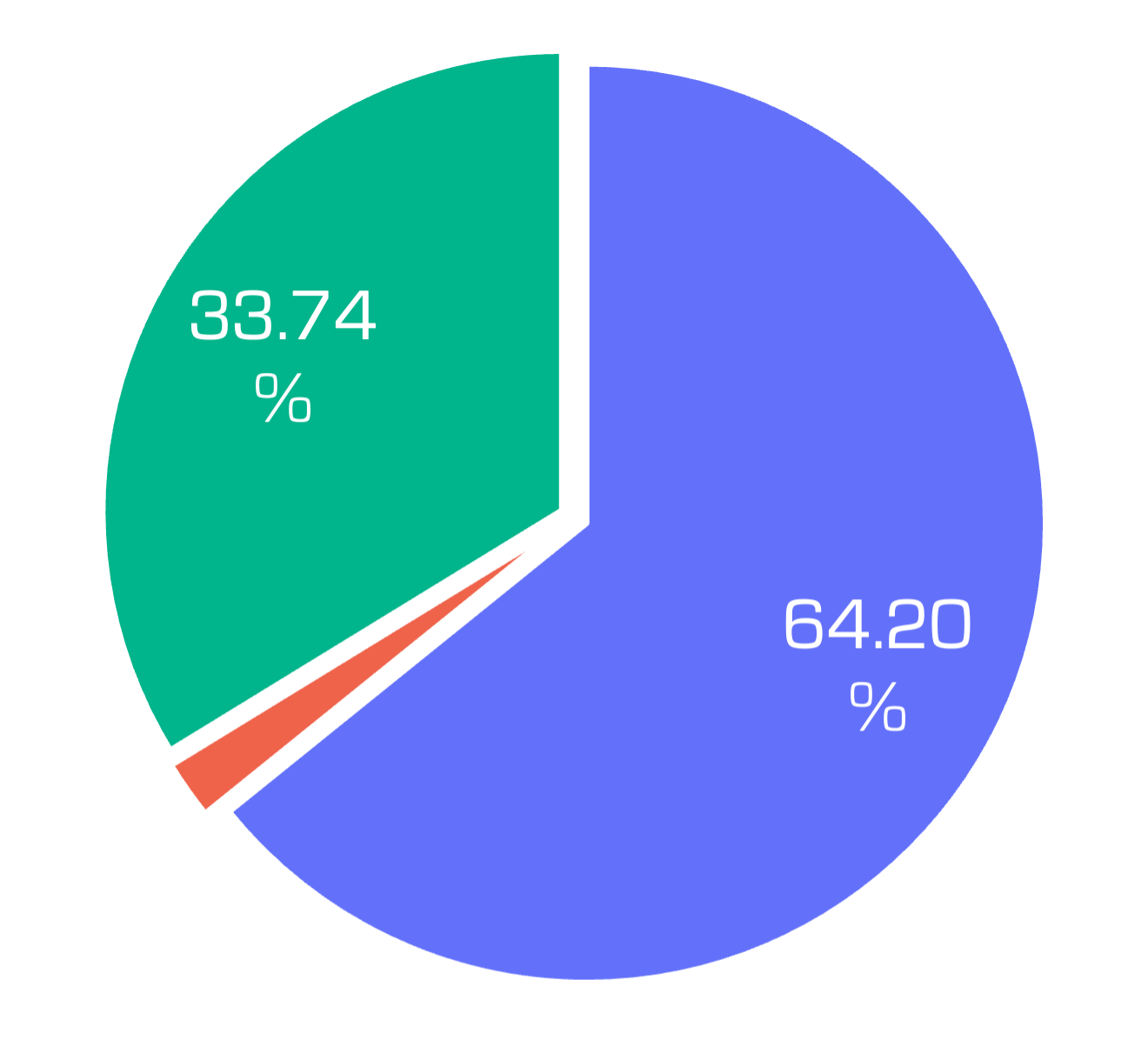}}\end{minipage}
    \\\midrule
    \texttt{rider} & \texttt{car} & \texttt{truck} & \texttt{bus}
    \\
    \begin{minipage}[b]{0.5\columnwidth}\centering\raisebox{-.5\height}{\includegraphics[width=\linewidth]{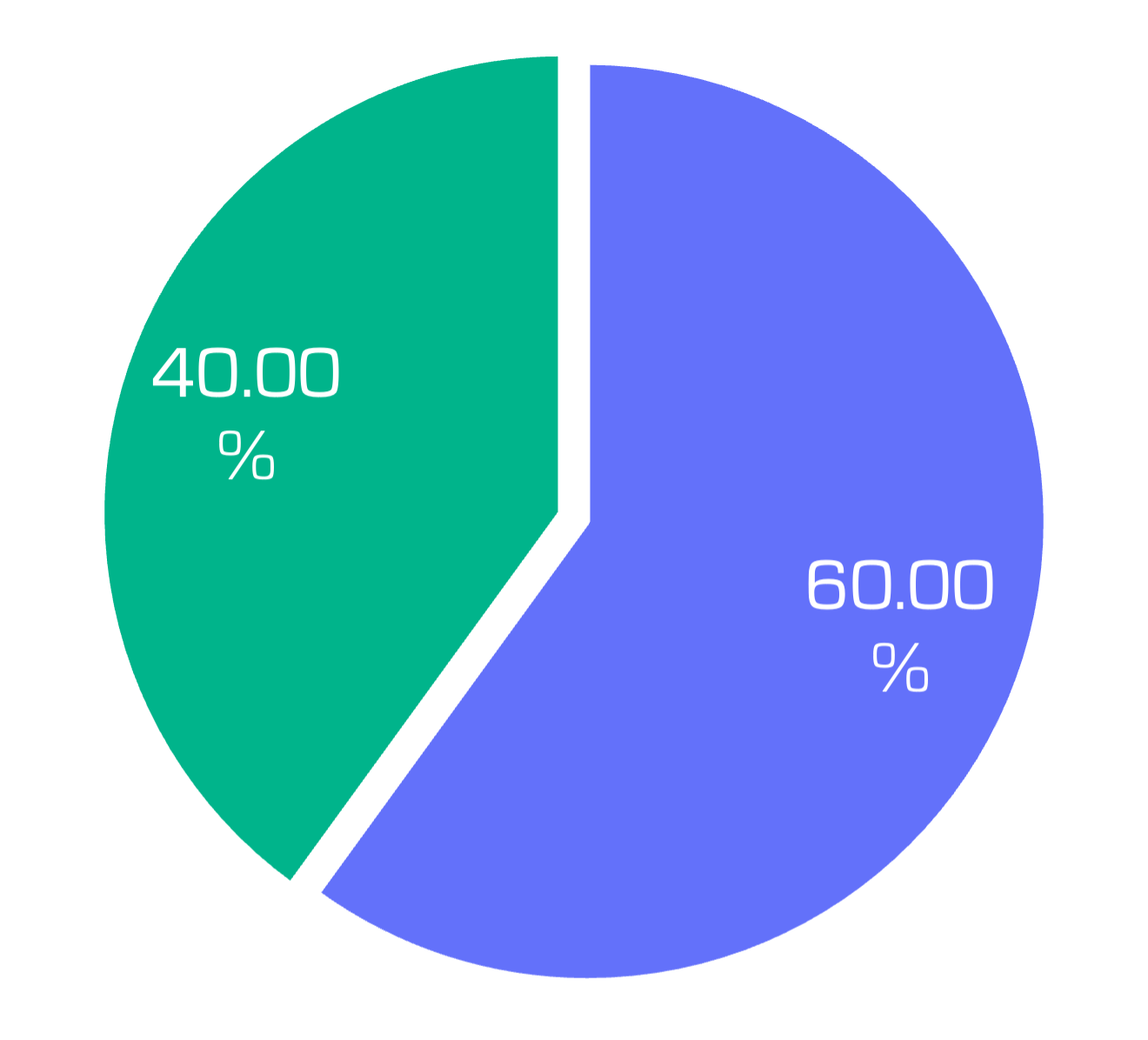}}\end{minipage} & \begin{minipage}[b]{0.5\columnwidth}\centering\raisebox{-.5\height}{\includegraphics[width=\linewidth]{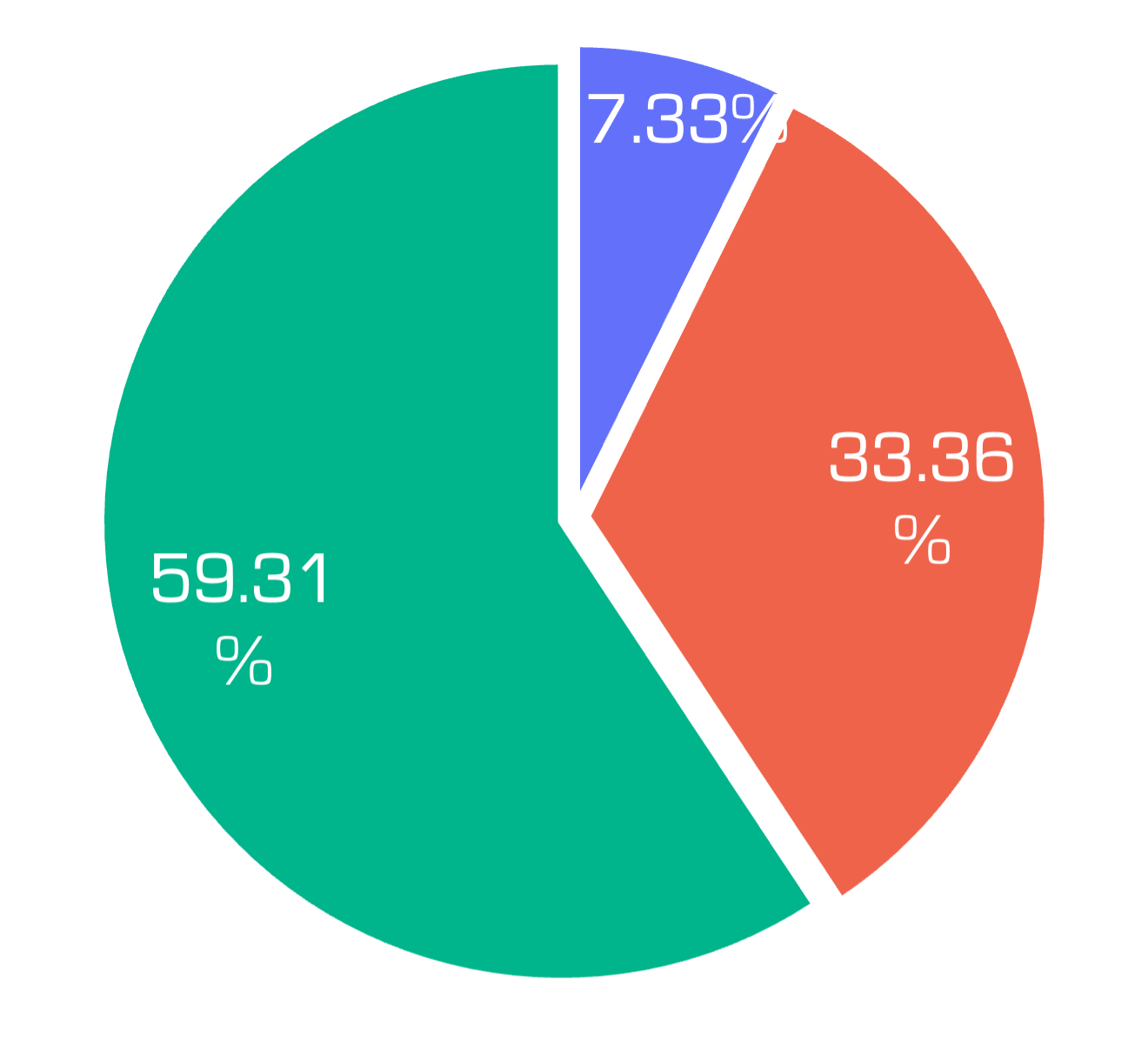}}\end{minipage} & \begin{minipage}[b]{0.5\columnwidth}\centering\raisebox{-.5\height}{\includegraphics[width=\linewidth]{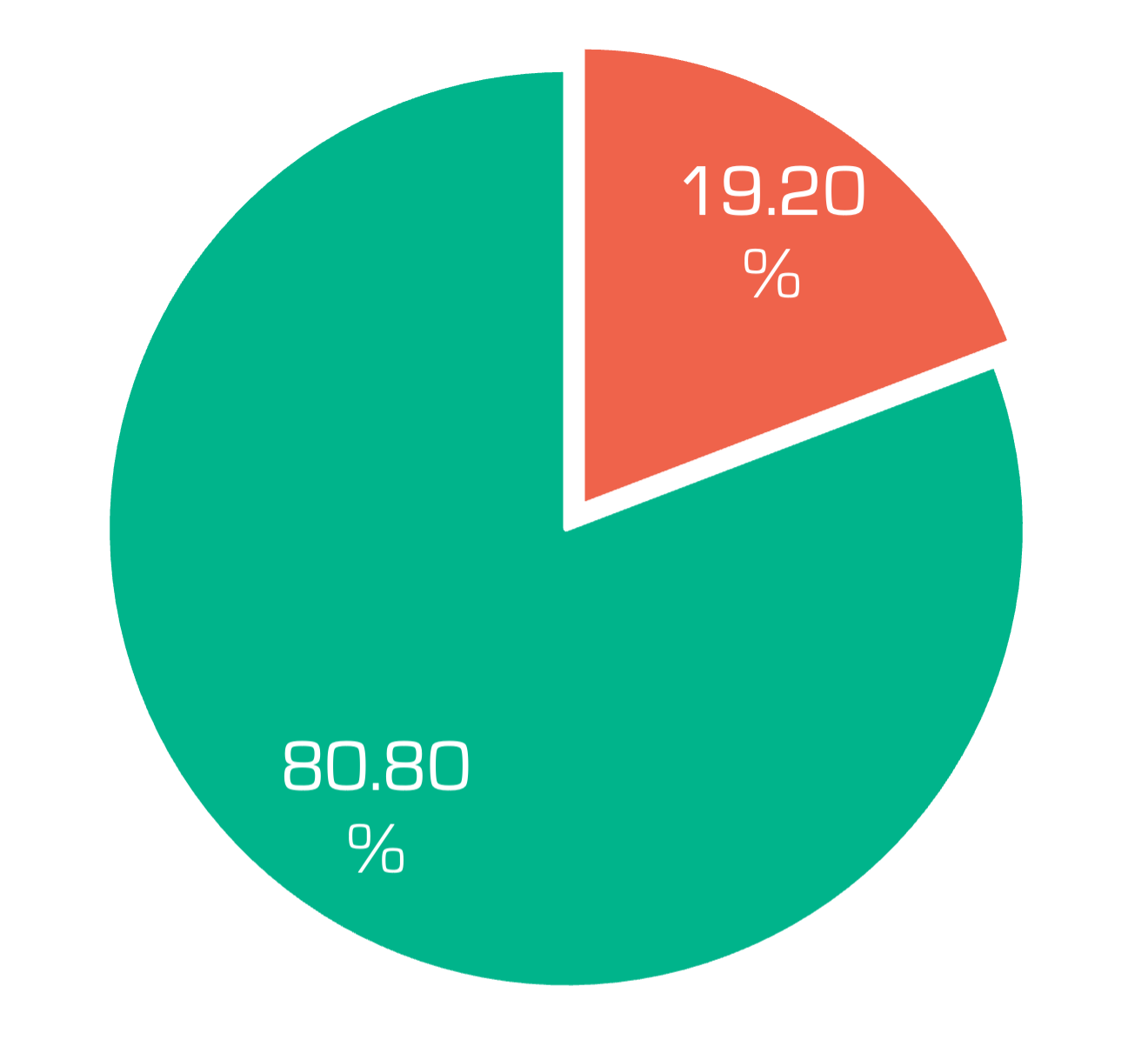}}\end{minipage} & \begin{minipage}[b]{0.5\columnwidth}\centering\raisebox{-.5\height}{\includegraphics[width=\linewidth]{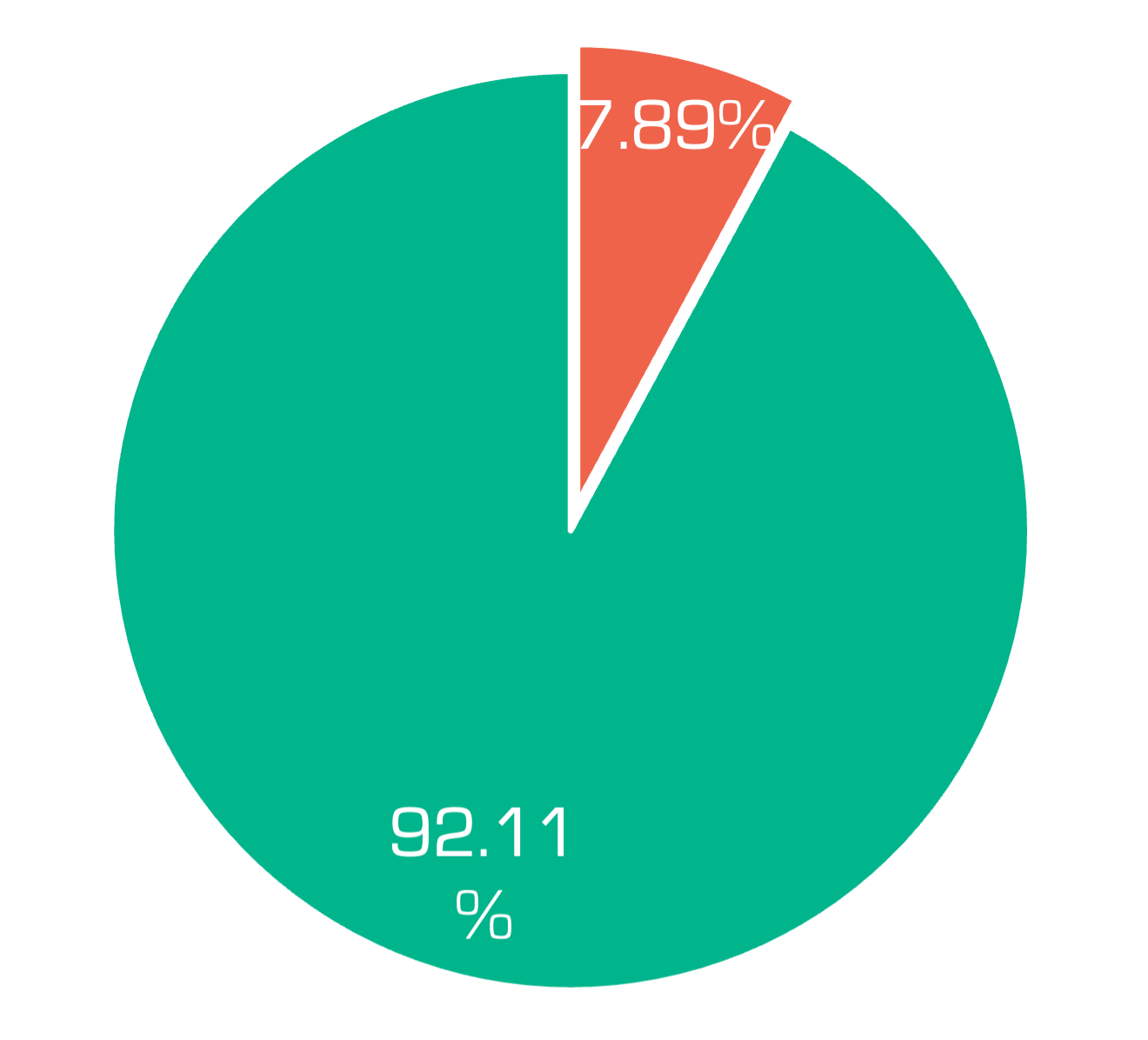}}\end{minipage}
    \\\midrule
    \texttt{train} & \texttt{motorcycle} & \texttt{bicycle} & 
    \\
    \begin{minipage}[b]{0.5\columnwidth}\centering\raisebox{-.5\height}{\includegraphics[width=\linewidth]{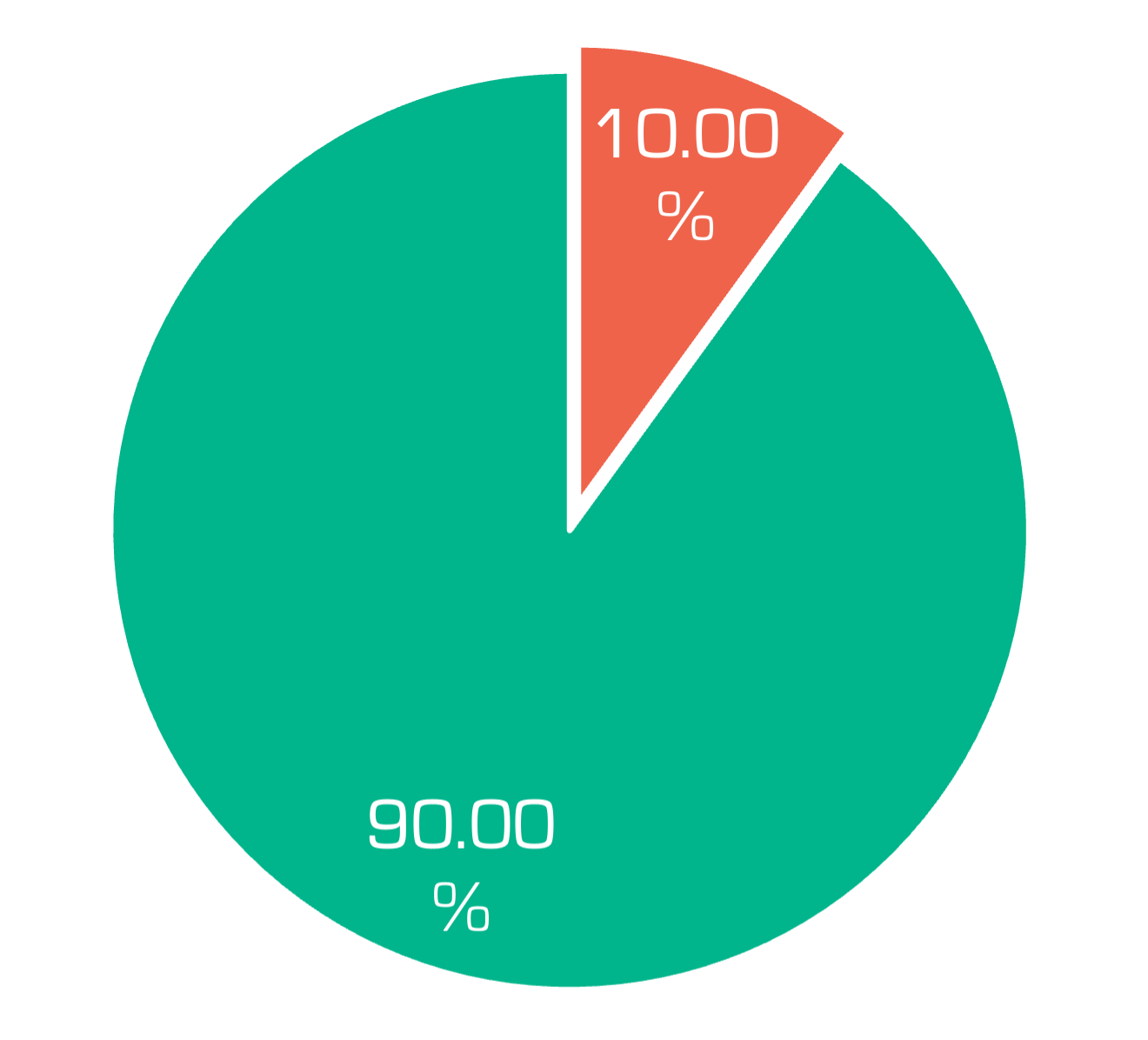}}\end{minipage} & \begin{minipage}[b]{0.5\columnwidth}\centering\raisebox{-.5\height}{\includegraphics[width=\linewidth]{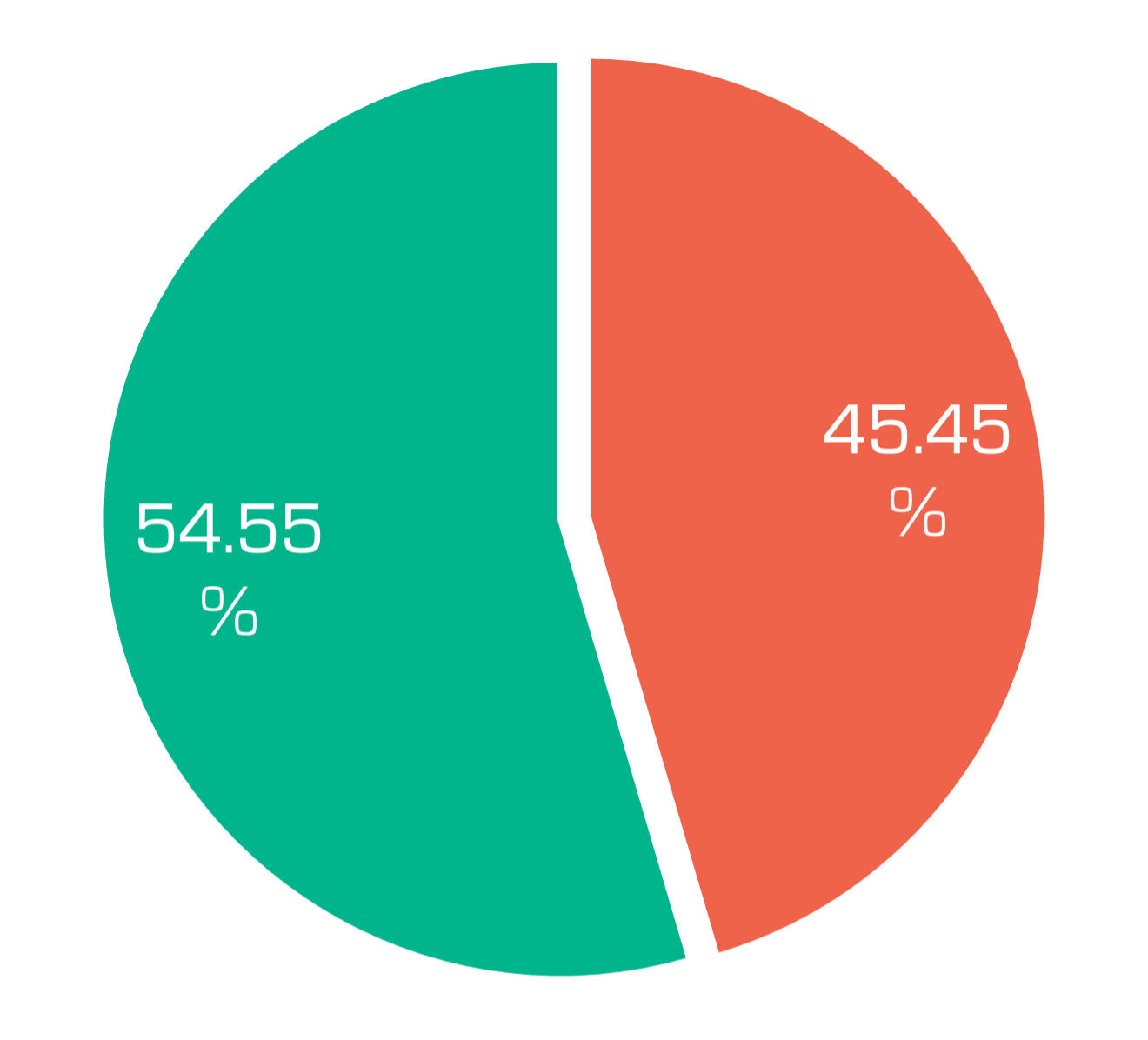}}\end{minipage} & \begin{minipage}[b]{0.5\columnwidth}\centering\raisebox{-.5\height}{\includegraphics[width=\linewidth]{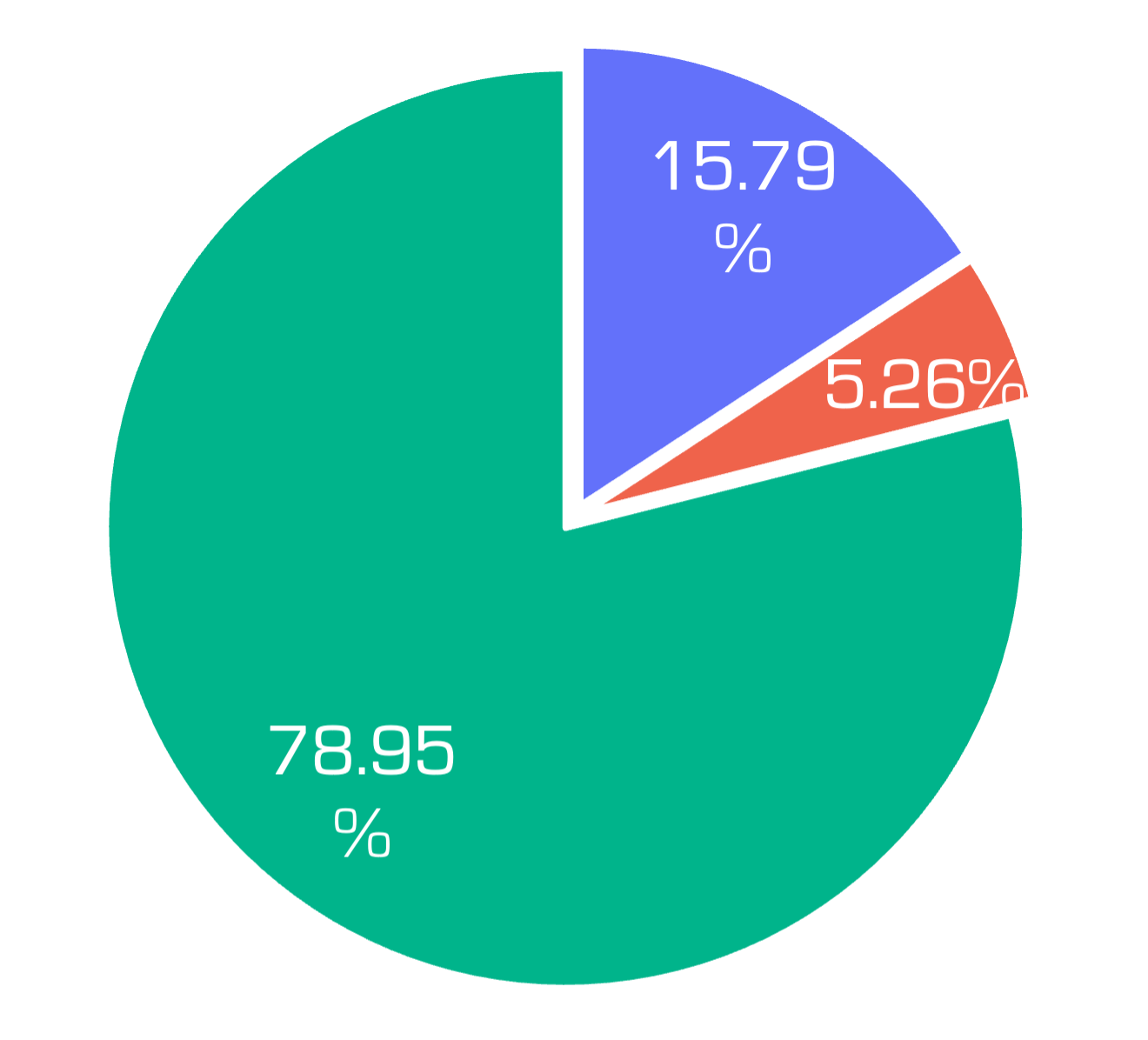}}\end{minipage} &
    \\\bottomrule
\end{tabular}}
\label{tab:pie_charts}
\vspace{-0.4cm}
\end{table*}

\begin{table}[t]
    \centering
    \caption{\textbf{The absolute platform-specific semantic distributions} among the \includegraphics[width=0.036\linewidth]{figures/icons/vehicle.png}~vehicle ($\mathcal{P}^{\textcolor{fly_green}{\mathbf{v}}}$), \includegraphics[width=0.042\linewidth]{figures/icons/drone.png}~drone ($\mathcal{P}^{\textcolor{fly_red}{\mathbf{d}}}$), and \includegraphics[width=0.038\linewidth]{figures/icons/quadruped.png}~quadruped ($\mathcal{P}^{\textcolor{fly_blue}{\mathbf{q}}}$) platforms, respectively, in the proposed \textbf{\textit{\textcolor{fly_green}{E}\textcolor{fly_red}{X}\textcolor{fly_blue}{Po}}} benchmark.}
    \vspace{-0.2cm}
    \resizebox{\linewidth}{!}{
    \begin{tabular}{l|p{47pt}<{\centering}|p{47pt}<{\centering}|p{47pt}<{\centering}}
        \toprule
        \textbf{Class} & \textbf{Vehicle} & \textbf{Drone} & \textbf{Quadruped}
        \\\midrule\midrule
        \textcolor{road}{$\blacksquare$}~\texttt{road} & $21.94\%$ & $34.51\%$ & $18.98\%$
        \\
        \textcolor{sidewalk}{$\blacksquare$}~\texttt{sidewalk} & $6.63\%$ & $6.36\%$ & $7.09\%$
        \\
        \textcolor{building}{$\blacksquare$}~\texttt{building} & $24.91\%$ & $4.96\%$ & $12.93\%$
        \\
        \textcolor{wall}{$\blacksquare$}~\texttt{wall} & $0.47\%$ & $0.63\%$ & $5.46\%$ 
        \\
        \textcolor{fence}{$\blacksquare$}~\texttt{fence} & $0.55\%$ & $0.97\%$ & $4.91\%$
        \\
        \textcolor{pole}{$\blacksquare$}~\texttt{pole} & $2.21\%$ & $0.33\%$ & $0.56\%$
        \\
        \textcolor{traffic-light}{$\blacksquare$}~\texttt{traffic-light} & $0.22\%$ & $0.00\%$ & $0.00\%$
        \\
        \textcolor{traffic-sign}{$\blacksquare$}~\texttt{traffic-sign} & $0.45\%$ & $0.08\%$ & $0.01\%$
        \\
        \textcolor{vegetation}{$\blacksquare$}~\texttt{vegetation} & $23.77\%$ & $14.52\%$ & $26.65\%$
        \\
        \textcolor{terrain}{$\blacksquare$}~\texttt{terrain} & $1.78\%$ & $31.46\%$ & $12.18\%$
        \\
        \textcolor{background}{$\blacksquare$}~\texttt{sky} & $6.53\%$ & $1.63\%$ & $8.68\%$
        \\
        \textcolor{person}{$\blacksquare$}~\texttt{person} & $0.82\%$ & $0.05\%$ & $1.56\%$
        \\
        \textcolor{rider}{$\blacksquare$}~\texttt{rider} & $0.02\%$ & $0.00\%$ & $0.03\%$
        \\
        \textcolor{car}{$\blacksquare$}~\texttt{car} & $7.36\%$ & $4.14\%$ & $0.91\%$
        \\
        \textcolor{truck}{$\blacksquare$}~\texttt{truck} & $1.01\%$ & $0.24\%$ & $0.00\%$
        \\
        \textcolor{bus}{$\blacksquare$}~\texttt{bus} & $1.05\%$ & $0.09\%$ & $0.00\%$
        \\
        \textcolor{train}{$\blacksquare$}~\texttt{train} & $0.09\%$ & $0.01\%$ & $0.00\%$
        \\
        \textcolor{motorcycle}{$\blacksquare$}~\texttt{motorcycle} & $0.01\%$ & $0.01\%$ & $0.00\%$
        \\
        \textcolor{bicycle}{$\blacksquare$}~\texttt{bicycle} & $0.15\%$ & $0.01\%$ & $0.03\%$
        \\\midrule
        \textbf{Total} & \textcolor{fly_green}{$\mathbf{100\%}$} & \textcolor{fly_red}{$\mathbf{100\%}$} & \textcolor{fly_blue}{$\mathbf{100\%}$}
        \\\bottomrule
    \end{tabular}}
\label{tab:class_distribution}
\end{table}

\begin{itemize}
    \item \textbf{Platform-Specific Semantic Distributions:} The relative proportions of each semantic class across the three platforms are presented in \cref{tab:pie_charts}, with semantic occupations normalized to $1$. Notable discrepancies are observed among the platforms. 
    
    \begin{itemize}
        \item For instance, the \textit{drone} platform accounts for $45.75\%$ of the \texttt{road} class, attributed to its high-altitude perspective that captures expansive ground surfaces. In contrast, the \textit{vehicle} platform dominates classes such as \texttt{building}, \texttt{traffic-sign}, and all categories of \texttt{car}, reflecting its road-level viewpoint and focus on urban navigation. Similarly, all instances of \texttt{traffic-light} appear exclusively in the \textit{vehicle} platform, as this class is inherently associated with vehicle-centric scenarios.

        \item On the other hand, the \textit{quadruped} platform, with its low-height perspective, captures a higher proportion of \texttt{fence} ($76.36\%$), \texttt{wall} ($83.23\%$), and similar semantic categories. This aligns with its tendency to perceive surroundings closer to ground level, making it better suited for mixed indoor-outdoor environments. 

        \item As for the \textit{drone} platform, a significant proportion of \texttt{terrain} ($69.26\%$) is captured due to its elevated viewpoint, which provides a broader landscape perspective. This platform also includes a notable share of \texttt{car}-related classes, such as \texttt{truck} ($19.20\%$), \texttt{bus} ($7.89\%$), and \texttt{motorcycle} ($45.45\%$), reflecting its ability to observe these objects from a unique vantage point that complements ground-level perspectives.
        
        \item Each platform thus exhibits distinct semantic distributions, emphasizing the importance of tailored domain adaptation strategies for robust cross-platform event perception.
    \end{itemize}

    \item \textbf{Absolute Semantic Distributions:} We calculate the absolute semantic occupations for each platform and present the statistics in \cref{tab:class_distribution}. As shown, the distributions for all three platforms exhibit a long-tailed nature, reflecting real-world event camera scenarios where certain static classes dominate while dynamic and small-object classes occur less frequently.
    \begin{itemize}
        \item The majority classes for the \textit{vehicle} platform are \texttt{building} ($24.91\%$), \texttt{vegetation} ($23.77\%$), and \texttt{road} ($21.94\%$). These static classes dominate due to the platform’s road-level perspective, which frequently encounters large, continuous structures and roadside greenery. In contrast, small and dynamic classes, such as \texttt{rider} ($0.02\%$) and \texttt{motorcycle} ($0.01\%$), are underrepresented, underscoring the vehicle platform’s bias towards large, static objects in its operating environment.

        \item The \textit{drone} platform primarily captures \texttt{road} ($34.51\%$), \texttt{terrain} ($31.46\%$), and \texttt{vegetation} ($14.52\%$). This is due to its high-altitude perspective, which provides expansive views of ground surfaces and surrounding landscapes. Dynamic classes, such as different categories of \texttt{car}, are underrepresented because they occupy less visual space from the drone’s viewpoint compared to static, large-area features.

        \item We also observe that the \texttt{quadruped} platform exhibits notably higher proportions of \texttt{sky} ($8.68\%$), \texttt{wall} ($5.46\%$), and \texttt{fence} ($4.91\%$) compared to the other two platforms. This is attributed to its low-altitude perspective, which captures more vertical structures and surrounding boundaries, as well as frequent mixed indoor-outdoor scenarios. Unlike the \texttt{vehicle} and \texttt{drone} platforms, \texttt{quadruped} data features a more balanced representation of close-range objects and environmental details.
    \end{itemize}
\end{itemize}

These platform-specific statistics provide a comprehensive understanding of the challenges in cross-platform adaptation, emphasizing the need for robust event camera perception models capable of handling diverse semantic distributions and environmental contexts.

\subsection{License}
The \textbf{\textit{\textcolor{fly_green}{E}\textcolor{fly_red}{X}\textcolor{fly_blue}{Po}}} benchmark is released under the Attribution-ShareAlike 4.0 International (CC BY-SA 4.0)\footnote{\url{https://creativecommons.org/licenses/by-sa/4.0/legalcode}.} license.

\section{Event Activation Prior: Formulation}
\label{sec:prior_details}
In event-based cross-platform adaptation, each platform introduces unique activation patterns due to variations in sensor perspectives, motion dynamics, and environmental conditions. The Event Activation Prior (EAP) captures these platform-specific activation patterns and encourages confident predictions by leveraging the classic entropy minimization framework. In this section, we elaborate on the formulation of our proposed EAP in more detail.

\subsection{Problem Formulation}
In our setting, we address cross-platform adaptation across three distinct event data domains: \textit{vehicle}, \textit{drone}, and \textit{quadruped}, referred to as $\mathcal{D} =\{\mathcal{P}^{\textcolor{fly_green}{\mathbf{v}}}, \mathcal{P}^{\textcolor{fly_red}{\mathbf{d}}}, \mathcal{P}^{\textcolor{fly_blue}{\mathbf{q}}} \}$, respectively.
Each domain contains:
\begin{itemize}
    \item \textbf{Event Voxel Grids}: $\mathbf{V} \in \mathbb{R}^{T \times H \times W}$, where $T$ is the number of temporal bins and $(H, W)$ are the spatial dimensions of the event sensor.
    
    \item \textbf{Semantic Labels} (source domain only):  $y \in \mathbb{R}^{H \times W}$, where each pixel corresponds to one of $C$ pre-defined semantic classes.
\end{itemize}

In our cross-platform adaptation problem, we assume access to fully labeled data from a source domain while only having access to unlabeled data from a target domain. The objective is to leverage both the labeled source data and the unlabeled target data to train an event camera perception model that can perform well on the target domain. This adaptation is challenging because each platform captures data from distinct perspectives, motion patterns, and environmental contexts.

\subsection{EAP: Motivation \& Formulation}
EAP is designed to guide cross-platform adaptation by leveraging platform-specific event activation patterns. Events are triggered by changes in brightness due to motion, making certain regions in the event data -- characterized by frequent activations -- highly informative. By minimizing entropy in these regions, we hope to encourage the model to make confident predictions that align with the target domain’s unique motion-triggered patterns, which in turn improve the perception performance.

\subsection{Likelihood for Supervised Loss}
For labeled data from the source domain $ \mathcal{P}^{\mathrm{src}} \in \mathcal{D}$, we train our event camera perception model by maximizing the likelihood of the ground truth labels. This likelihood, $P(y | \mathbf{V})$, forms the supervised loss term:
\begin{equation}
    L(\theta) = -\sum_{\mathbf{V} \in \mathcal{P}^{\mathrm{src}}} \log P(y | \mathbf{V}; \theta)~,
\end{equation}
where $\theta$ represents the model parameters. This supervised loss anchors the model’s learning in well-labeled source data, providing a foundation for generalization.

Since we lack labeled data in the target domain, we define the EAP to help the model leverage unlabeled data by minimizing prediction uncertainty in \textit{high-activation regions} of the target domain. These regions, $\mathbf{S} \subset \{0, 1, \dots, H-1\} \times \{0, 1, \dots, W-1\}$, are identified based on the characteristic event activations in each platform.
To achieve this, EAP follows the principle of entropy minimization, where we aim to:
\begin{itemize}
    \item Identify high-activation regions $\mathbf{S}$ in the target domain.

    \item Minimize the conditional entropy $H(y_{\mathbf{S}} | \mathbf{V}_{\mathbf{S}}, \mathbf{S})$ in these regions, promoting confident predictions that align with target-specific patterns.
\end{itemize}

\subsection{Formulating EAP}
To incorporate the EAP into the model, we enforce a prior on $\theta$ that reduces entropy in high-activation regions $\mathbf{S}$ of the target domain. Following the maximum entropy principle \cite{grandvalet2004entropy-minimization}, we express this as a soft regularization:
\begin{equation}
    \mathbb{E}_{\theta}\left[ H(\mathbf{V}_{\mathbf{S}}, y_{\mathbf{S}} | \mathbf{S}) \right] \leq c~,
\end{equation}
where $c$ is a small constant enforcing high confidence in predictions. Using the principle of maximum entropy, we obtain:
\begin{align}
    P(\theta) & \propto \exp\left( -\lambda H(\mathbf{V}_{\mathbf{S}}, y_{\mathbf{S}} | \mathbf{S}) \right)~,
    \\
    & \propto \exp\left( -\lambda H(y_{\mathbf{S}} | \mathbf{V}_{\mathbf{S}}, \mathbf{S}) \right)~,
\end{align}
where $\lambda > 0$ is the Lagrange multiplier corresponding to constant $c$, which balances the effect of EAP on the model’s training objective.

\subsection{Empirical Estimation of EAP}
To implement the EAP, we estimate the conditional entropy $H(y | \mathbf{V}, \mathbf{S})$ by focusing on high-activation regions $\mathbf{S}$ in the target domain. This conditional entropy captures prediction uncertainty within the specific spatial region $\mathbf{S}$, allowing us to concentrate adaptation efforts on regions aligned with platform-specific activations. Using an empirical plug-in estimator, we approximate this entropy as:
\begin{equation}
    H_{\mathrm{emp}}(y | \mathbf{V}, \mathbf{S}) = \mathbb{E}_{\mathbf{V}, y, \mathbf{S}} \left[ \hat{P}(y | \mathbf{V}, \mathbf{S}) \log \hat{P}(y | \mathbf{V}, \mathbf{S}) \right],
\end{equation}
where $\hat{P}(y | \mathbf{V}, \mathbf{S})$ is the empirical prediction probability conditioned on the event voxel grid $\mathbf{V}$ and restricted to region $\mathbf{S}$. By minimizing $H_{\mathrm{emp}}(y | \mathbf{V}, \mathbf{S})$, we encourage confident predictions within these regions, aligning the model’s predictions with the target domain’s activation patterns.

\subsection{Integrating EAP into the Training Objective}
To incorporate EAP into the model’s training, we define the overall objective function as a maximum-a-posteriori (MAP) estimation:
\begin{equation}
C(\theta) = \mathcal{L}(\theta) - \lambda H_{\mathrm{emp}}(y | \mathbf{V}, \mathbf{S})~,
\end{equation}
where $\mathcal{L}(\theta)$ represents the supervised loss on source data. $H_{\mathrm{emp}}(y | \mathbf{V}, \mathbf{S})$ minimizes uncertainty in the target domain by leveraging EAP over high-activation regions.

By focusing on high-activation areas, the event camera perception model learns to adapt to the target domain’s unique event-triggered patterns, achieving robust adaptation across platforms. This approach captures and emphasizes platform-specific activation patterns, making EAP an effective regularization for confident adaptation in event-based cross-platform scenarios.

\section{Event Activation Prior: Observation}
\label{sec:observations}

In this section, we provide concrete evidence supporting the proposed Event Activation Prior (EAP) by analyzing the platform-specific activation patterns in both static and dynamic regions. The evidence is presented through class distribution statistics and maps, which highlight the unique activation characteristics of each platform.

\begin{table*}[t]
    \centering
    \caption{
        \textbf{The class distribution maps} of static classes among the \includegraphics[width=0.018\linewidth]{figures/icons/vehicle.png}~vehicle ($\mathcal{P}^{\textcolor{fly_green}{\mathbf{v}}}$), \includegraphics[width=0.021\linewidth]{figures/icons/drone.png}~drone ($\mathcal{P}^{\textcolor{fly_red}{\mathbf{d}}}$), and \includegraphics[width=0.019\linewidth]{figures/icons/quadruped.png}~quadruped ($\mathcal{P}^{\textcolor{fly_blue}{\mathbf{q}}}$) platforms, respectively, in the proposed \textbf{\textit{\textcolor{fly_green}{E}\textcolor{fly_red}{X}\textcolor{fly_blue}{Po}}} benchmark. The brighter the color, the higher the probability of occurrences. Best viewed in colors.}
    \vspace{-0.2cm}
    \resizebox{0.93\linewidth}{!}{
    \begin{tabular}{p{20pt}<{\centering}|p{80pt}<{\centering}|p{40pt}<{\centering}|c|c|c}
    \toprule
    \multirow{2}{*}{\textbf{ID}} & \multirow{2}{*}{\textbf{Class}} & \multirow{2}{*}{\textbf{Type}} & \includegraphics[width=0.024\linewidth]{figures/icons/vehicle.png} & \includegraphics[width=0.03\linewidth]{figures/icons/drone.png} & \includegraphics[width=0.024\linewidth]{figures/icons/quadruped.png}
    \\
    & & & \textbf{\textcolor{fly_green}{vehicle}} ($\mathcal{P}^{\textcolor{fly_green}{\mathbf{v}}}$) & \textbf{\textcolor{fly_red}{drone}} ($\mathcal{P}^{\textcolor{fly_red}{\mathbf{d}}}$) & \textbf{\textcolor{fly_blue}{quadruped}} ($\mathcal{P}^{\textcolor{fly_blue}{\mathbf{q}}}$)
    \\\midrule\midrule
    $0$ & \texttt{road} & static & \begin{minipage}[b]{0.44\columnwidth}\centering\raisebox{-.4\height}{\includegraphics[width=\linewidth]{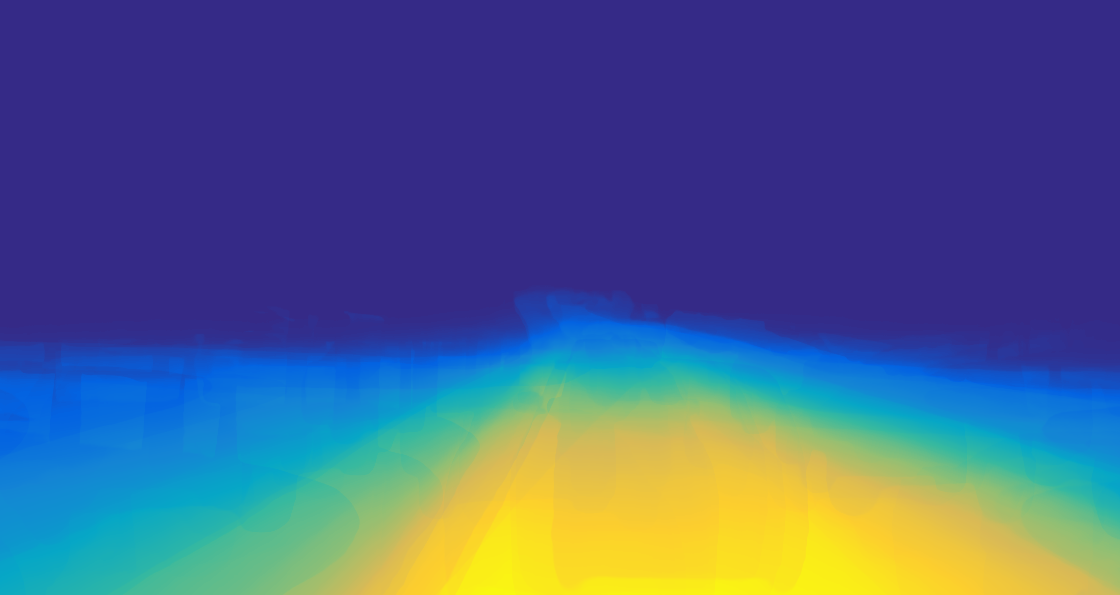}}\end{minipage} & \begin{minipage}[b]{0.44\columnwidth}\centering\raisebox{-.4\height}{\includegraphics[width=\linewidth]{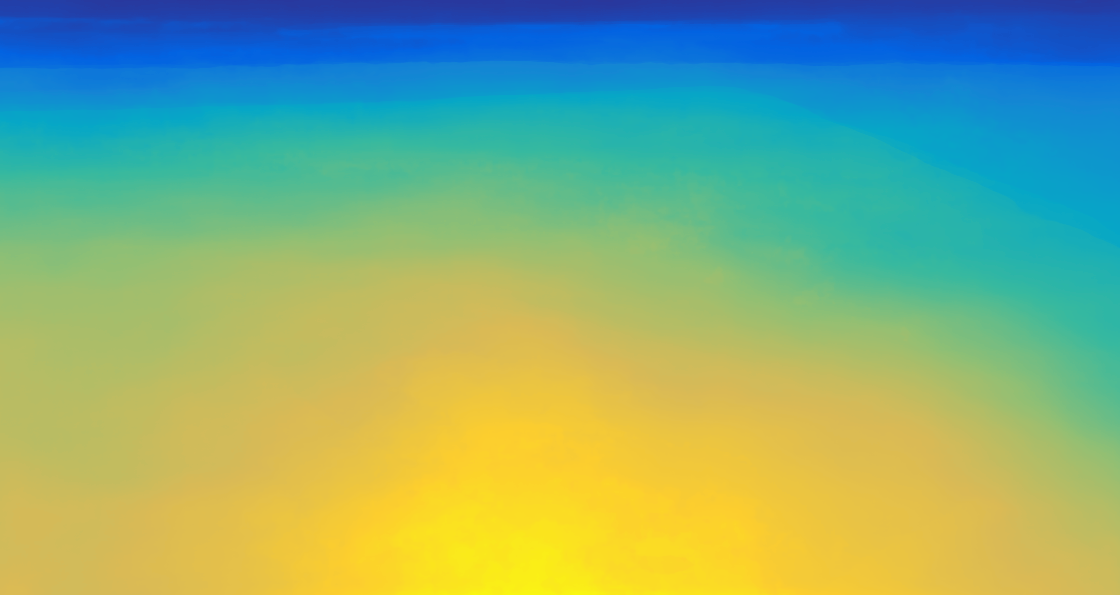}}\end{minipage} & \begin{minipage}[b]{0.44\columnwidth}\centering\raisebox{-.4\height}{\includegraphics[width=\linewidth]{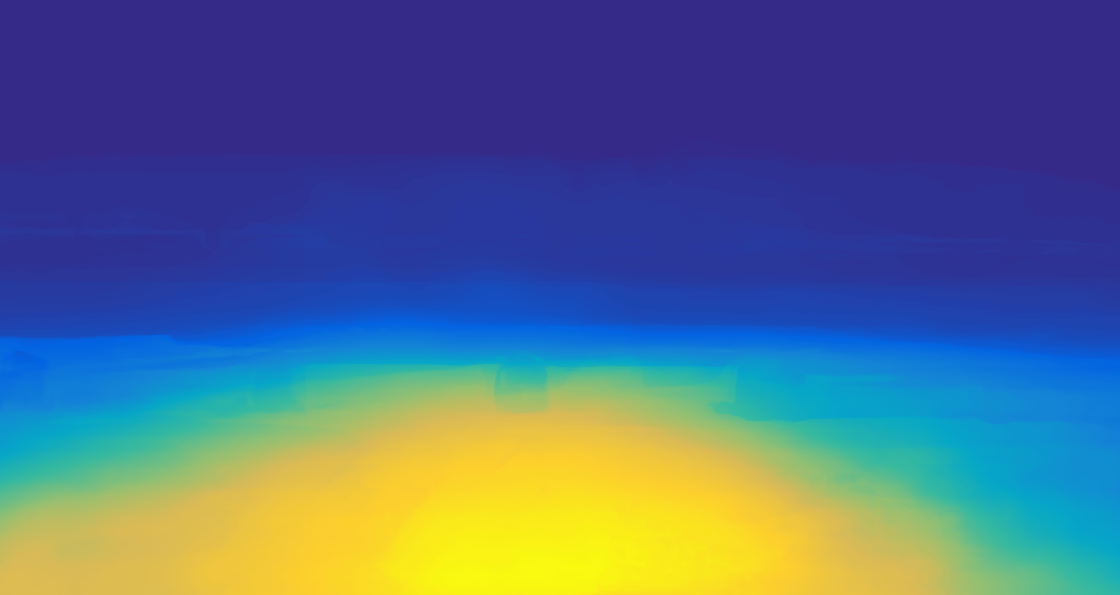}}\end{minipage}
    \\\midrule
    $1$ & \texttt{sidewalk} & static & \begin{minipage}[b]{0.44\columnwidth}\centering\raisebox{-.4\height}{\includegraphics[width=\linewidth]{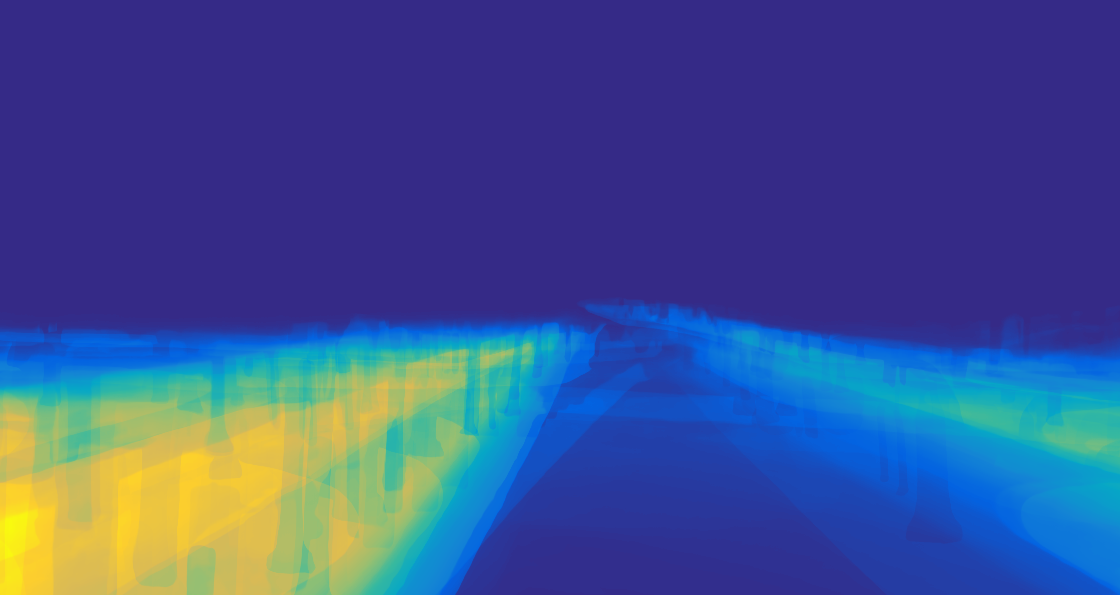}}\end{minipage} & \begin{minipage}[b]{0.44\columnwidth}\centering\raisebox{-.4\height}{\includegraphics[width=\linewidth]{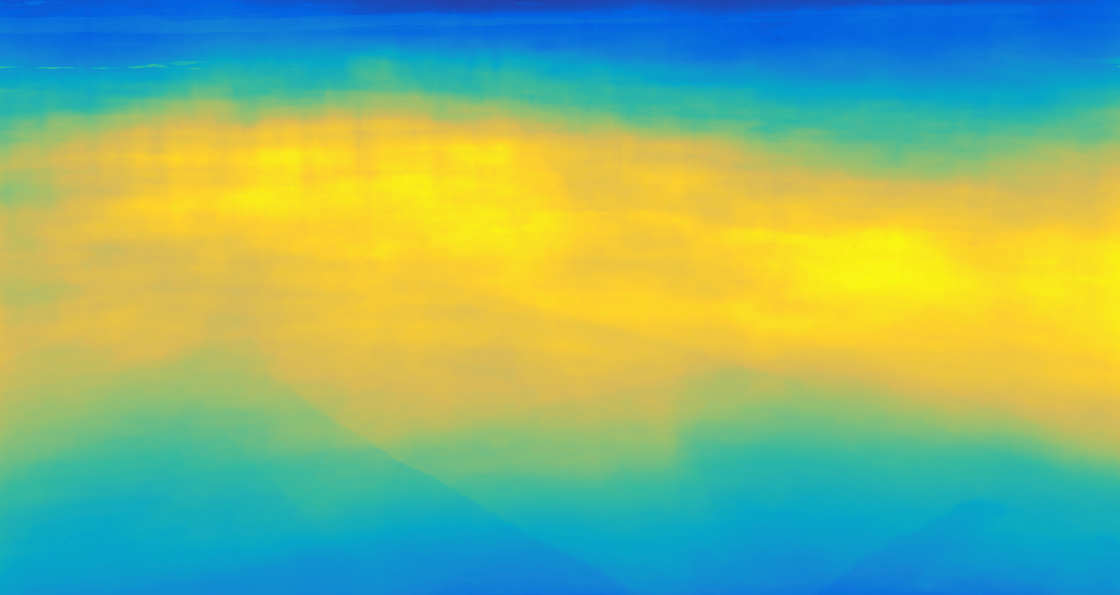}}\end{minipage} & \begin{minipage}[b]{0.44\columnwidth}\centering\raisebox{-.4\height}{\includegraphics[width=\linewidth]{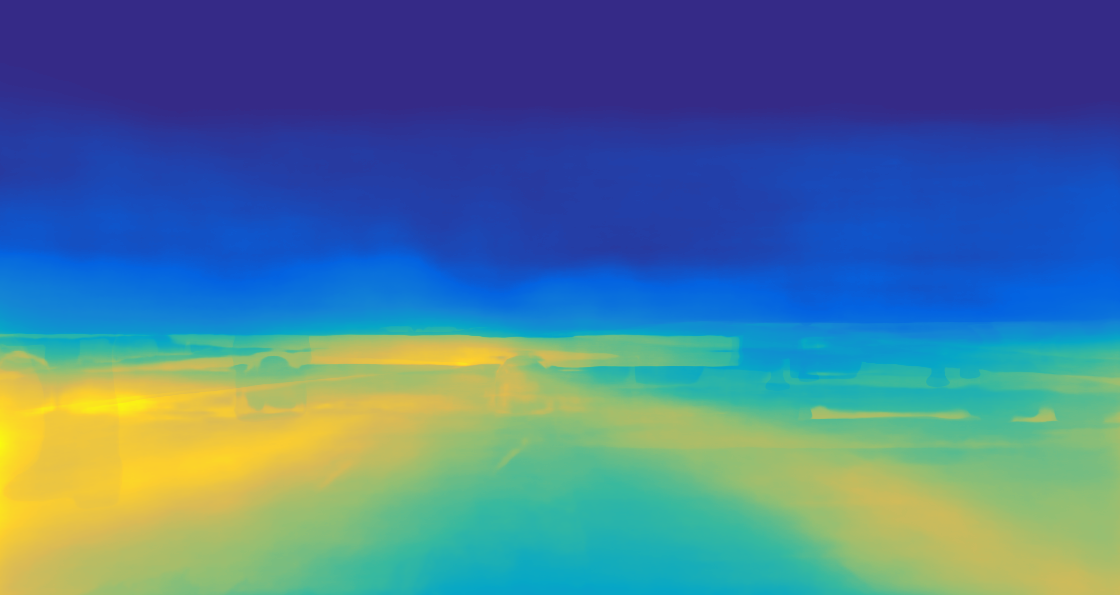}}\end{minipage}
    \\\midrule
    $2$ & \texttt{building} & static & \begin{minipage}[b]{0.44\columnwidth}\centering\raisebox{-.4\height}{\includegraphics[width=\linewidth]{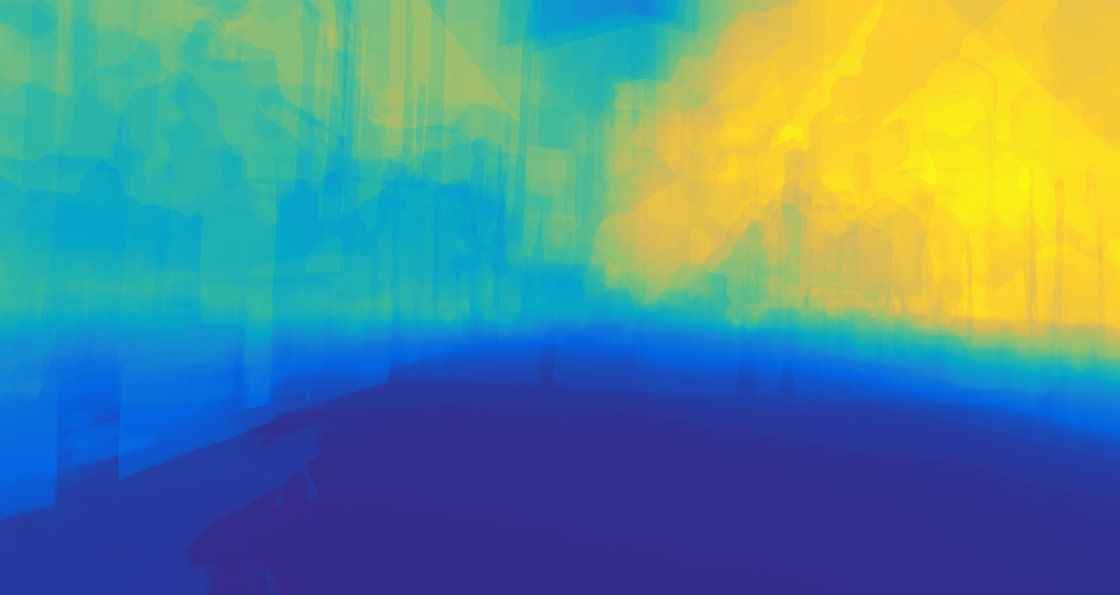}}\end{minipage} & \begin{minipage}[b]{0.44\columnwidth}\centering\raisebox{-.4\height}{\includegraphics[width=\linewidth]{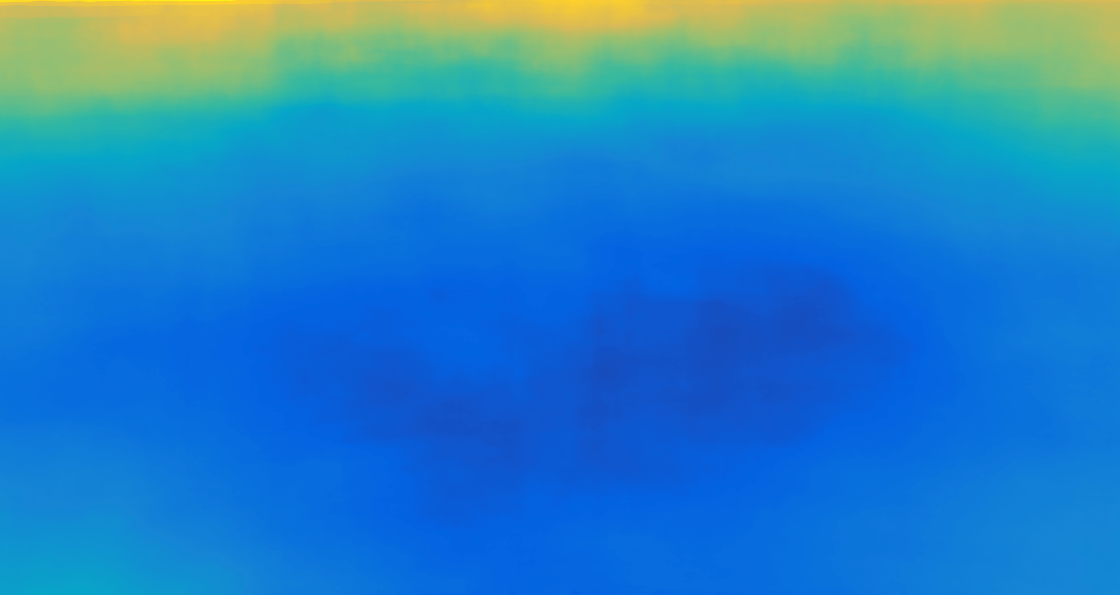}}\end{minipage} & \begin{minipage}[b]{0.44\columnwidth}\centering\raisebox{-.4\height}{\includegraphics[width=\linewidth]{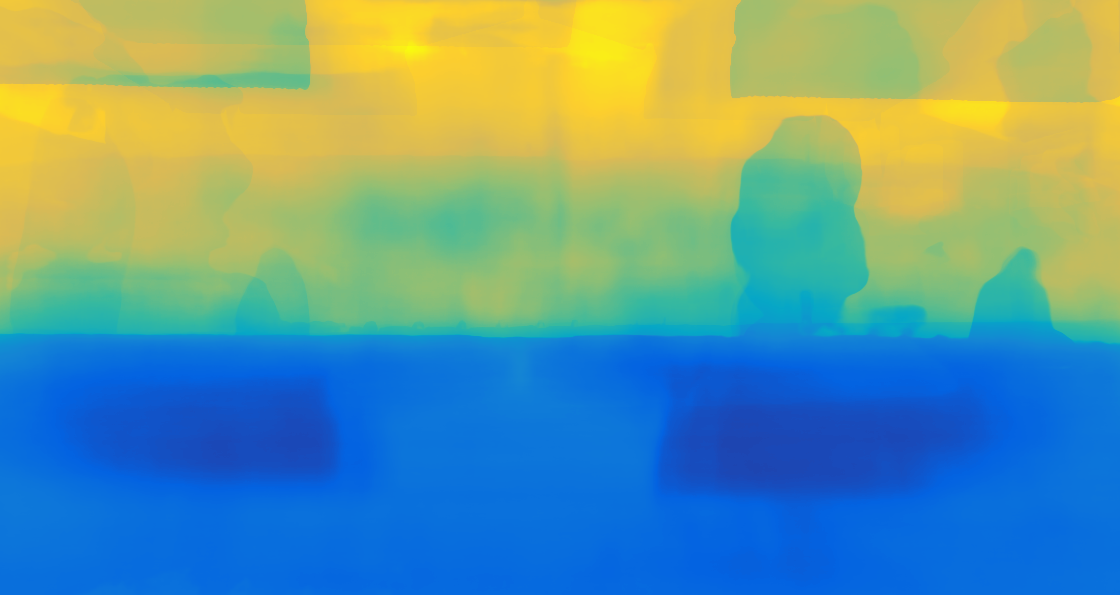}}\end{minipage}
    \\\midrule
    $3$ & \texttt{wall} & static & \begin{minipage}[b]{0.44\columnwidth}\centering\raisebox{-.4\height}{\includegraphics[width=\linewidth]{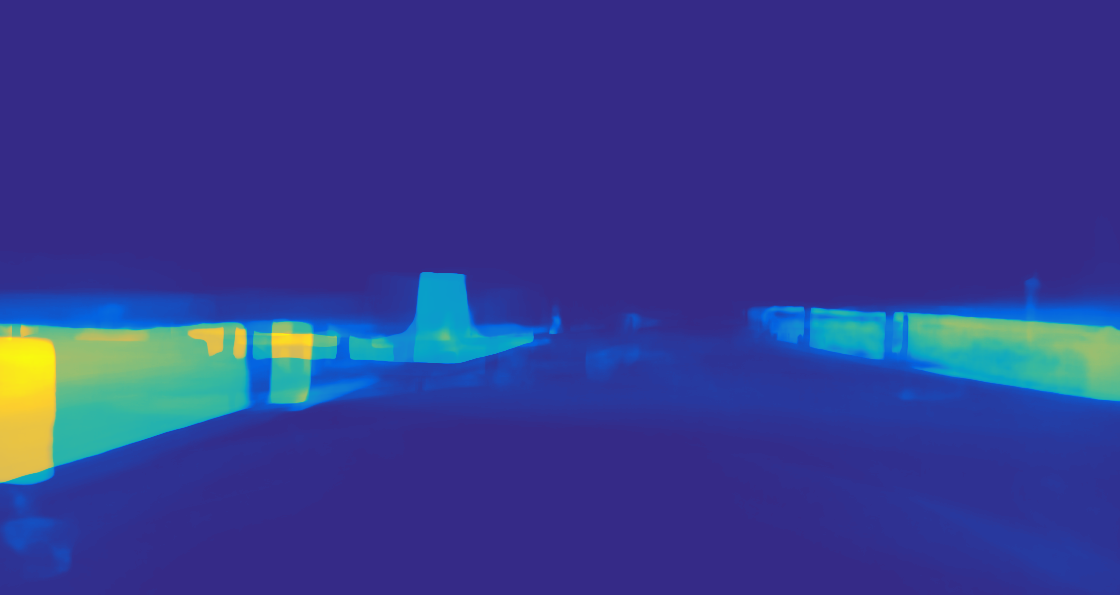}}\end{minipage} & \begin{minipage}[b]{0.44\columnwidth}\centering\raisebox{-.4\height}{\includegraphics[width=\linewidth]{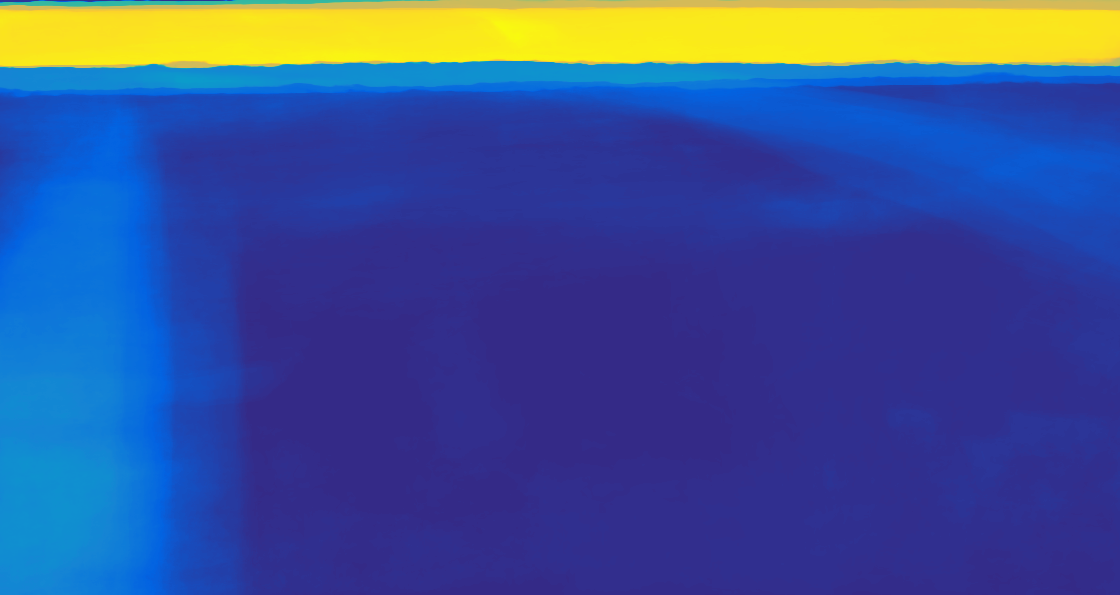}}\end{minipage} & \begin{minipage}[b]{0.44\columnwidth}\centering\raisebox{-.4\height}{\includegraphics[width=\linewidth]{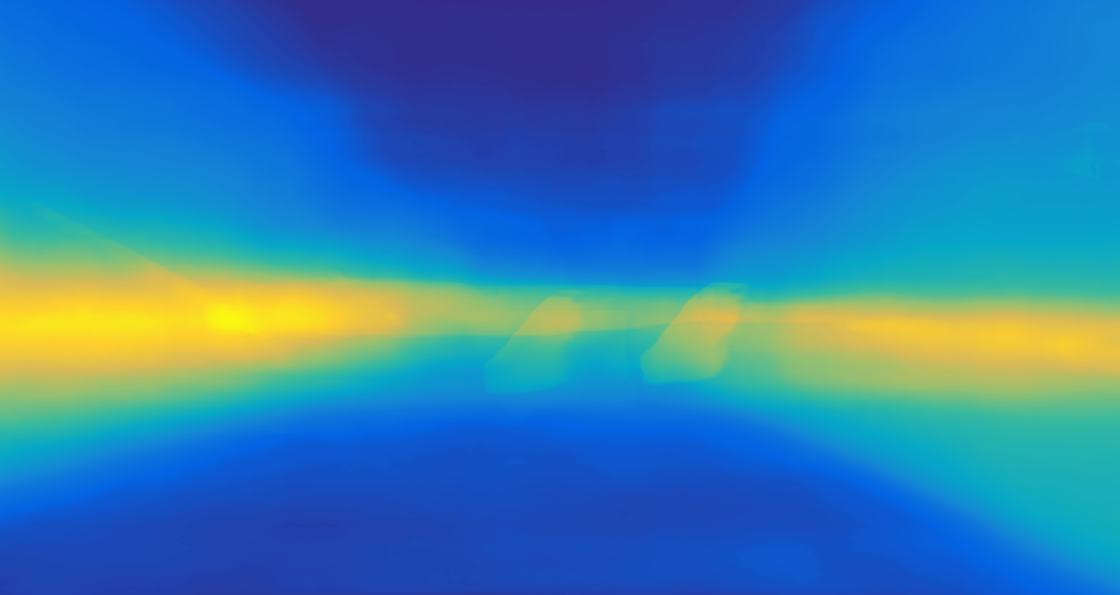}}\end{minipage}
    \\\midrule
    $4$ & \texttt{fence} & static & \begin{minipage}[b]{0.44\columnwidth}\centering\raisebox{-.4\height}{\includegraphics[width=\linewidth]{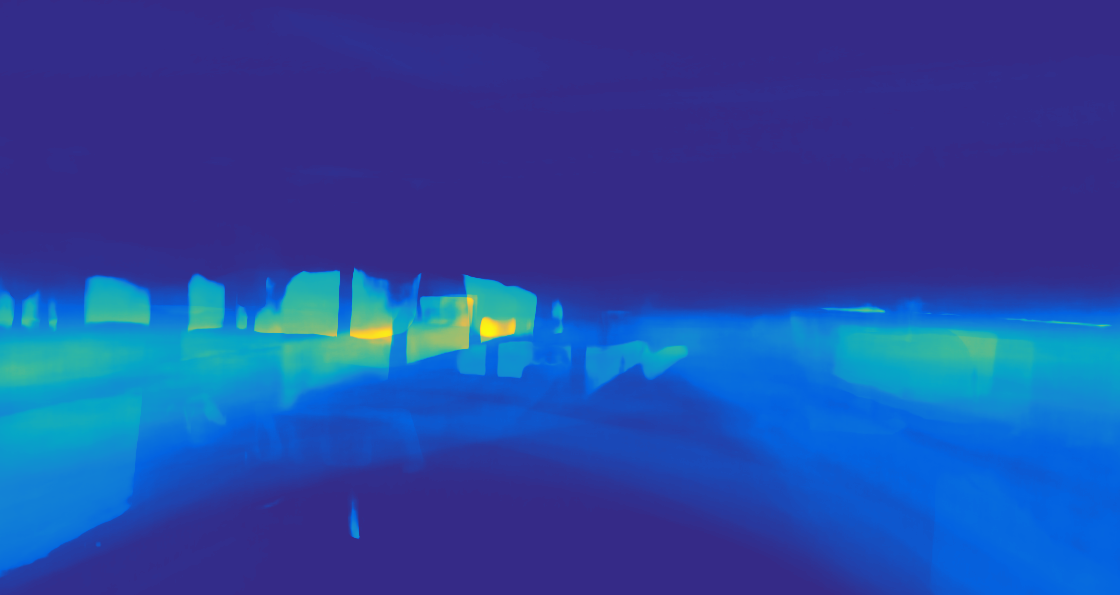}}\end{minipage} & \begin{minipage}[b]{0.44\columnwidth}\centering\raisebox{-.4\height}{\includegraphics[width=\linewidth]{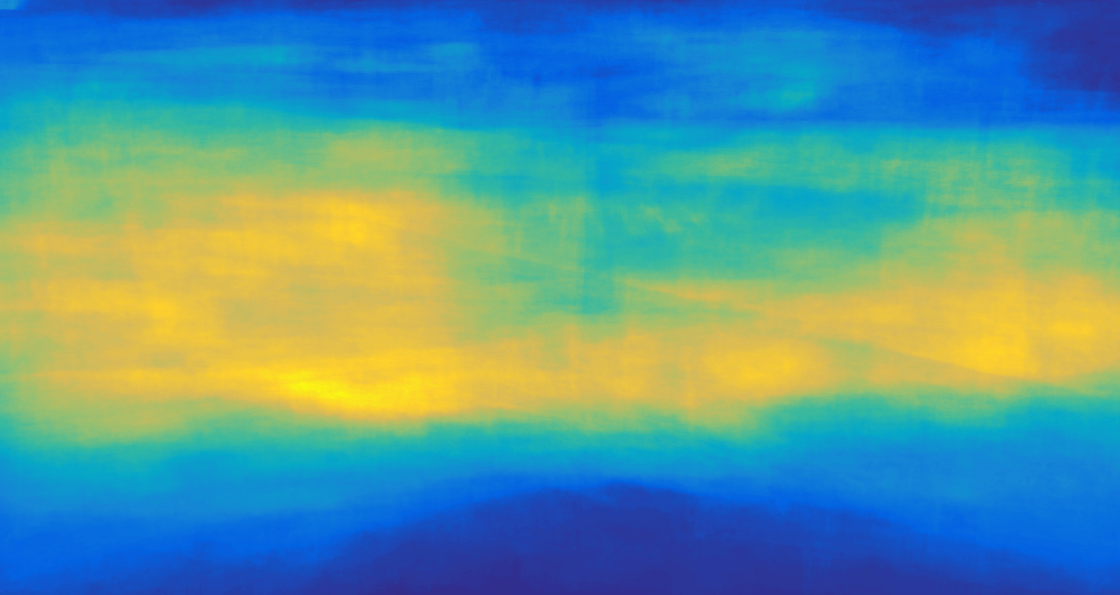}}\end{minipage} & \begin{minipage}[b]{0.44\columnwidth}\centering\raisebox{-.4\height}{\includegraphics[width=\linewidth]{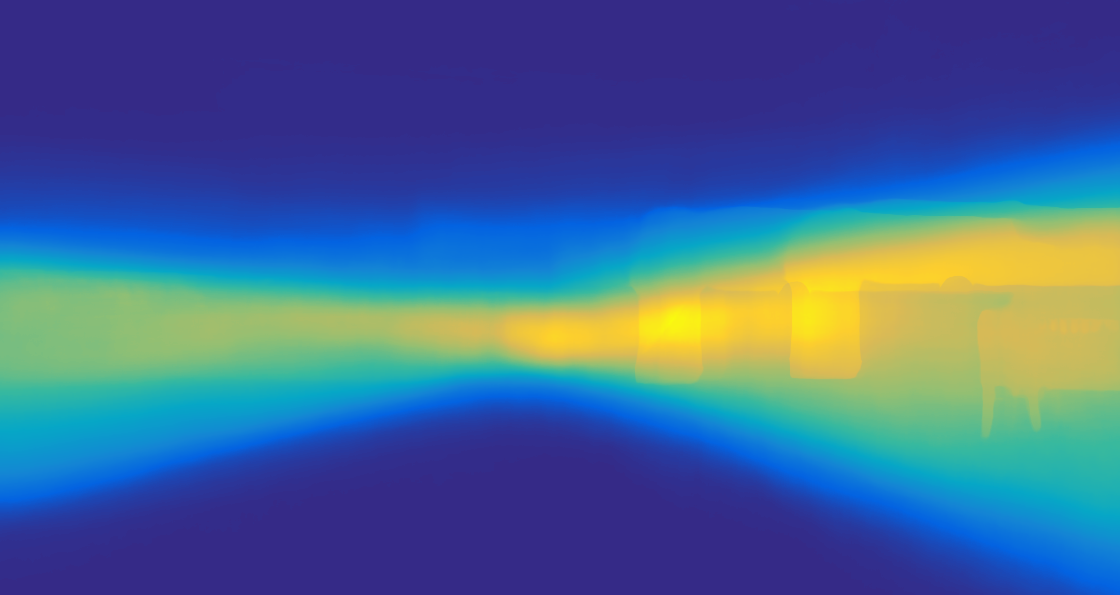}}\end{minipage}
    \\\midrule
    $5$ & \texttt{pole} & static & \begin{minipage}[b]{0.44\columnwidth}\centering\raisebox{-.4\height}{\includegraphics[width=\linewidth]{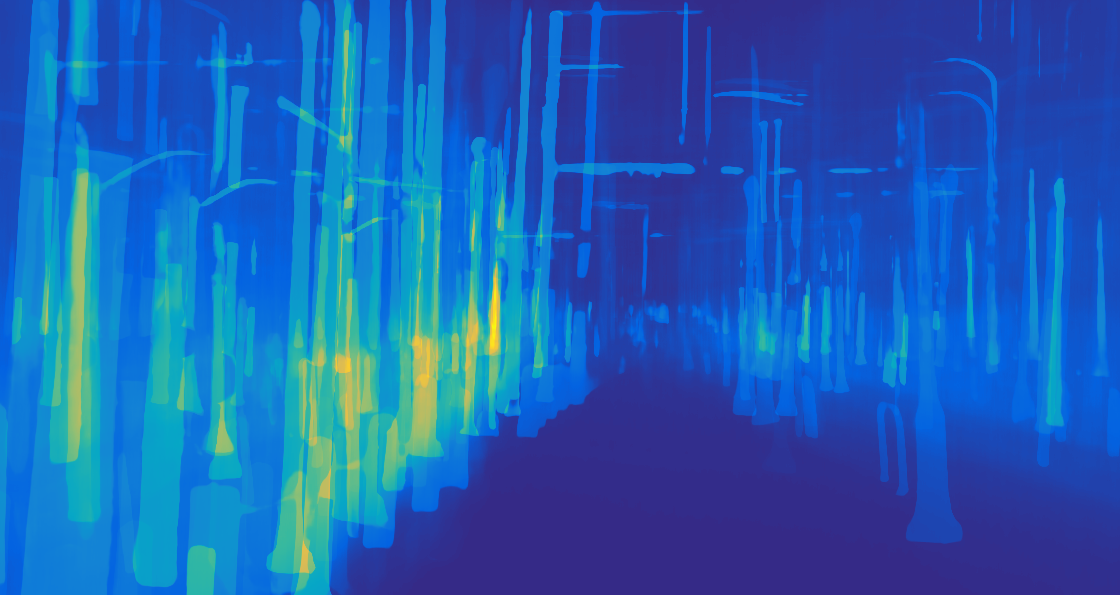}}\end{minipage} & \begin{minipage}[b]{0.44\columnwidth}\centering\raisebox{-.4\height}{\includegraphics[width=\linewidth]{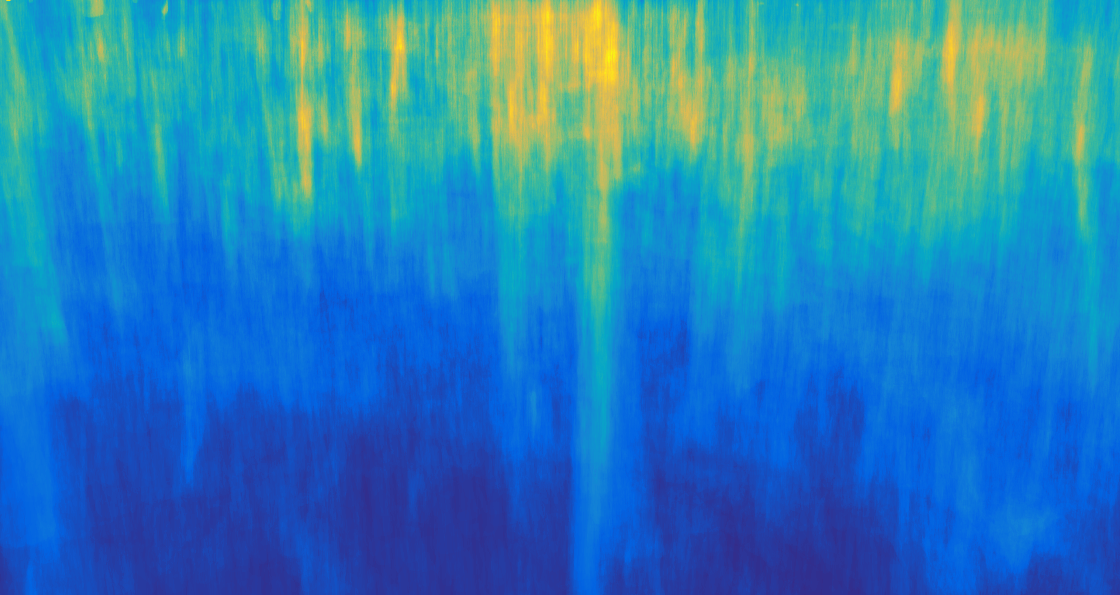}}\end{minipage} & \begin{minipage}[b]{0.44\columnwidth}\centering\raisebox{-.4\height}{\includegraphics[width=\linewidth]{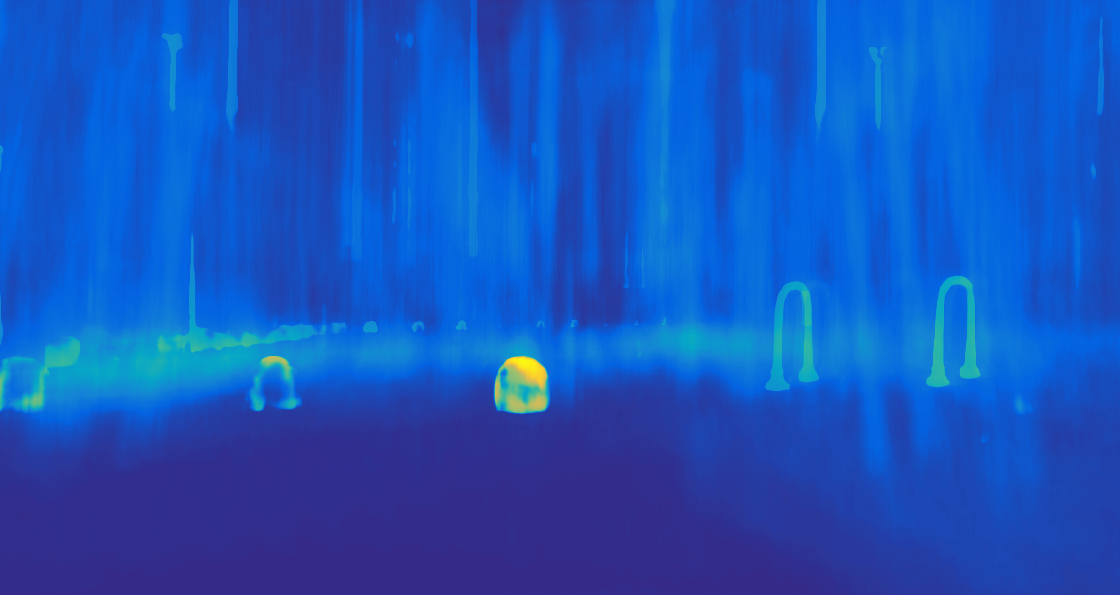}}\end{minipage}
    \\\midrule
    $6$ & \texttt{traffic-light} & static & \begin{minipage}[b]{0.44\columnwidth}\centering\raisebox{-.4\height}{\includegraphics[width=\linewidth]{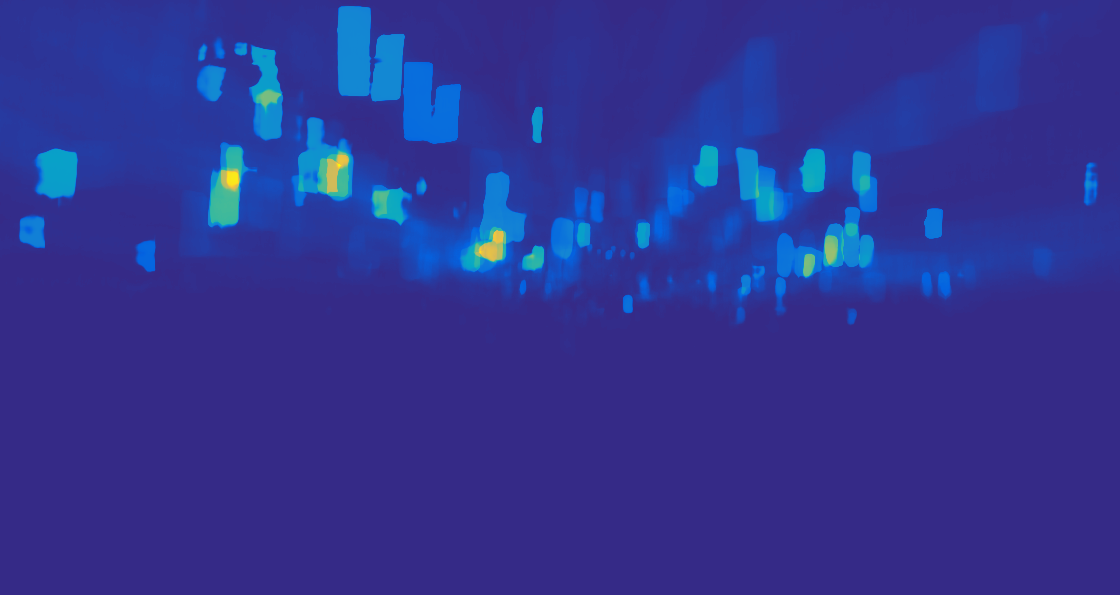}}\end{minipage} & \begin{minipage}[b]{0.44\columnwidth}\centering\raisebox{-.4\height}{\includegraphics[width=\linewidth]{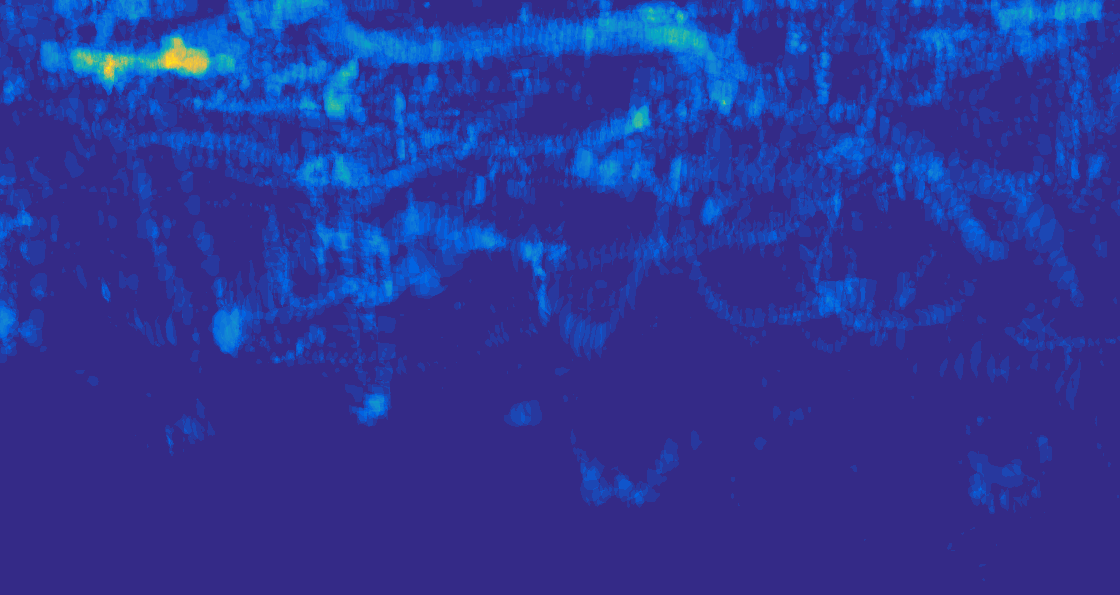}}\end{minipage} & \begin{minipage}[b]{0.44\columnwidth}\centering\raisebox{-.4\height}{\includegraphics[width=\linewidth]{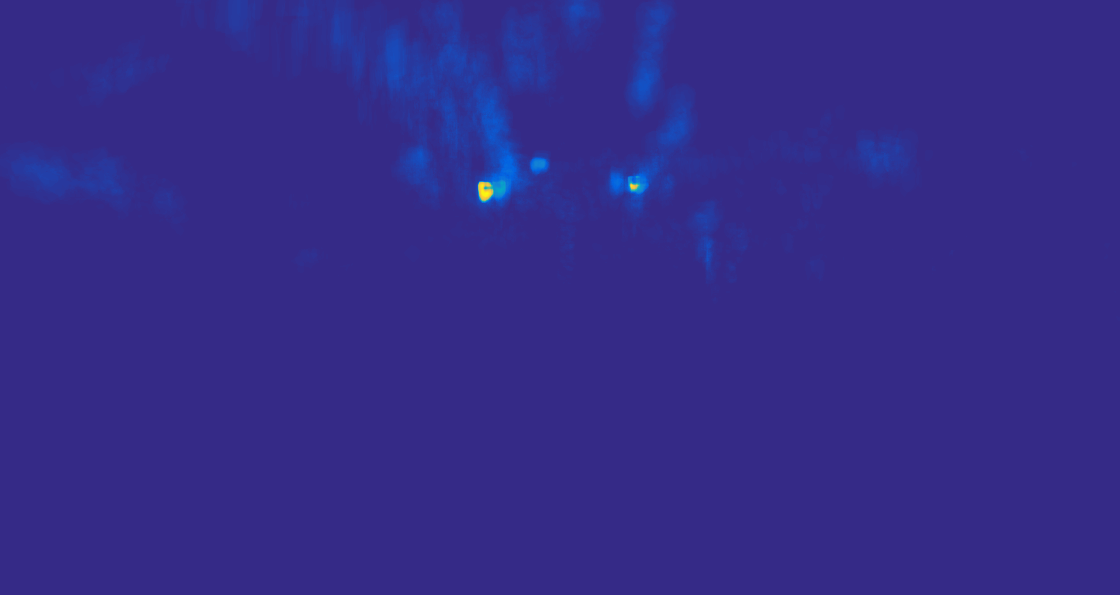}}\end{minipage}
    \\\midrule
    $7$ & \texttt{traffic-sign} & static & \begin{minipage}[b]{0.44\columnwidth}\centering\raisebox{-.4\height}{\includegraphics[width=\linewidth]{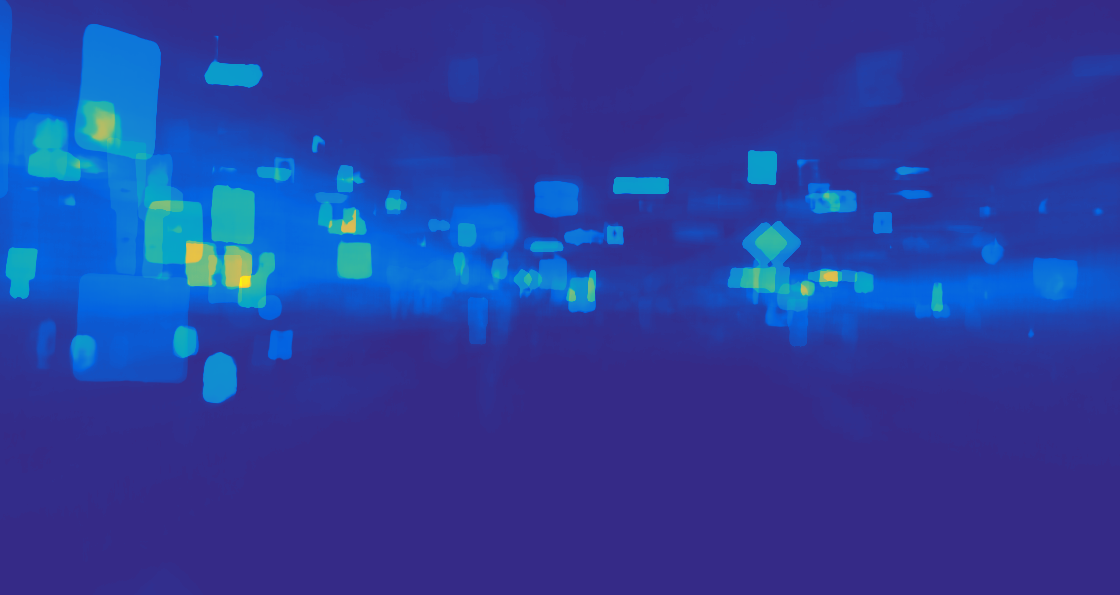}}\end{minipage} & \begin{minipage}[b]{0.44\columnwidth}\centering\raisebox{-.4\height}{\includegraphics[width=\linewidth]{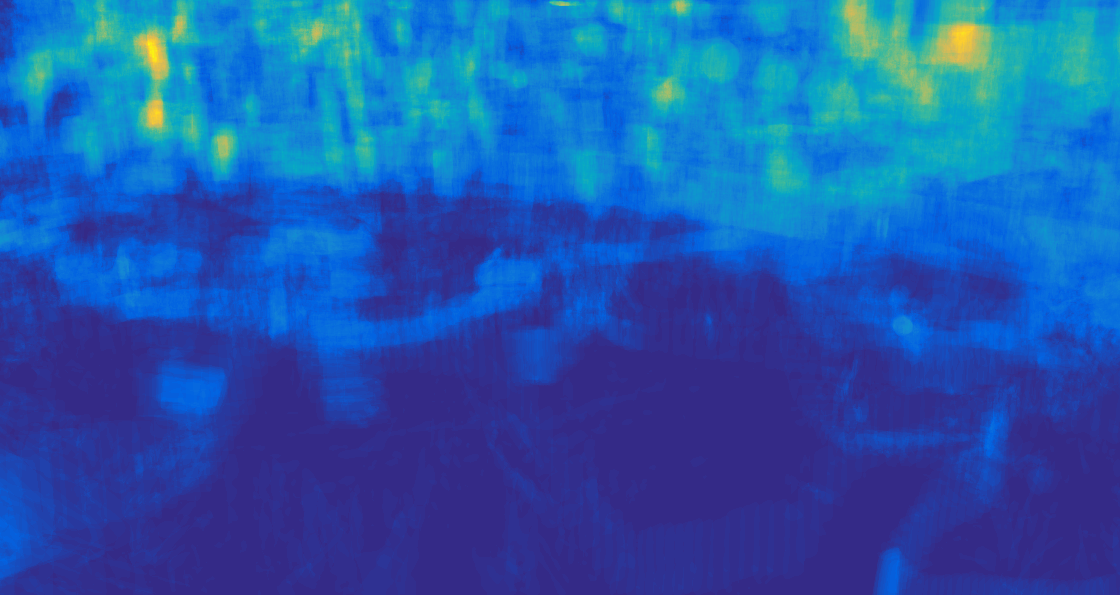}}\end{minipage} & \begin{minipage}[b]{0.44\columnwidth}\centering\raisebox{-.4\height}{\includegraphics[width=\linewidth]{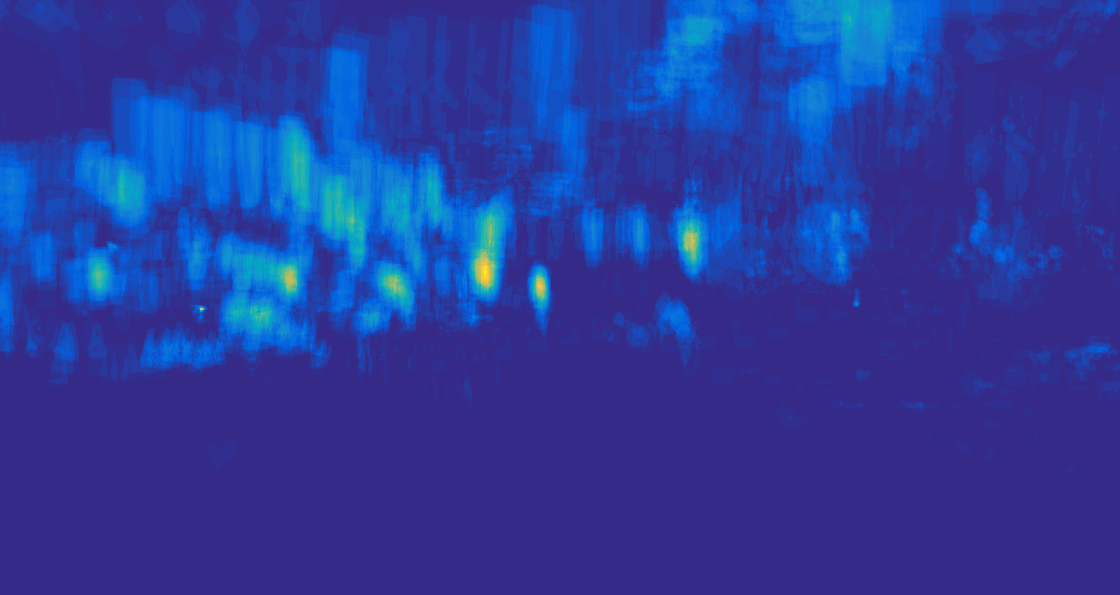}}\end{minipage}
    \\\midrule
    $8$ & \texttt{vegetation} & static & \begin{minipage}[b]{0.44\columnwidth}\centering\raisebox{-.4\height}{\includegraphics[width=\linewidth]{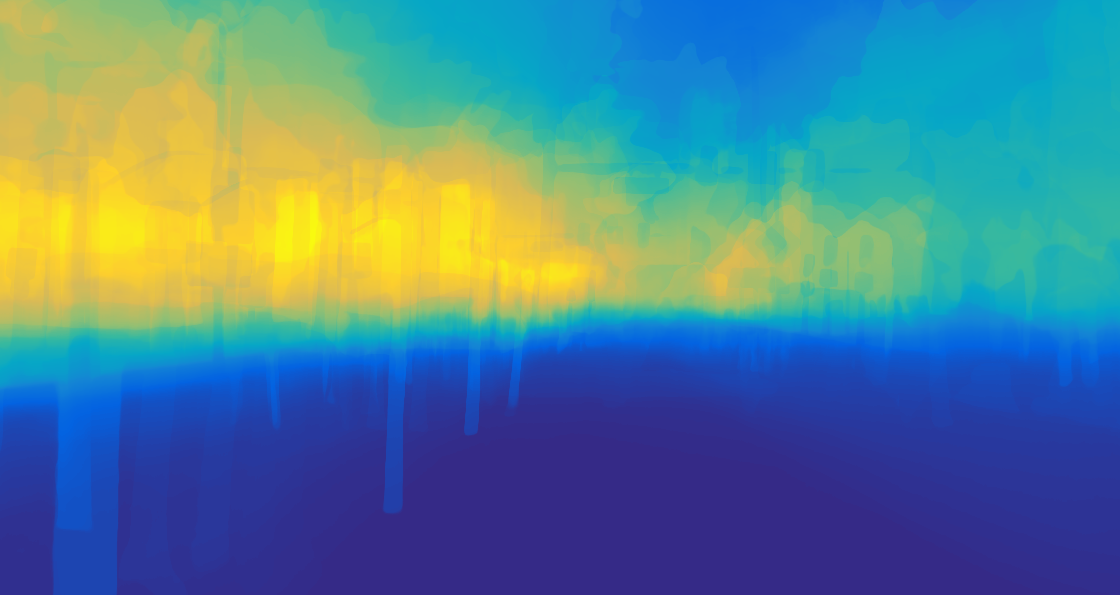}}\end{minipage} & \begin{minipage}[b]{0.44\columnwidth}\centering\raisebox{-.4\height}{\includegraphics[width=\linewidth]{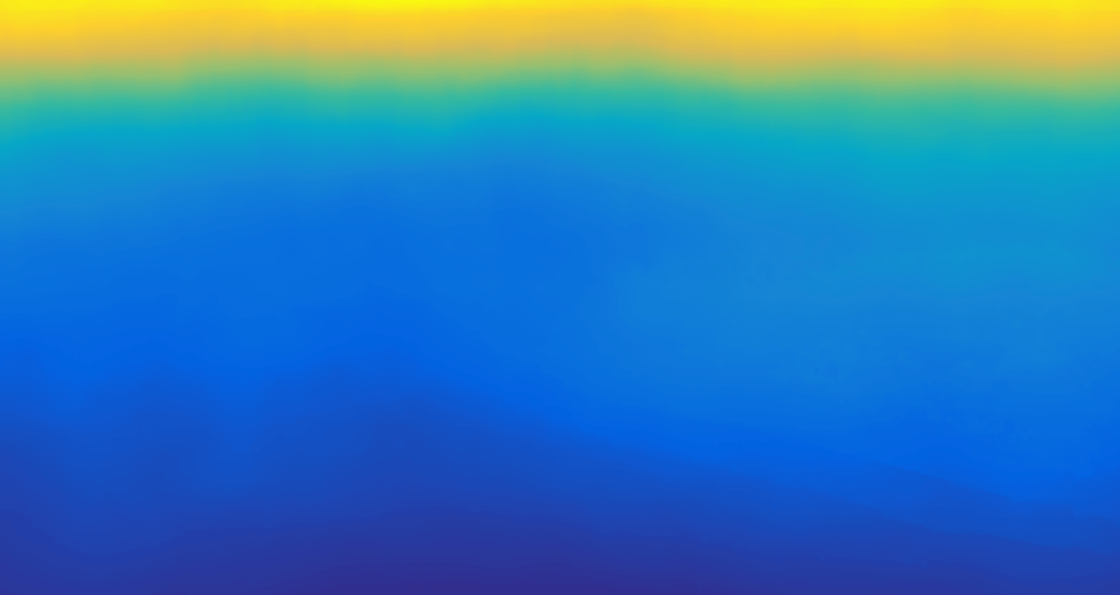}}\end{minipage} & \begin{minipage}[b]{0.44\columnwidth}\centering\raisebox{-.4\height}{\includegraphics[width=\linewidth]{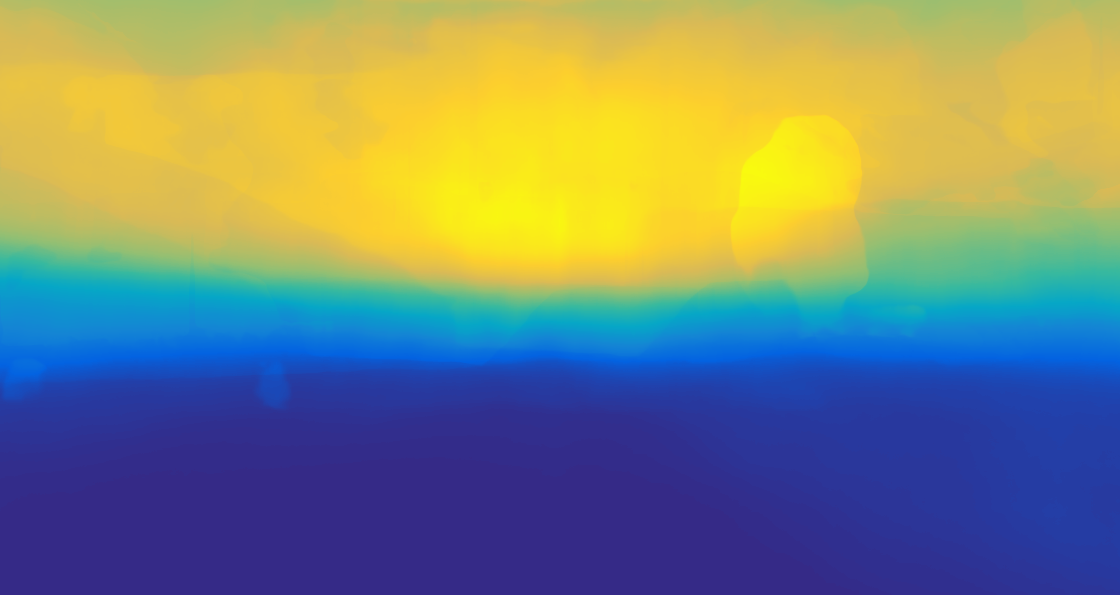}}\end{minipage}
    \\\midrule
    $9$ & \texttt{terrain} & static & \begin{minipage}[b]{0.44\columnwidth}\centering\raisebox{-.4\height}{\includegraphics[width=\linewidth]{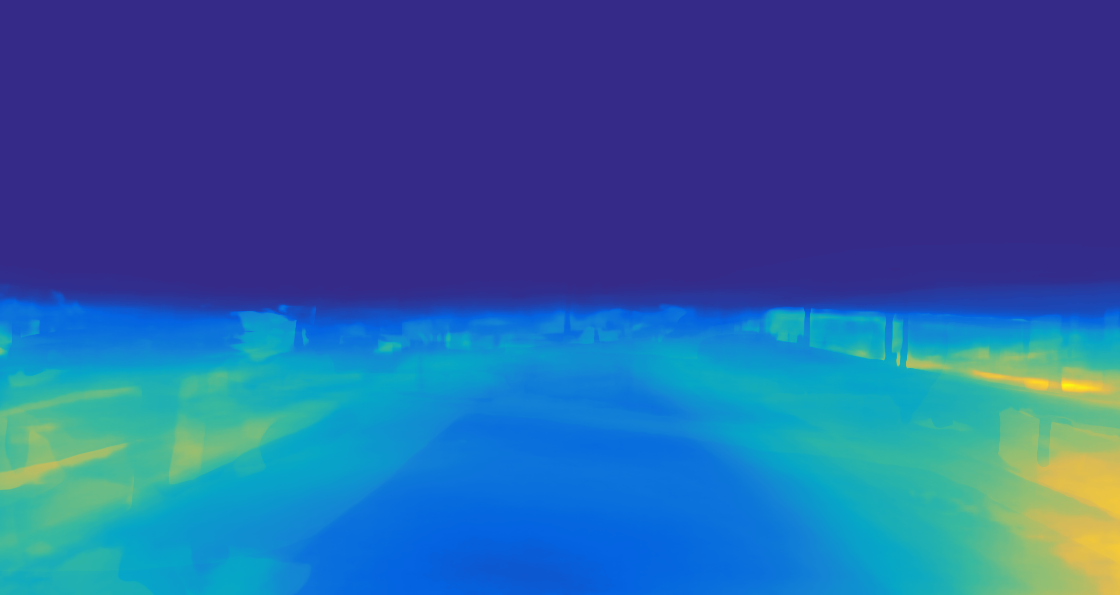}}\end{minipage} & \begin{minipage}[b]{0.44\columnwidth}\centering\raisebox{-.4\height}{\includegraphics[width=\linewidth]{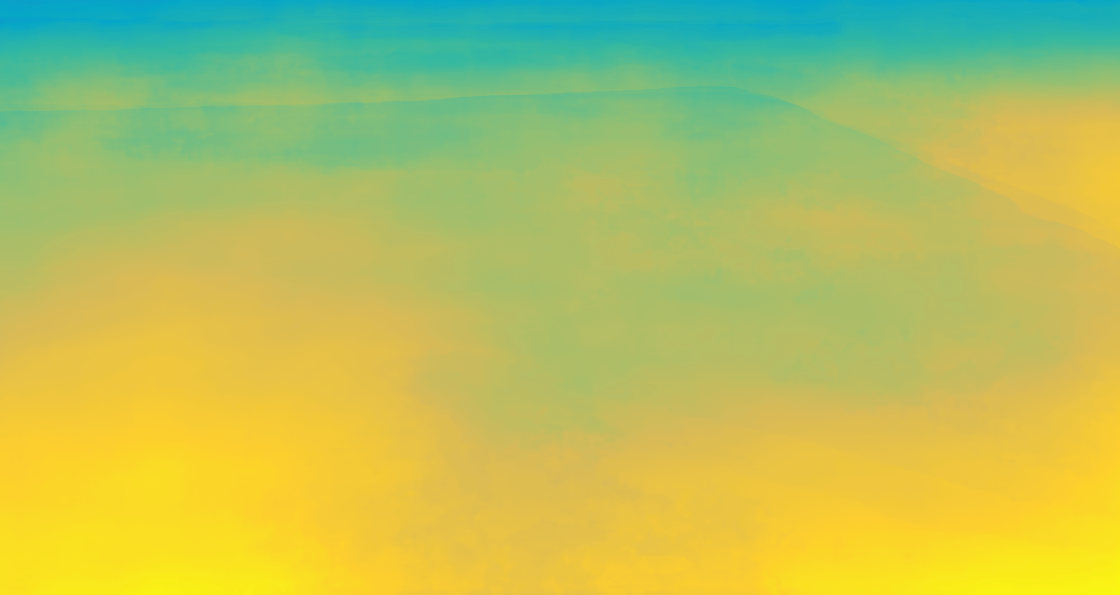}}\end{minipage} & \begin{minipage}[b]{0.44\columnwidth}\centering\raisebox{-.4\height}{\includegraphics[width=\linewidth]{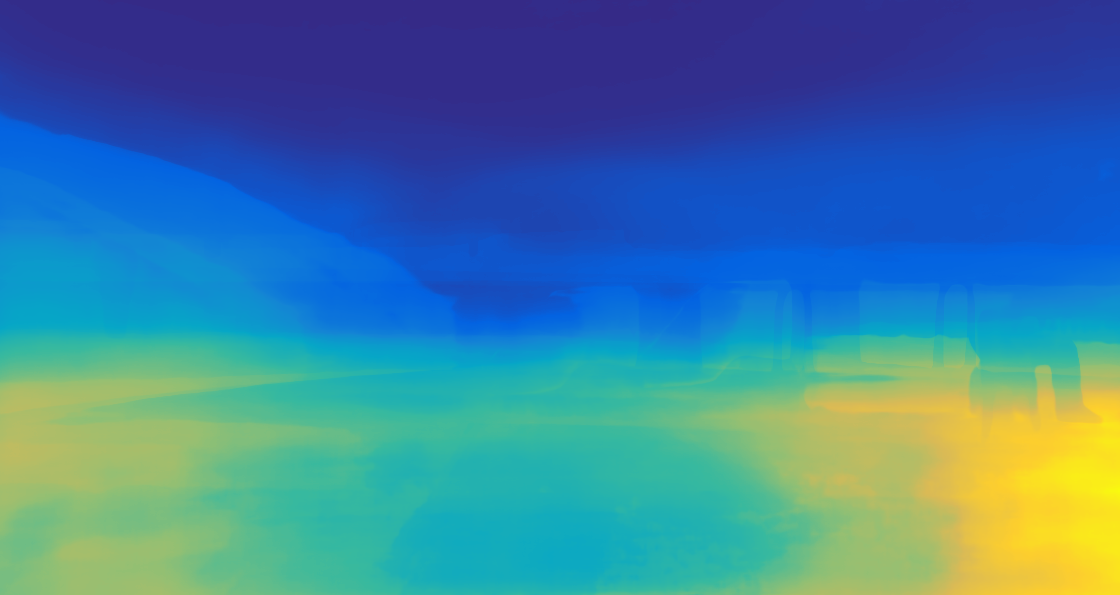}}\end{minipage}
    \\\midrule
    $10$ & \texttt{sky} & static & \begin{minipage}[b]{0.44\columnwidth}\centering\raisebox{-.4\height}{\includegraphics[width=\linewidth]{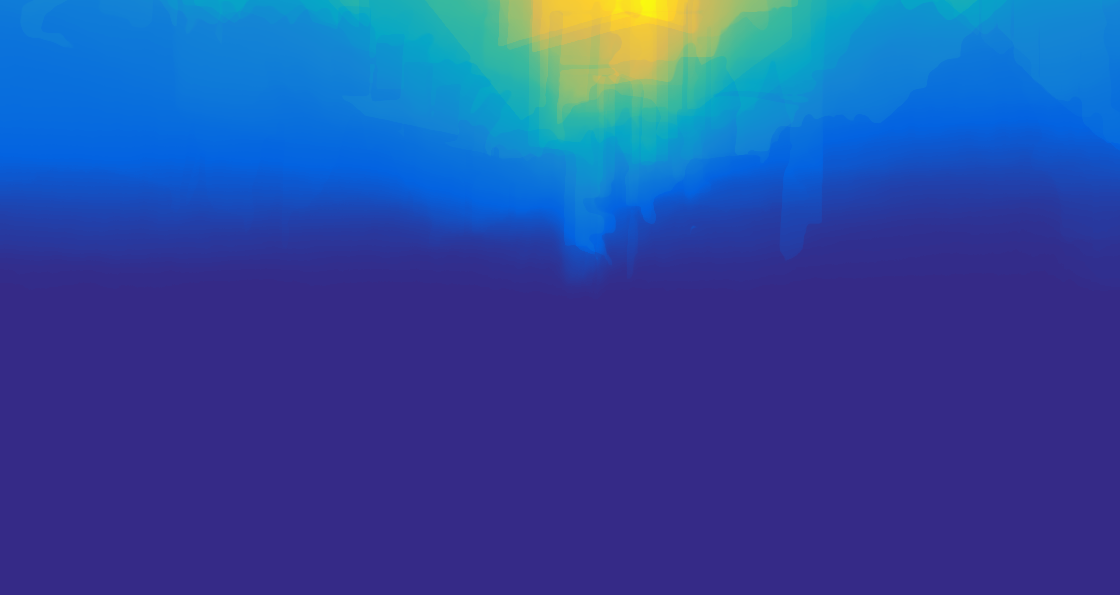}}\end{minipage} & \begin{minipage}[b]{0.44\columnwidth}\centering\raisebox{-.4\height}{\includegraphics[width=\linewidth]{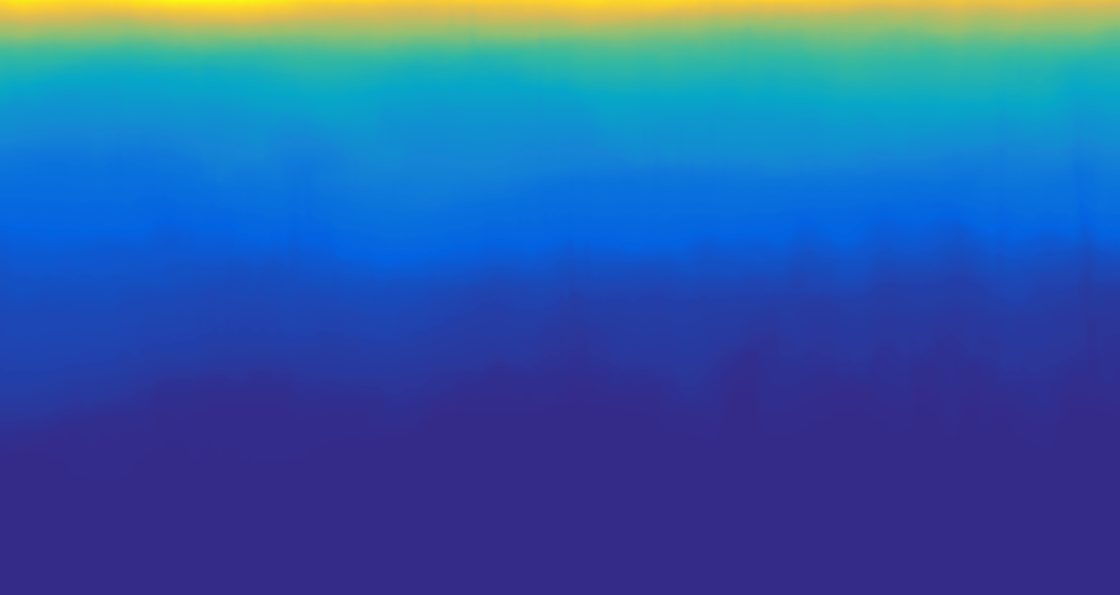}}\end{minipage} & \begin{minipage}[b]{0.44\columnwidth}\centering\raisebox{-.4\height}{\includegraphics[width=\linewidth]{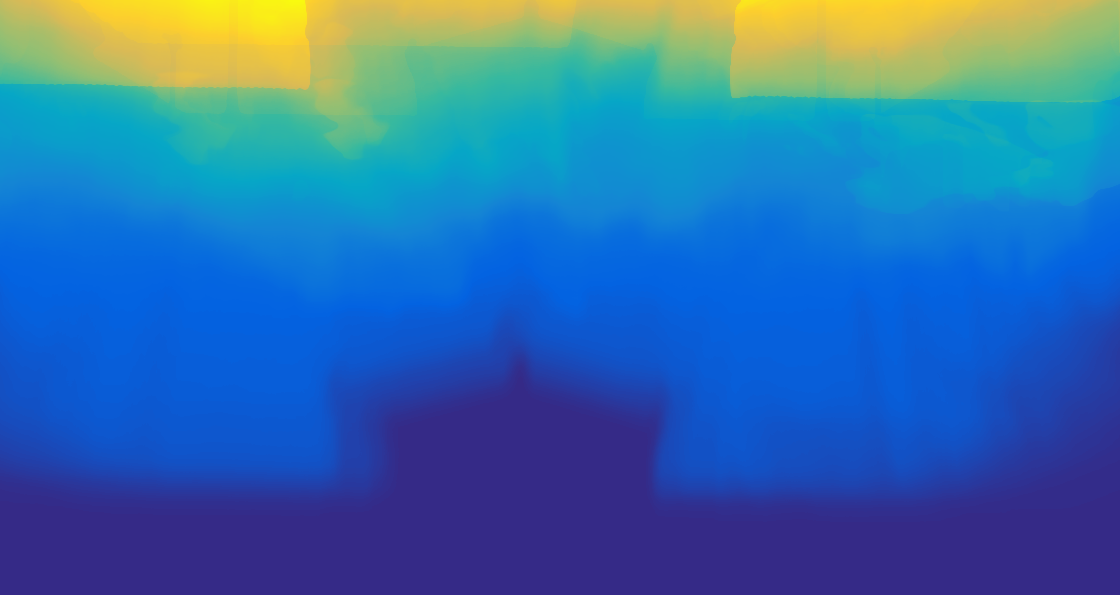}}\end{minipage}
    \\\bottomrule
\end{tabular}}
\label{tab:prior_static}
\vspace{-0.4cm}
\end{table*}

\begin{table*}[t]
    \centering
    \caption{
        \textbf{The class distribution maps} of dynamic classes among the \includegraphics[width=0.018\linewidth]{figures/icons/vehicle.png}~vehicle ($\mathcal{P}^{\textcolor{fly_green}{\mathbf{v}}}$), \includegraphics[width=0.021\linewidth]{figures/icons/drone.png}~drone ($\mathcal{P}^{\textcolor{fly_red}{\mathbf{d}}}$), and \includegraphics[width=0.019\linewidth]{figures/icons/quadruped.png}~quadruped ($\mathcal{P}^{\textcolor{fly_blue}{\mathbf{q}}}$) platforms, respectively, in the proposed \textbf{\textit{\textcolor{fly_green}{E}\textcolor{fly_red}{X}\textcolor{fly_blue}{Po}}} benchmark. The brighter the color, the higher the probability of occurrences. Best viewed in colors.}
    \vspace{-0.2cm}
    \resizebox{0.94\linewidth}{!}{
    \begin{tabular}{p{20pt}<{\centering}|p{80pt}<{\centering}|p{40pt}<{\centering}|c|c|c}
    \toprule
    \multirow{2}{*}{\textbf{ID}} & \multirow{2}{*}{\textbf{Class}} & \multirow{2}{*}{\textbf{Type}} & \includegraphics[width=0.024\linewidth]{figures/icons/vehicle.png} & \includegraphics[width=0.03\linewidth]{figures/icons/drone.png} & \includegraphics[width=0.024\linewidth]{figures/icons/quadruped.png}
    \\
    & & & \textbf{\textcolor{fly_green}{vehicle}} ($\mathcal{P}^{\textcolor{fly_green}{\mathbf{v}}}$) & \textbf{\textcolor{fly_red}{drone}} ($\mathcal{P}^{\textcolor{fly_red}{\mathbf{d}}}$) & \textbf{\textcolor{fly_blue}{quadruped}} ($\mathcal{P}^{\textcolor{fly_blue}{\mathbf{q}}}$)
    \\\midrule\midrule
    $11$ & \texttt{person} & dynamic & \begin{minipage}[b]{0.44\columnwidth}\centering\raisebox{-.4\height}{\includegraphics[width=\linewidth]{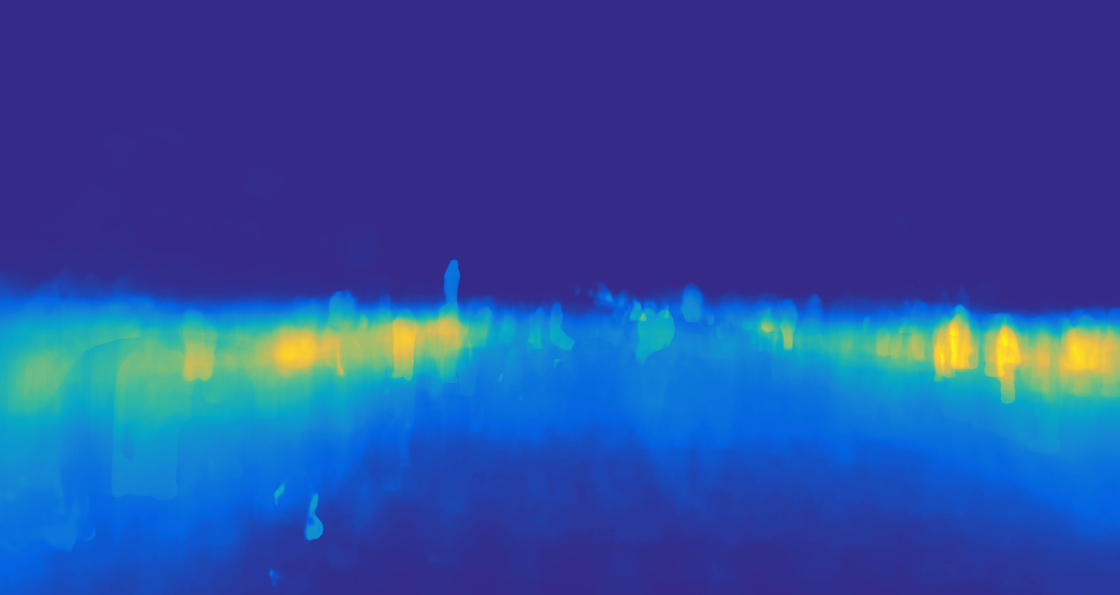}}\end{minipage} & \begin{minipage}[b]{0.44\columnwidth}\centering\raisebox{-.4\height}{\includegraphics[width=\linewidth]{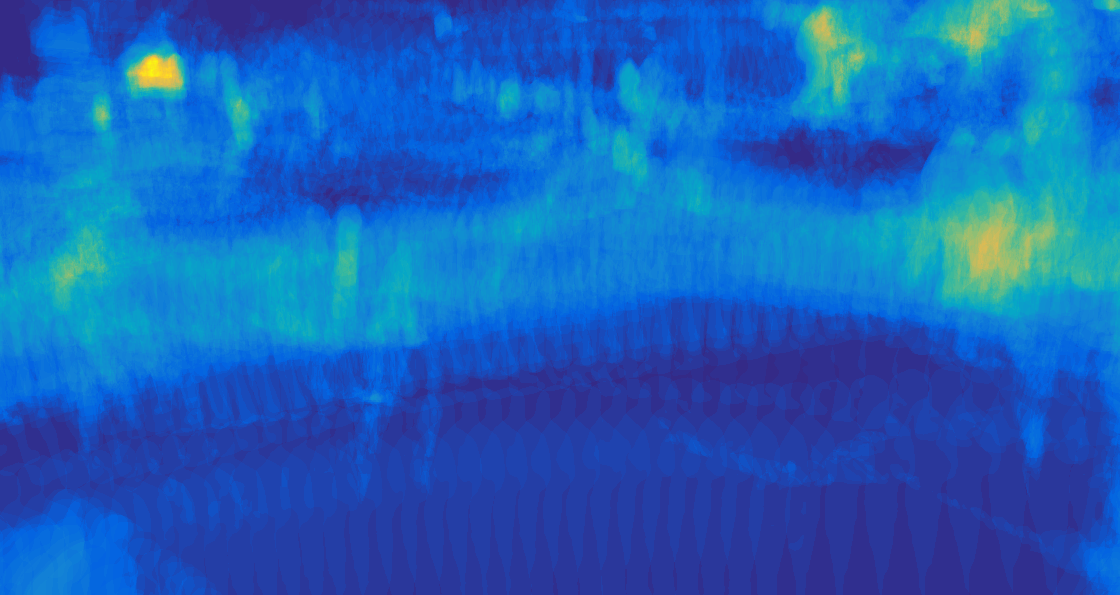}}\end{minipage} & \begin{minipage}[b]{0.44\columnwidth}\centering\raisebox{-.4\height}{\includegraphics[width=\linewidth]{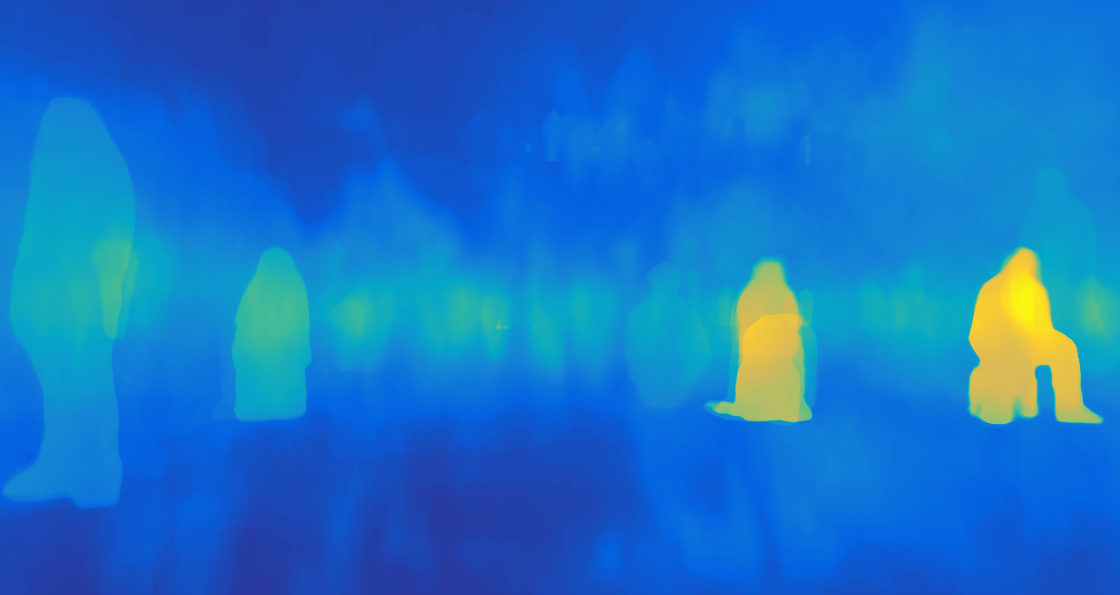}}\end{minipage}
    \\\midrule
    $12$ & \texttt{rider} & dynamic & \begin{minipage}[b]{0.44\columnwidth}\centering\raisebox{-.4\height}{\includegraphics[width=\linewidth]{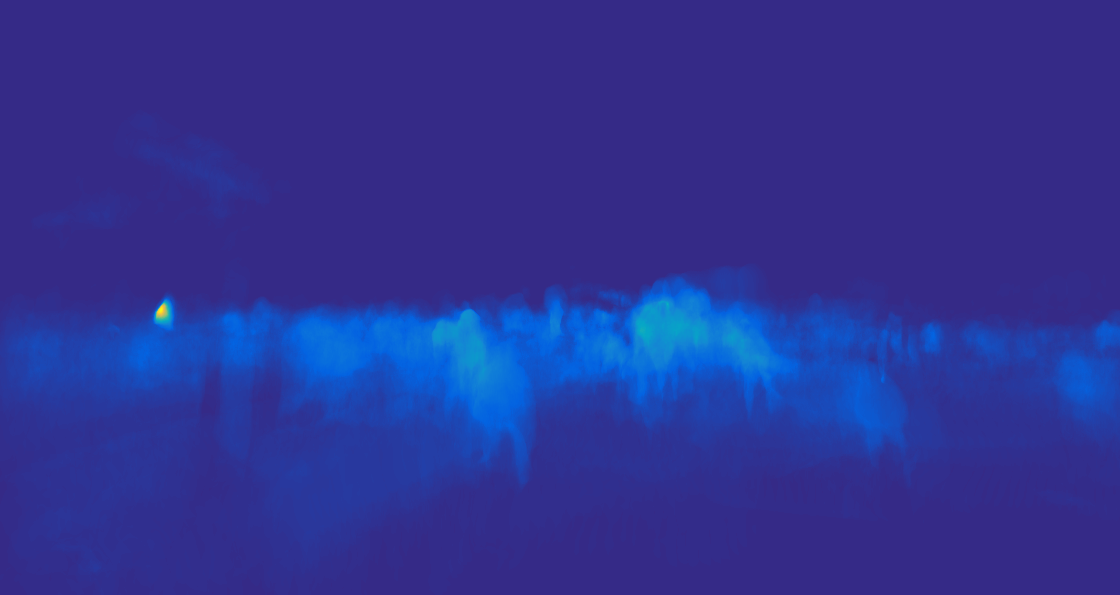}}\end{minipage} & \begin{minipage}[b]{0.44\columnwidth}\centering\raisebox{-.4\height}{\includegraphics[width=\linewidth]{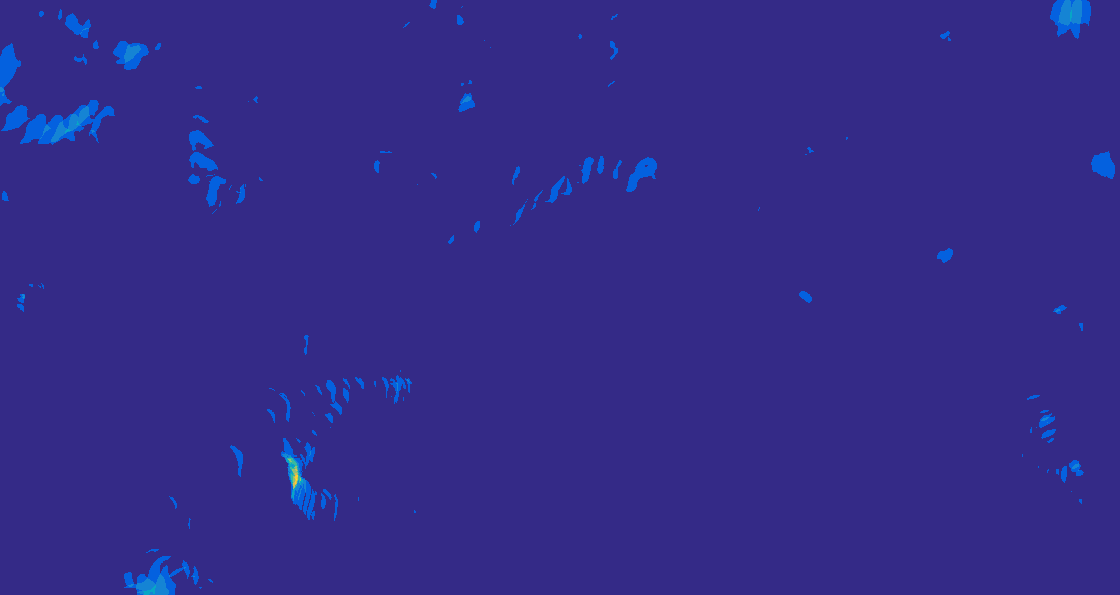}}\end{minipage} & \begin{minipage}[b]{0.44\columnwidth}\centering\raisebox{-.4\height}{\includegraphics[width=\linewidth]{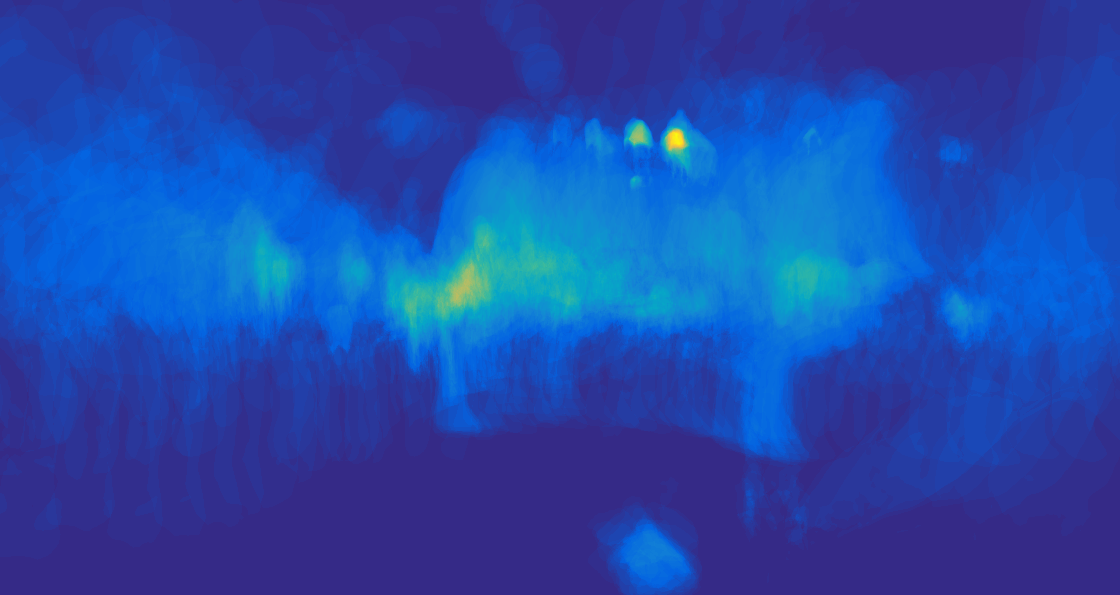}}\end{minipage}
    \\\midrule
    $13$ & \texttt{car} & dynamic & \begin{minipage}[b]{0.44\columnwidth}\centering\raisebox{-.4\height}{\includegraphics[width=\linewidth]{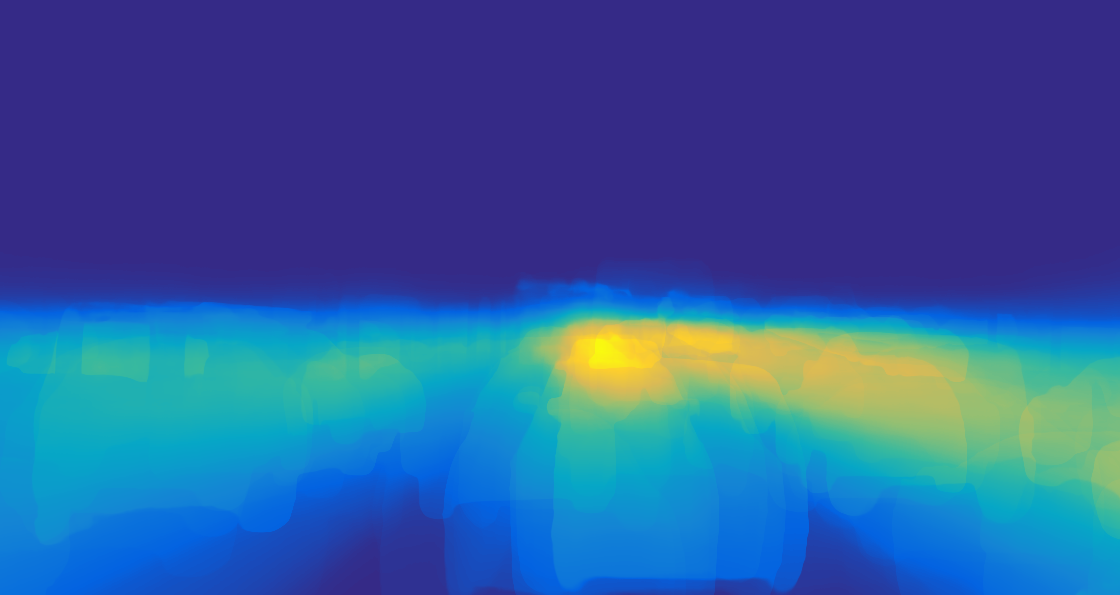}}\end{minipage} & \begin{minipage}[b]{0.44\columnwidth}\centering\raisebox{-.4\height}{\includegraphics[width=\linewidth]{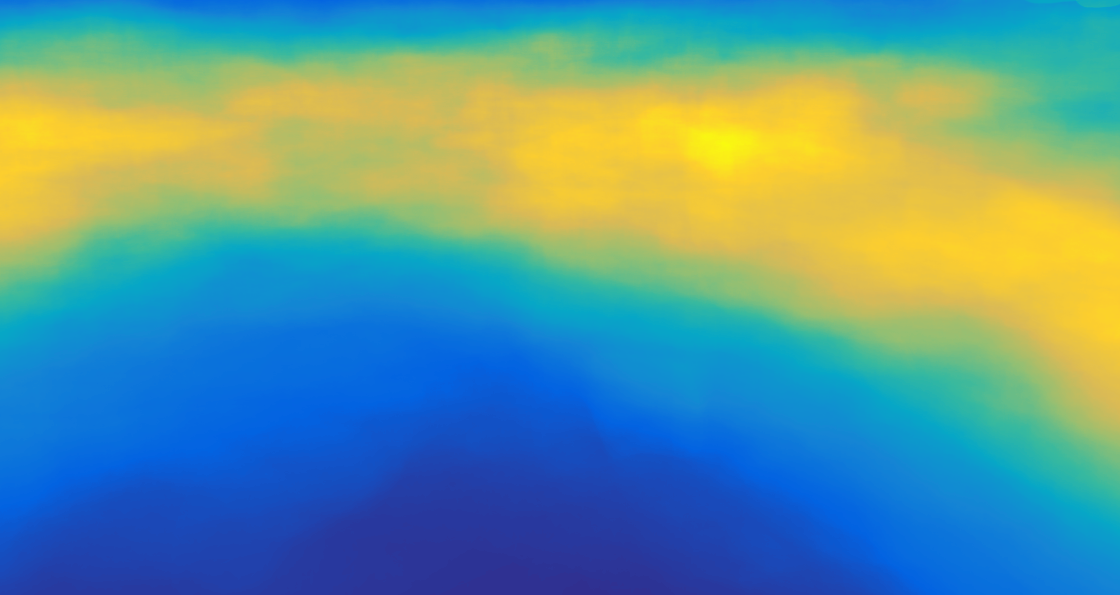}}\end{minipage} & \begin{minipage}[b]{0.44\columnwidth}\centering\raisebox{-.4\height}{\includegraphics[width=\linewidth]{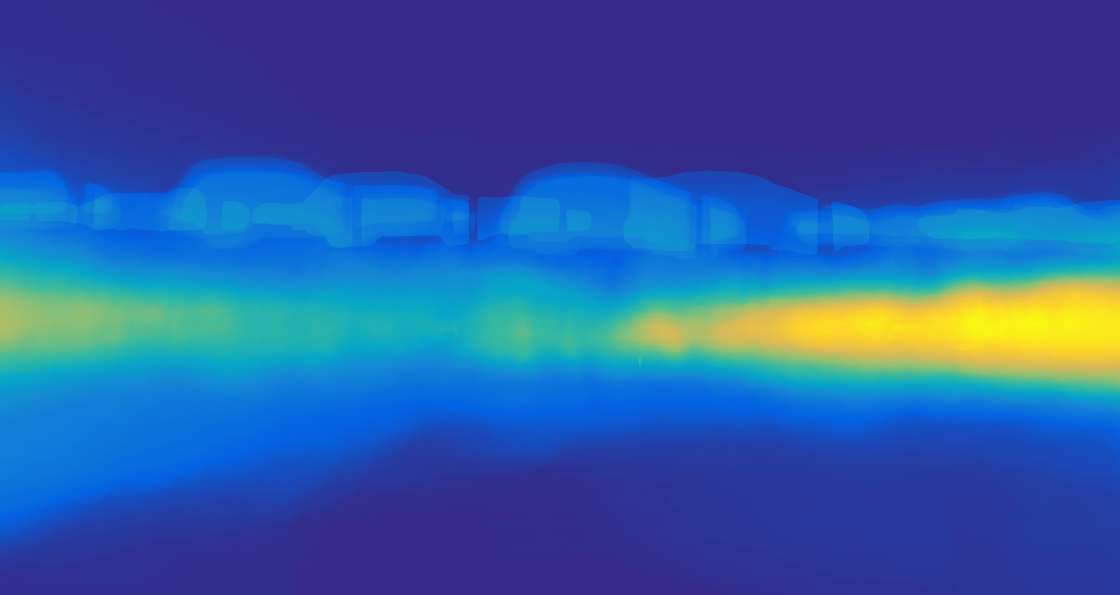}}\end{minipage}
    \\\midrule
    $14$ & \texttt{truck} & dynamic & \begin{minipage}[b]{0.44\columnwidth}\centering\raisebox{-.4\height}{\includegraphics[width=\linewidth]{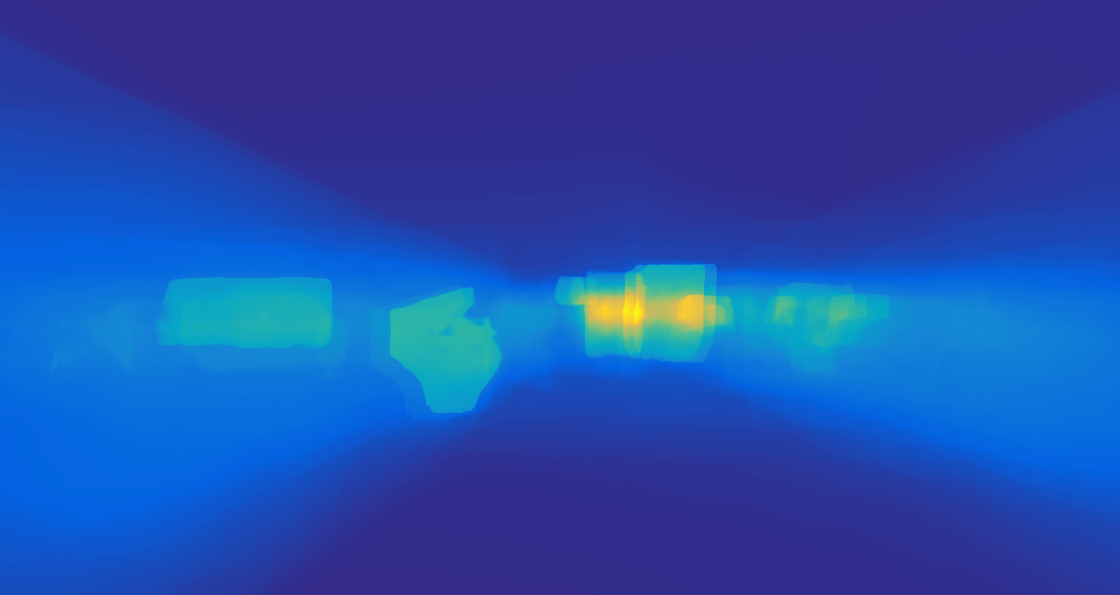}}\end{minipage} & \begin{minipage}[b]{0.44\columnwidth}\centering\raisebox{-.4\height}{\includegraphics[width=\linewidth]{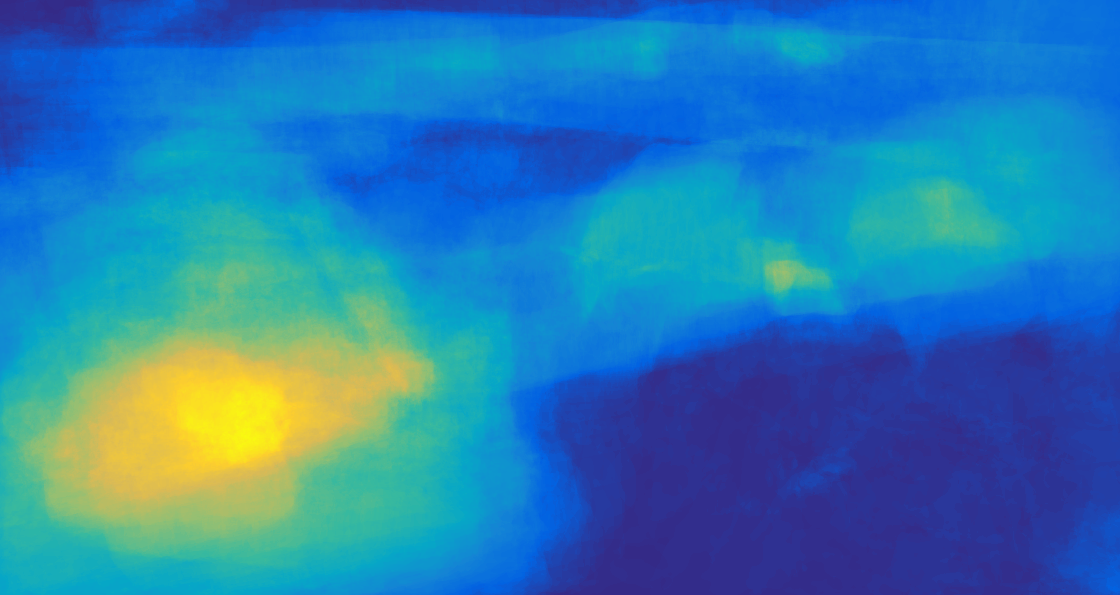}}\end{minipage} & \begin{minipage}[b]{0.44\columnwidth}\centering\raisebox{-.4\height}{\includegraphics[width=\linewidth]{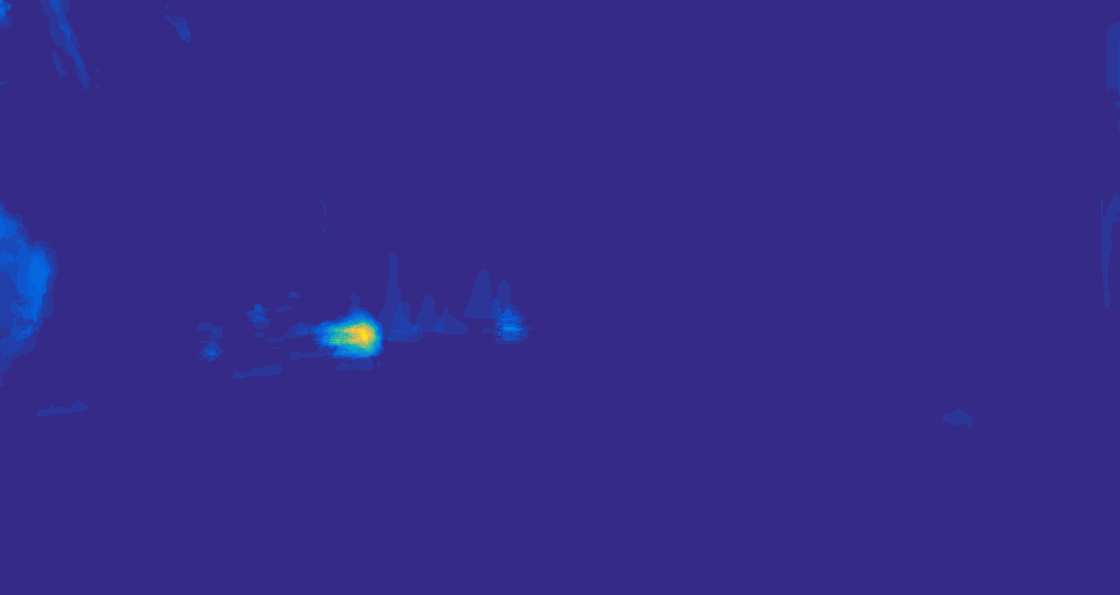}}\end{minipage}
    \\\midrule
    $15$ & \texttt{bus} & dynamic & \begin{minipage}[b]{0.44\columnwidth}\centering\raisebox{-.4\height}{\includegraphics[width=\linewidth]{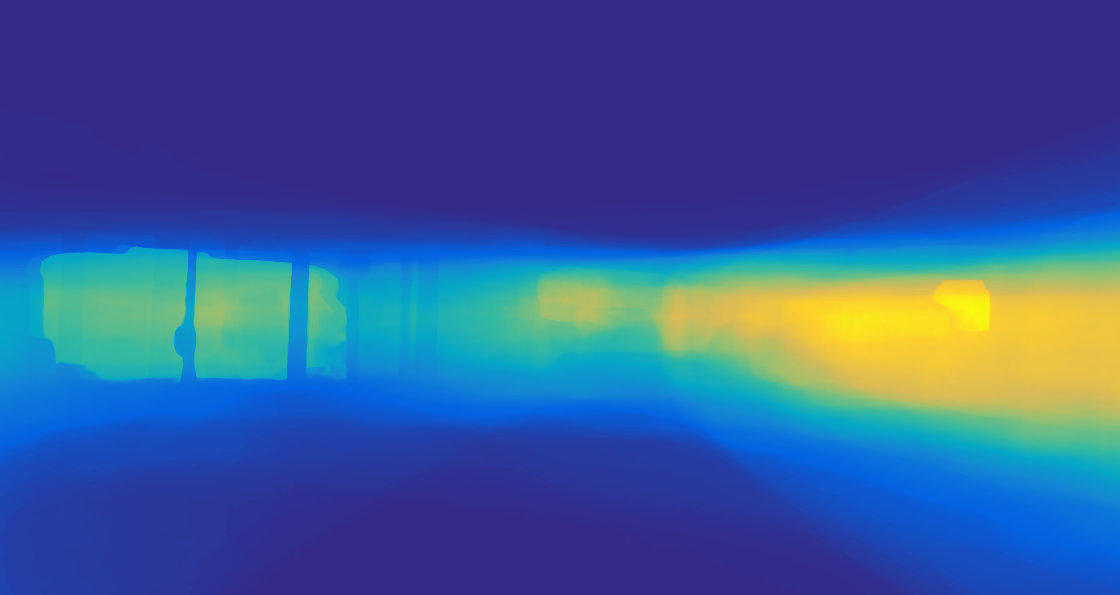}}\end{minipage} & \begin{minipage}[b]{0.44\columnwidth}\centering\raisebox{-.4\height}{\includegraphics[width=\linewidth]{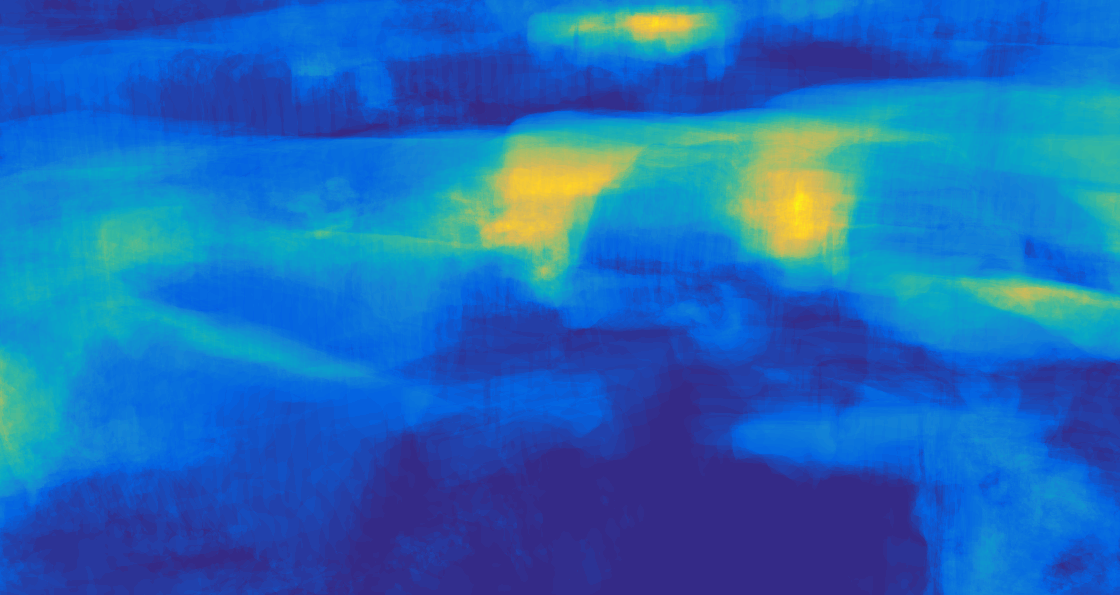}}\end{minipage} & \begin{minipage}[b]{0.44\columnwidth}\centering\raisebox{-.4\height}{\includegraphics[width=\linewidth]{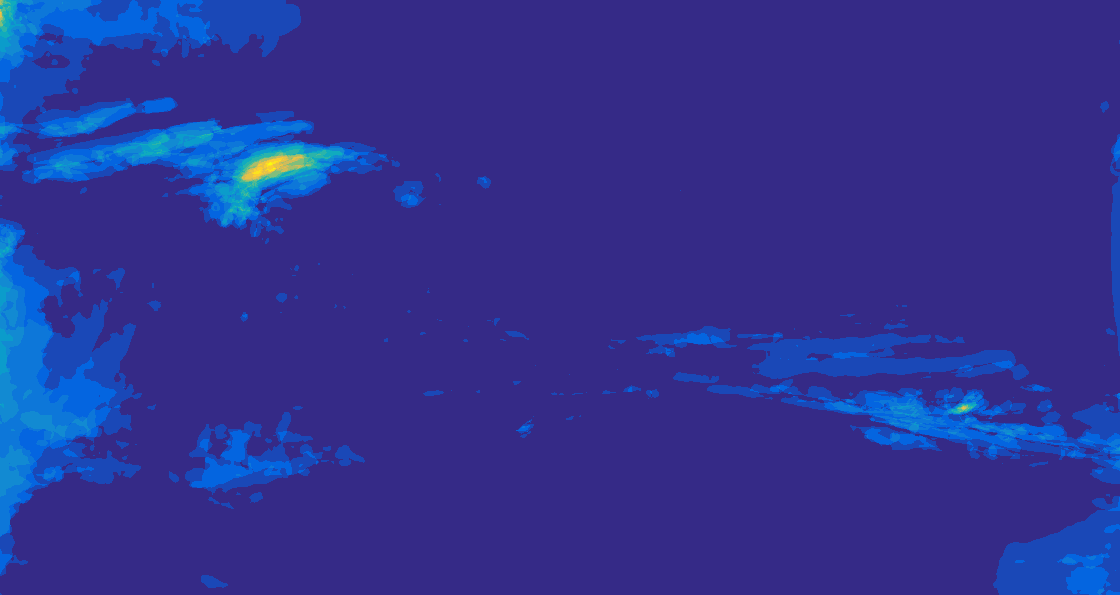}}\end{minipage}
    \\\midrule
    $16$ & \texttt{train} & dynamic & \begin{minipage}[b]{0.44\columnwidth}\centering\raisebox{-.4\height}{\includegraphics[width=\linewidth]{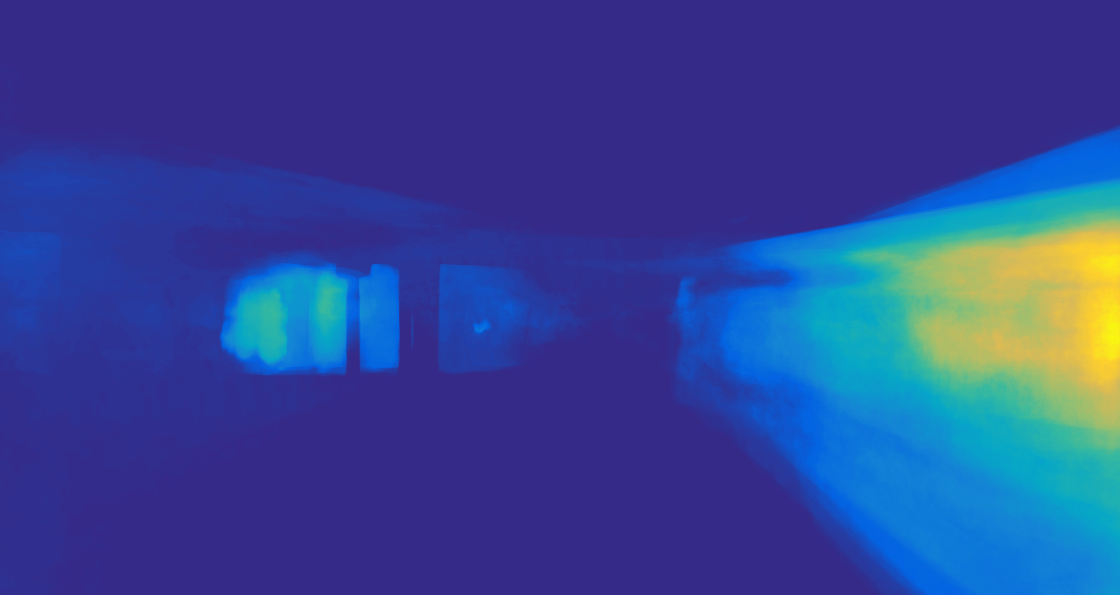}}\end{minipage} & \begin{minipage}[b]{0.44\columnwidth}\centering\raisebox{-.4\height}{\includegraphics[width=\linewidth]{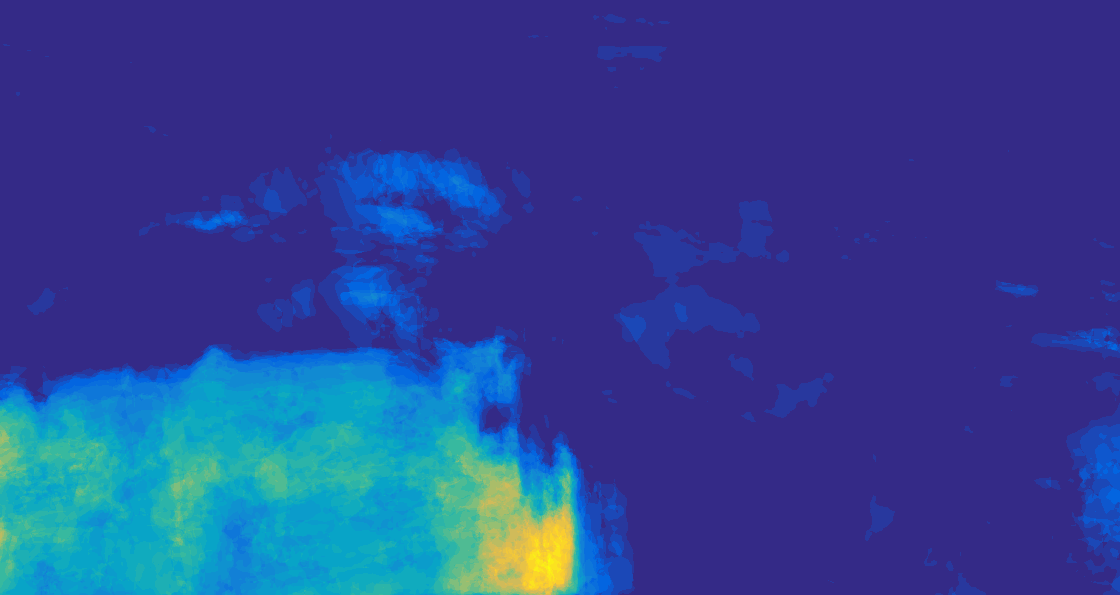}}\end{minipage} & \begin{minipage}[b]{0.44\columnwidth}\centering\raisebox{-.4\height}{\includegraphics[width=\linewidth]{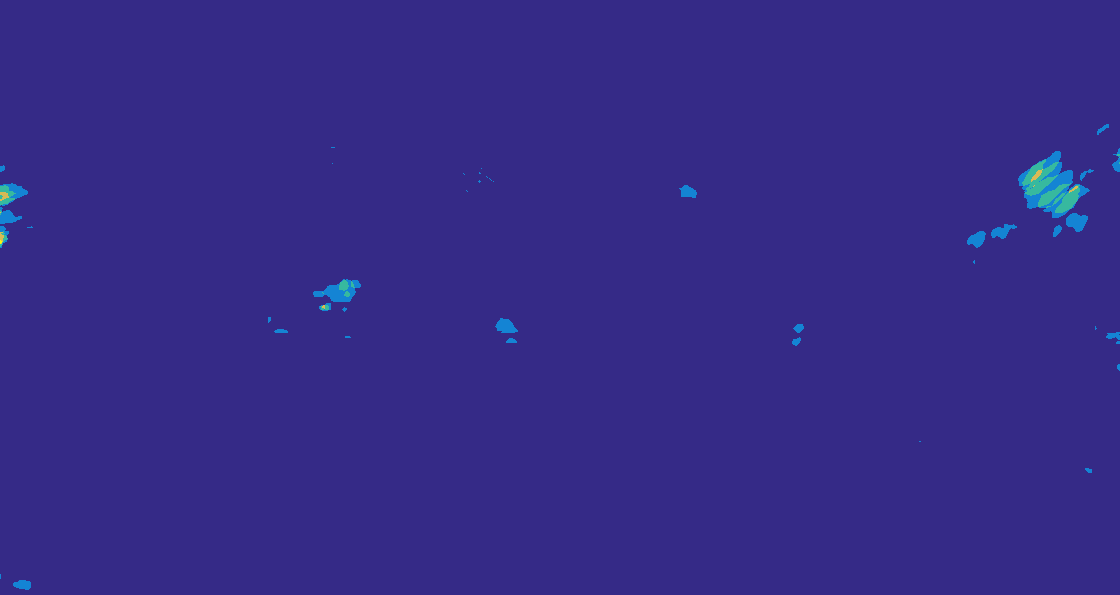}}\end{minipage}
    \\\midrule
    $17$ & \texttt{motorcycle} & dynamic & \begin{minipage}[b]{0.44\columnwidth}\centering\raisebox{-.4\height}{\includegraphics[width=\linewidth]{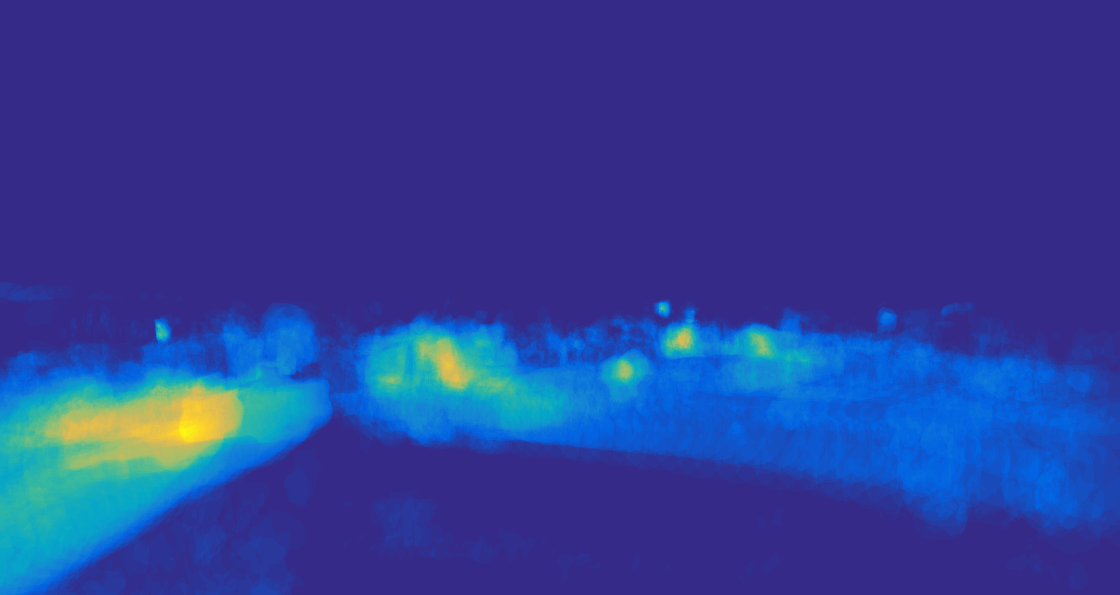}}\end{minipage} & \begin{minipage}[b]{0.44\columnwidth}\centering\raisebox{-.4\height}{\includegraphics[width=\linewidth]{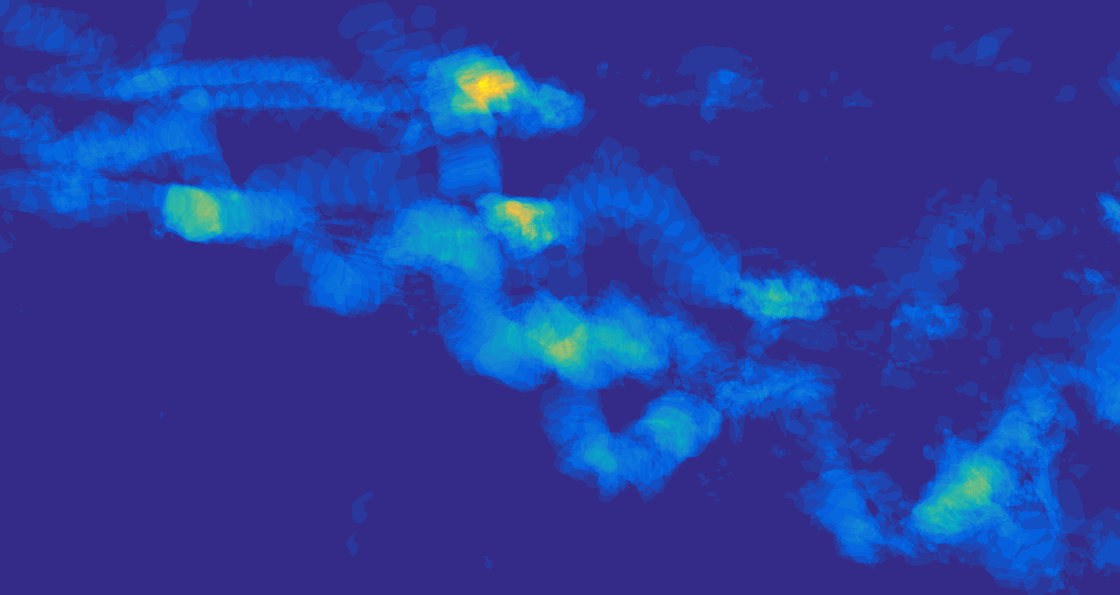}}\end{minipage} & \begin{minipage}[b]{0.44\columnwidth}\centering\raisebox{-.4\height}{\includegraphics[width=\linewidth]{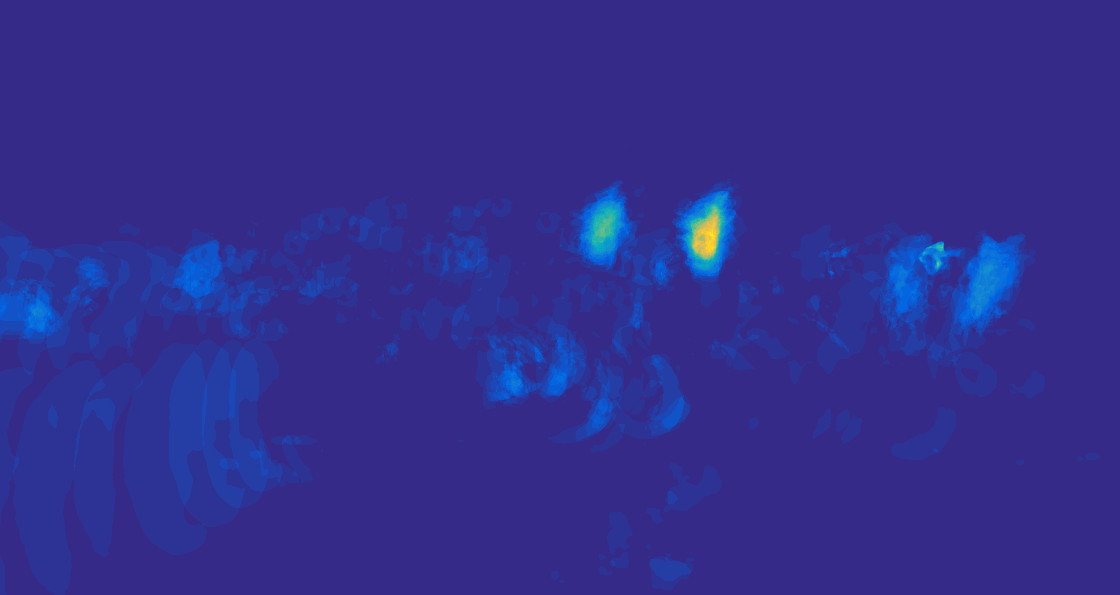}}\end{minipage}
    \\\midrule
    $18$ & \texttt{bicycle} & dynamic & \begin{minipage}[b]{0.44\columnwidth}\centering\raisebox{-.4\height}{\includegraphics[width=\linewidth]{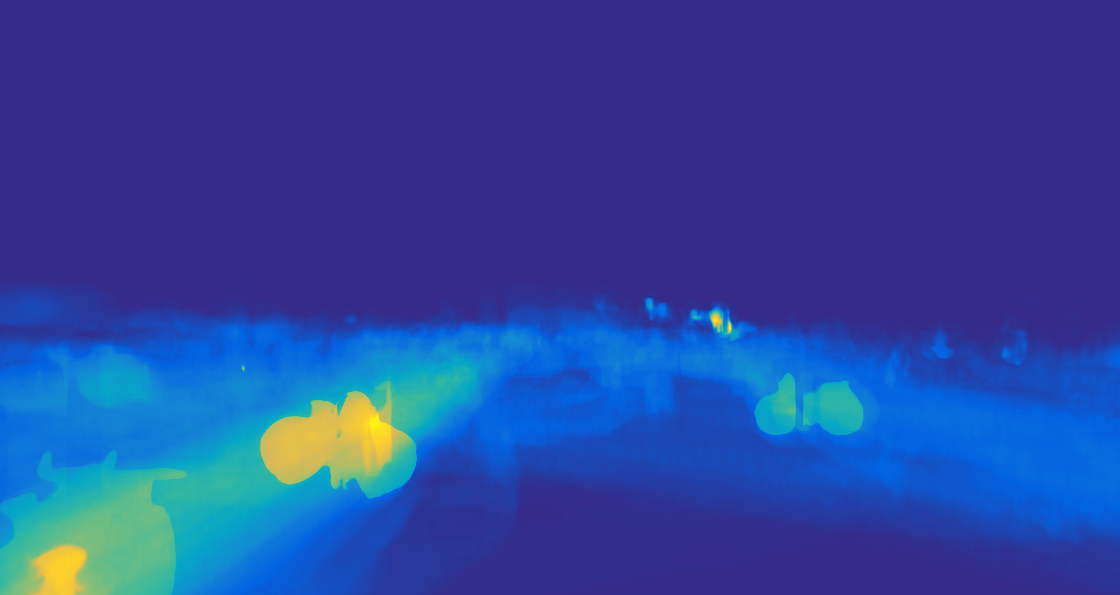}}\end{minipage} & \begin{minipage}[b]{0.44\columnwidth}\centering\raisebox{-.4\height}{\includegraphics[width=\linewidth]{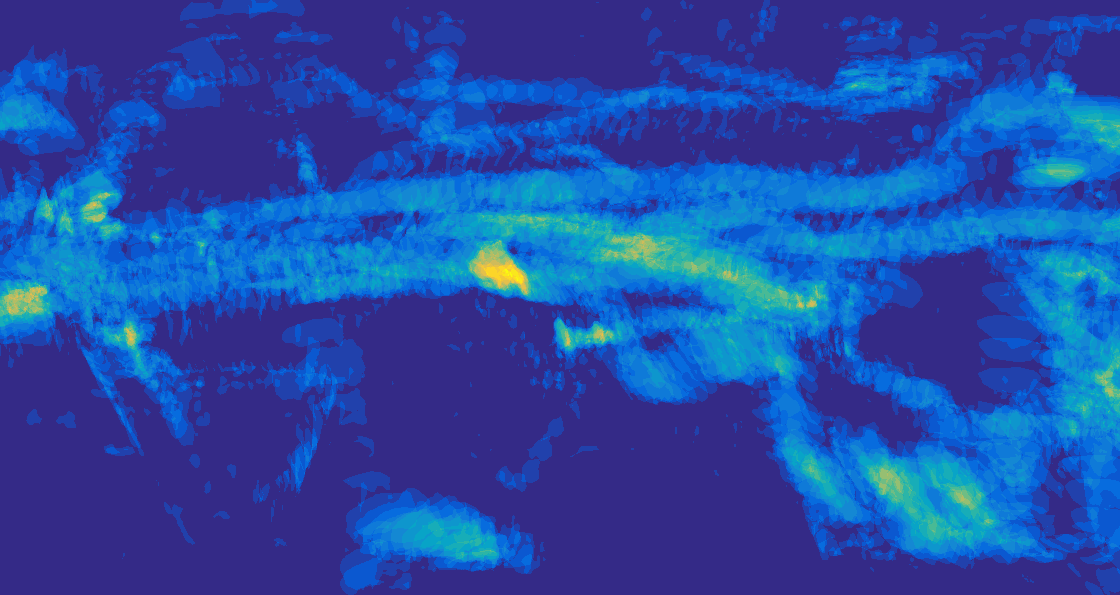}}\end{minipage} & \begin{minipage}[b]{0.44\columnwidth}\centering\raisebox{-.4\height}{\includegraphics[width=\linewidth]{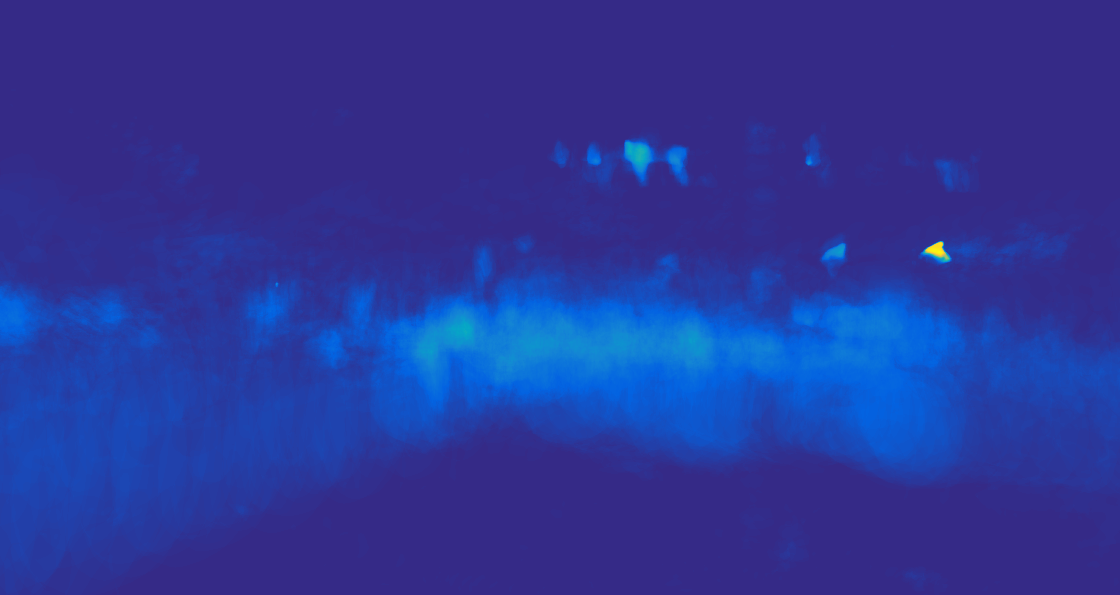}}\end{minipage}
    \\\bottomrule
\end{tabular}}
\label{tab:prior_dynamic}
\end{table*}

\subsection{Class Distribution Statistics}

As discussed in \cref{sec_supp:bench_definition} and \cref{sec_supp:platform_statisctis}, the same semantic class exhibits notable discrepancies across the three platforms, influenced by their unique perspectives, motion dynamics, and environmental contexts. Such discrepancies emphasize the need for spatial priors, as formulated in EAP, to account for platform-specific variations.

For example, the class \texttt{road} dominates the \textit{drone} platform ($45.75\%$) due to its high-altitude perspective capturing extensive ground-level surfaces, while in \textit{vehicle} ($21.94\%$) and \textit{quadruped} ($15.42\%$) platforms, this class appears more localized. Dynamic classes such as \texttt{car} and \texttt{person} show higher prominence in the \textit{vehicle} platform, consistent with its traffic-oriented scenarios, while being less frequent in \textit{drone} and \textit{quadruped} data due to limited proximity and perspectives for capturing such objects. Static classes like \texttt{vegetation} and \texttt{building} exhibit significant variation in coverage due to platform-specific viewpoints, with \textit{drone} capturing broader fields of view compared to the ground-level perspectives of \textit{vehicle} and \textit{quadruped}.

These statistics reinforce the hypothesis that leveraging spatial priors informed by class-specific activation patterns can significantly enhance cross-platform adaptation.

\subsection{Class Distribution Maps}

\cref{tab:prior_static} and \cref{tab:prior_dynamic} present the activation proportions for static and dynamic classes, respectively, across the \textit{vehicle}, \textit{drone}, and \textit{quadruped} platforms. These heatmaps reveal distinct spatial coverage and density patterns for each platform, which serve as the foundation for the proposed EAP. These tables highlight the following key observations:
\begin{itemize}
    \item In the \textit{vehicle} platform, the \texttt{road} class is highly concentrated in the lower-central region, reflecting the ground-level perspective. In contrast, \textit{drone} exhibits a broader, more evenly distributed pattern due to its high-altitude viewpoint capturing expansive ground surfaces. The \textit{quadruped} platform shows a localized, narrower distribution, aligning with its lower vantage point.

    \item The \textit{vehicle} platform exhibits dense, vertically structured priors for \texttt{building}, consistent with urban driving scenarios. Meanwhile, \textit{drone} and \textit{quadruped} display sparser coverage, with \textit{drone} capturing larger landscape-level structures and \textit{quadruped} focusing on closer, localized regions. A similar pattern applies to some \texttt{car} classes, such as \texttt{bus}, \texttt{train}, \texttt{motorcycle}, and \texttt{bicycle}.

    \item The \texttt{pole} and \texttt{traffic-light} classes are distinctly prominent in the \textit{vehicle} platform due to urban driving environments. The \textit{drone} platform shows certain occurrences, while the \textit{quadruped} platform captures sporadic patterns that align with its lower viewpoint.

    \item For majority classes, such as \texttt{vegetation}, \texttt{terrain}, and \texttt{sky}, the spatial distribution for \textit{vehicle} and \textit{drone} is broader and denser, reflecting outdoor scenarios with natural elements. The \textit{quadruped} platform captures localized \texttt{vegetation} mainly from the upper half of the field of view, often in close proximity to its route.
\end{itemize}

These heatmaps demonstrate the inherent semantic and spatial discrepancies across platforms, highlighting the necessity of incorporating spatial priors into the cross-platform adaptation process. By leveraging these platform-specific semantic distributions, the EAP enables more confident and domain-aligned predictions, ensuring effective adaptation across diverse operational contexts.

\begin{table*}[t]
    \centering
    \caption{
        \textbf{The event-triggered activation maps} among the \includegraphics[width=0.018\linewidth]{figures/icons/vehicle.png}~vehicle ($\mathcal{P}^{\textcolor{fly_green}{\mathbf{v}}}$), \includegraphics[width=0.021\linewidth]{figures/icons/drone.png}~drone ($\mathcal{P}^{\textcolor{fly_red}{\mathbf{d}}}$), and \includegraphics[width=0.019\linewidth]{figures/icons/quadruped.png}~quadruped ($\mathcal{P}^{\textcolor{fly_blue}{\mathbf{q}}}$) platforms, respectively, in the proposed \textbf{\textit{\textcolor{fly_green}{E}\textcolor{fly_red}{X}\textcolor{fly_blue}{Po}}} benchmark. The brighter the color, the higher the probability of occurrences. Best viewed in colors.}
    \vspace{-0.2cm}
    \resizebox{\linewidth}{!}{
    \begin{tabular}{c|c|c|c}
    \toprule
    \multirow{2}{*}{\textbf{Class}} & \includegraphics[width=0.024\linewidth]{figures/icons/vehicle.png} & \includegraphics[width=0.03\linewidth]{figures/icons/drone.png} & \includegraphics[width=0.024\linewidth]{figures/icons/quadruped.png}
    \\
    & \textbf{\textcolor{fly_green}{vehicle}} ($\mathcal{P}^{\textcolor{fly_green}{\mathbf{v}}}$) & \textbf{\textcolor{fly_red}{drone}} ($\mathcal{P}^{\textcolor{fly_red}{\mathbf{d}}}$) & \textbf{\textcolor{fly_blue}{quadruped}} ($\mathcal{P}^{\textcolor{fly_blue}{\mathbf{q}}}$)
    \\\midrule\midrule
    \texttt{road} & \begin{minipage}[b]{0.66\columnwidth}\centering\raisebox{-.5\height}{\includegraphics[width=\linewidth]{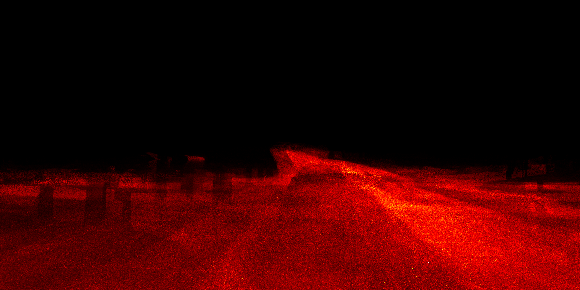}}\end{minipage} & \begin{minipage}[b]{0.66\columnwidth}\centering\raisebox{-.5\height}{\includegraphics[width=\linewidth]{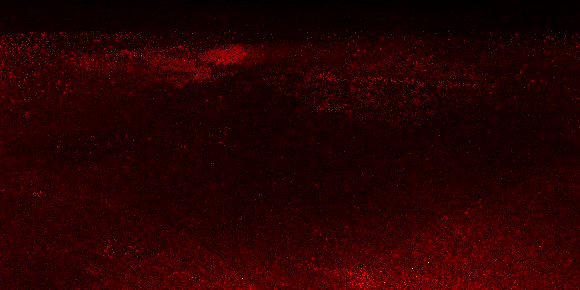}}\end{minipage} & \begin{minipage}[b]{0.66\columnwidth}\centering\raisebox{-.5\height}{\includegraphics[width=\linewidth]{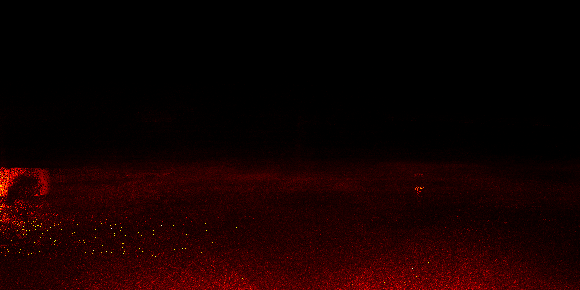}}\end{minipage}
    \\\midrule
    \texttt{building} & \begin{minipage}[b]{0.66\columnwidth}\centering\raisebox{-.5\height}{\includegraphics[width=\linewidth]{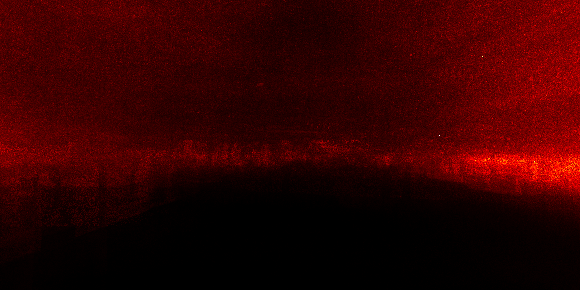}}\end{minipage} & \begin{minipage}[b]{0.66\columnwidth}\centering\raisebox{-.5\height}{\includegraphics[width=\linewidth]{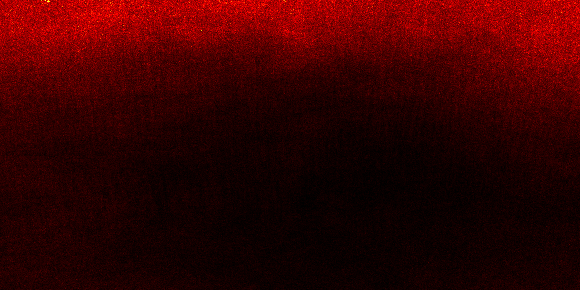}}\end{minipage} & \begin{minipage}[b]{0.66\columnwidth}\centering\raisebox{-.5\height}{\includegraphics[width=\linewidth]{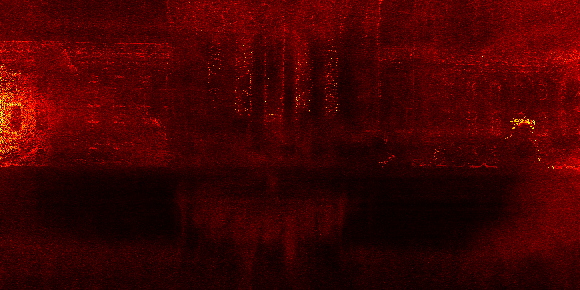}}\end{minipage}
    \\\midrule
    \texttt{car} & \begin{minipage}[b]{0.66\columnwidth}\centering\raisebox{-.5\height}{\includegraphics[width=\linewidth]{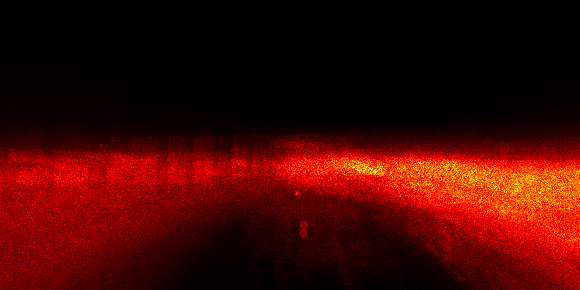}}\end{minipage} & \begin{minipage}[b]{0.66\columnwidth}\centering\raisebox{-.5\height}{\includegraphics[width=\linewidth]{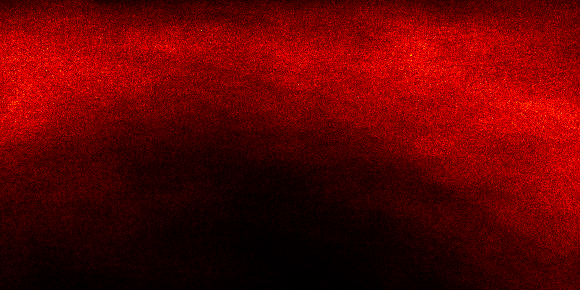}}\end{minipage} & \begin{minipage}[b]{0.66\columnwidth}\centering\raisebox{-.5\height}{\includegraphics[width=\linewidth]{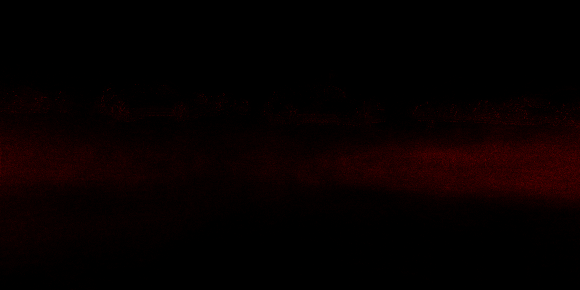}}\end{minipage}
    \\\midrule
    \texttt{fence} & \begin{minipage}[b]{0.66\columnwidth}\centering\raisebox{-.5\height}{\includegraphics[width=\linewidth]{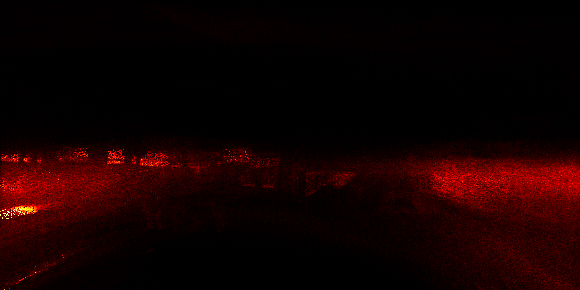}}\end{minipage} & \begin{minipage}[b]{0.66\columnwidth}\centering\raisebox{-.5\height}{\includegraphics[width=\linewidth]{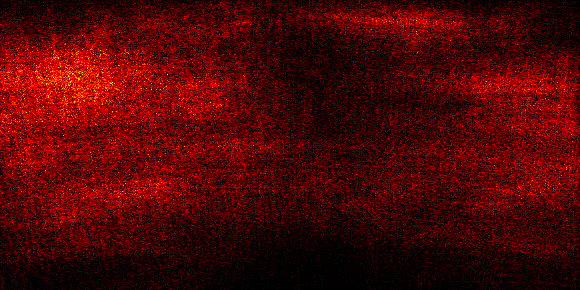}}\end{minipage} & \begin{minipage}[b]{0.66\columnwidth}\centering\raisebox{-.5\height}{\includegraphics[width=\linewidth]{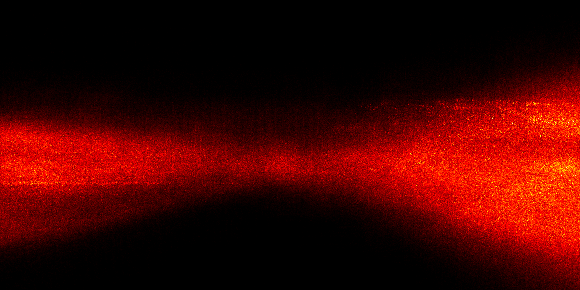}}\end{minipage}
    \\\midrule
    \texttt{pole} & \begin{minipage}[b]{0.66\columnwidth}\centering\raisebox{-.5\height}{\includegraphics[width=\linewidth]{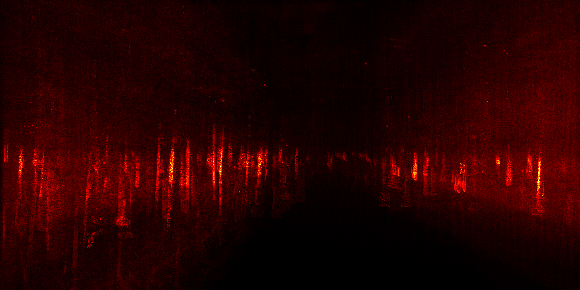}}\end{minipage} & \begin{minipage}[b]{0.66\columnwidth}\centering\raisebox{-.5\height}{\includegraphics[width=\linewidth]{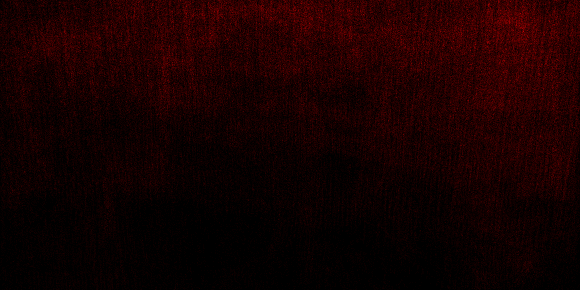}}\end{minipage} & \begin{minipage}[b]{0.66\columnwidth}\centering\raisebox{-.5\height}{\includegraphics[width=\linewidth]{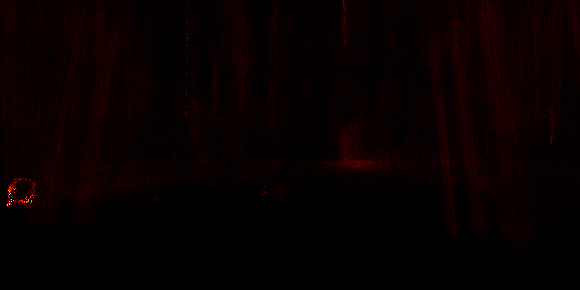}}\end{minipage}
    \\\midrule
    \texttt{sidewalk} & \begin{minipage}[b]{0.66\columnwidth}\centering\raisebox{-.5\height}{\includegraphics[width=\linewidth]{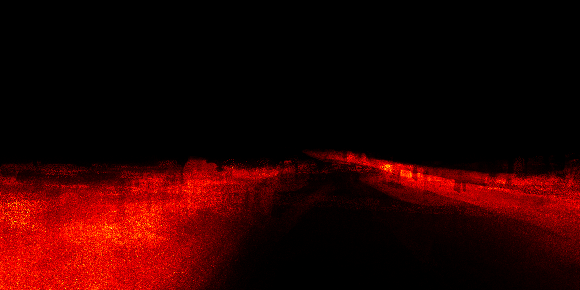}}\end{minipage} & \begin{minipage}[b]{0.66\columnwidth}\centering\raisebox{-.5\height}{\includegraphics[width=\linewidth]{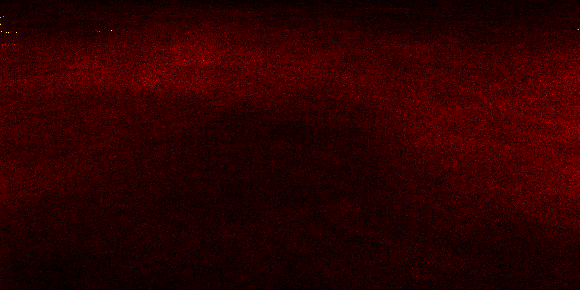}}\end{minipage} & \begin{minipage}[b]{0.66\columnwidth}\centering\raisebox{-.5\height}{\includegraphics[width=\linewidth]{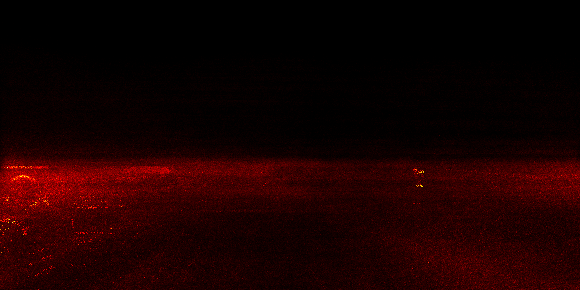}}\end{minipage}
    \\\midrule
    \texttt{vegetation} & \begin{minipage}[b]{0.66\columnwidth}\centering\raisebox{-.5\height}{\includegraphics[width=\linewidth]{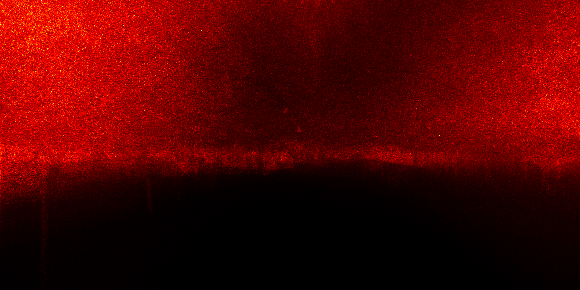}}\end{minipage} & \begin{minipage}[b]{0.66\columnwidth}\centering\raisebox{-.5\height}{\includegraphics[width=\linewidth]{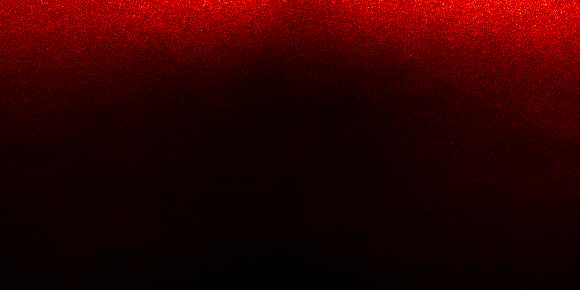}}\end{minipage} & \begin{minipage}[b]{0.66\columnwidth}\centering\raisebox{-.5\height}{\includegraphics[width=\linewidth]{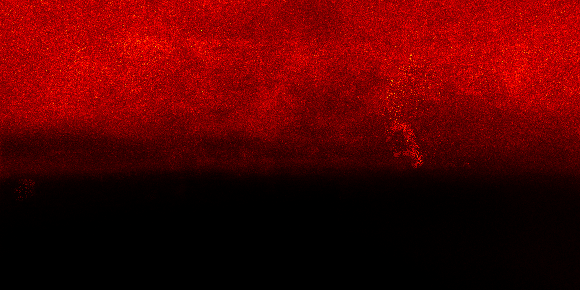}}\end{minipage}
    \\\midrule
    \texttt{wall} & \begin{minipage}[b]{0.66\columnwidth}\centering\raisebox{-.5\height}{\includegraphics[width=\linewidth]{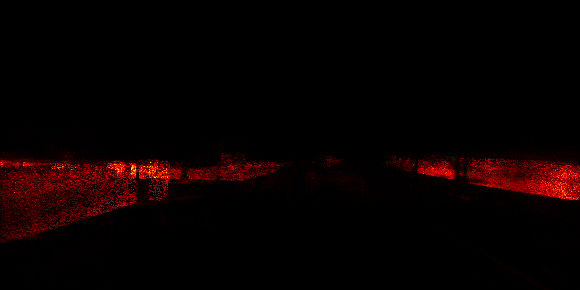}}\end{minipage} & \begin{minipage}[b]{0.66\columnwidth}\centering\raisebox{-.5\height}{\includegraphics[width=\linewidth]{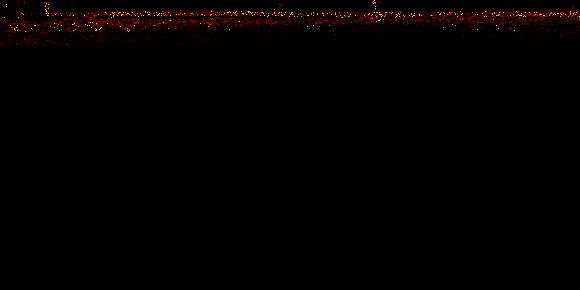}}\end{minipage} & \begin{minipage}[b]{0.66\columnwidth}\centering\raisebox{-.5\height}{\includegraphics[width=\linewidth]{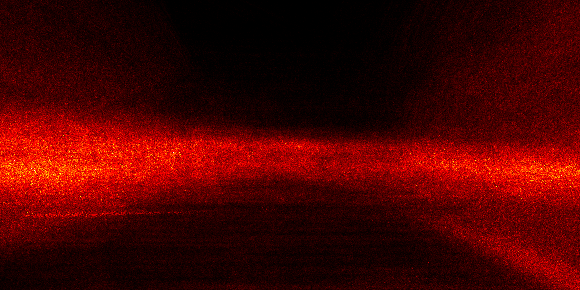}}\end{minipage}
    \\\bottomrule
\end{tabular}}
\label{tab:activation_maps}
\end{table*}

\subsection{Event-Triggered Activation Maps}
Our EAP-driven event data mixing technique builds on the assumption that event-triggered activations are closely linked to semantic distributions, as these activations reflect dynamic and structural changes captured by event cameras. To validate this assumption, we calculate probability maps of event-triggered activations for all semantic classes and present the results in \cref{tab:activation_maps}.

These maps reveal a striking correlation between event-triggered activations and semantic class distributions. Specifically, the event-triggered activations in static classes such as \texttt{road}, \texttt{building}, and \texttt{vegetation} demonstrate strong spatial consistency across platforms. For example, in the \textit{vehicle} platform, \texttt{road} activations are concentrated in the lower-central region, reflecting the expected viewpoint of ground-level sensors. Similarly, \texttt{building} activations align vertically, consistent with urban environments. This correlation underscores the utility of EAP in capturing spatially consistent priors for static classes.

For dynamic classes such as \texttt{car}, activations are more sporadic but still exhibit platform-specific patterns. The \textit{vehicle} platform shows dense activations in traffic-heavy areas, while the \textit{drone} platform captures broader distributions due to its high-altitude perspective. The \textit{quadruped} platform highlights localized activations near dynamic objects encountered in its immediate surroundings.

These observations reinforce the premise of EAP: that leveraging platform-specific activation patterns can guide adaptation by aligning predictions with the unique event-triggered dynamics of each platform. By incorporating these patterns into the adaptation process, EAP enhances confidence in predictions, particularly for challenging classes or underrepresented regions.
\section{Additional Experiment Results}
In this section, we provide additional results from our comparative and ablation experiments to further demonstrate the effectiveness and superiority of the proposed \textbf{\textit{\textcolor{fly_green}{Event}\textcolor{fly_red}{Fly}}} framework.

\subsection{Class-Wise Adaptation Results}
In the main body of this paper, due to space limits, we provide only the class-wise cross-platform adaptation results for the \includegraphics[width=0.04\linewidth]{figures/icons/vehicle.png}~vehicle ($\mathcal{P}^{\textcolor{fly_green}{\mathbf{v}}}$) to \includegraphics[width=0.045\linewidth]{figures/icons/drone.png}~drone ($\mathcal{P}^{\textcolor{fly_red}{\mathbf{d}}}$) and the \includegraphics[width=0.04\linewidth]{figures/icons/vehicle.png}~vehicle ($\mathcal{P}^{\textcolor{fly_green}{\mathbf{v}}}$) to \includegraphics[width=0.042\linewidth]{figures/icons/quadruped.png}~quadruped ($\mathcal{P}^{\textcolor{fly_blue}{\mathbf{q}}}$) settings.

In this supplementary file, we further provide the cross-platform adaptation results from the following settings:
\begin{itemize}
    \item \cref{tab:drone2vehicle}: Adaptation from the \includegraphics[width=0.045\linewidth]{figures/icons/drone.png}~drone ($\mathcal{P}^{\textcolor{fly_red}{\mathbf{d}}}$) platform to the \includegraphics[width=0.04\linewidth]{figures/icons/vehicle.png}~vehicle ($\mathcal{P}^{\textcolor{fly_green}{\mathbf{v}}}$) platform.

    \item \cref{tab:drone2quadruped}: Adaptation from the \includegraphics[width=0.045\linewidth]{figures/icons/drone.png}~drone ($\mathcal{P}^{\textcolor{fly_red}{\mathbf{d}}}$) platform to the \includegraphics[width=0.042\linewidth]{figures/icons/quadruped.png}~quadruped ($\mathcal{P}^{\textcolor{fly_blue}{\mathbf{q}}}$) platform.

    \item \cref{tab:quadruped2vehicle}: Adaptation from the \includegraphics[width=0.042\linewidth]{figures/icons/quadruped.png}~quadruped ($\mathcal{P}^{\textcolor{fly_blue}{\mathbf{q}}}$) platform to the \includegraphics[width=0.04\linewidth]{figures/icons/vehicle.png}~vehicle ($\mathcal{P}^{\textcolor{fly_green}{\mathbf{v}}}$) platform.

    \item \cref{tab:quadruped2drone}: Adaptation from the \includegraphics[width=0.042\linewidth]{figures/icons/quadruped.png}~quadruped ($\mathcal{P}^{\textcolor{fly_blue}{\mathbf{q}}}$) platform to the \includegraphics[width=0.045\linewidth]{figures/icons/drone.png}~drone ($\mathcal{P}^{\textcolor{fly_red}{\mathbf{d}}}$) platform.
\end{itemize}

Across all adaptation settings, our framework consistently achieves the highest accuracy (Acc), mean accuracy (mAcc), and mean Intersection over Union (mIoU), demonstrating its robustness in adapting event-based perception across platforms. Notably, \textbf{\textit{\textcolor{fly_green}{Event}\textcolor{fly_red}{Fly}}} outperforms prior methods such as MIC \cite{hoyer2023mic} and PLSR \cite{zhao2024plsr} by significant margins, particularly in complex settings such as the adaptation from \includegraphics[width=0.045\linewidth]{figures/icons/drone.png}~drone ($\mathcal{P}^{\textcolor{fly_red}{\mathbf{d}}}$) to \includegraphics[width=0.042\linewidth]{figures/icons/quadruped.png}~quadruped ($\mathcal{P}^{\textcolor{fly_blue}{\mathbf{q}}}$), and from \includegraphics[width=0.042\linewidth]{figures/icons/quadruped.png}~quadruped ($\mathcal{P}^{\textcolor{fly_blue}{\mathbf{q}}}$) to \includegraphics[width=0.045\linewidth]{figures/icons/drone.png}~drone ($\mathcal{P}^{\textcolor{fly_red}{\mathbf{d}}}$).

\begin{table*}[t]
    \centering
    \caption{
        \textbf{Benchmark results of event camera cross-platform adaptation} from \includegraphics[width=0.021\linewidth]{figures/icons/drone.png}~drone ($\mathcal{P}^{\textcolor{fly_red}{\mathbf{d}}}$) to \includegraphics[width=0.018\linewidth]{figures/icons/vehicle.png}~vehicle ($\mathcal{P}^{\textcolor{fly_green}{\mathbf{v}}}$). \textcolor{gray}{Target} denotes the model is trained with ground truth from the target domain. All scores are given in percentage (\%). The \textcolor{fly_red}{second best} and \textcolor{fly_green}{best} scores under each evaluation metric are highlighted in \colorbox{fly_red!12}{\textcolor{fly_red}{red}} and \colorbox{fly_green!12}{\textcolor{fly_green}{green}} colors, respectively.}
    \vspace{-0.25cm}
    \resizebox{\linewidth}{!}{
    \begin{tabular}{r|p{26pt}<{\centering}p{26pt}<{\centering}p{26pt}<{\centering}p{26pt}<{\centering}|p{26pt}<{\centering}p{26pt}<{\centering}p{26pt}<{\centering}p{26pt}<{\centering}p{26pt}<{\centering}p{26pt}<{\centering}p{26pt}<{\centering}p{26pt}<{\centering}p{26pt}<{\centering}p{26pt}<{\centering}p{26pt}<{\centering}}
    \toprule
    \textbf{Method} & \textbf{Acc} & \textbf{mAcc} & \textbf{mIoU} & \textbf{fIoU} & \rotatebox{90}{\textcolor{background}{$\blacksquare$}~ground~} & \rotatebox{90}{\textcolor{building}{$\blacksquare$}~build} & \rotatebox{90}{\textcolor{fence}{$\blacksquare$}~fence} & \rotatebox{90}{\textcolor{person}{$\blacksquare$}~person} & \rotatebox{90}{\textcolor{pole}{$\blacksquare$}~pole} & \rotatebox{90}{\textcolor{road}{$\blacksquare$}~road} & \rotatebox{90}{\textcolor{sidewalk}{$\blacksquare$}~walk} & \rotatebox{90}{\textcolor{vegetation}{$\blacksquare$}~veg} & \rotatebox{90}{\textcolor{car}{$\blacksquare$}~car} & \rotatebox{90}{\textcolor{wall}{$\blacksquare$}~wall} & \rotatebox{90}{\textcolor{traffic-sign}{$\blacksquare$}~sign}
    \\\midrule\midrule
    \textcolor{gray}{Source-Only}~\textcolor{gray}{$\circ$} & \textcolor{gray}{$57.91$} & \textcolor{gray}{$29.97$} & \textcolor{gray}{$20.79$} & \textcolor{gray}{$11.72$} & \textcolor{gray}{$52.64$} & \textcolor{gray}{$30.04$} & \textcolor{gray}{$0.82$} & \textcolor{gray}{$0.35$} & \textcolor{gray}{$11.27$} & \textcolor{gray}{$46.96$} & \textcolor{gray}{$7.48$} & \textcolor{gray}{$46.51$} & \textcolor{gray}{$31.99$} & \textcolor{gray}{$0.00$} & \textcolor{gray}{$0.68$}
    \\\midrule
    AdaptSegNet \cite{tsai2018adaptsegnet} & $68.29$ & $39.99$ & $29.55$ & $13.72$ & $41.79$ & $56.46$ & $0.53$ & $2.80$ & $20.75$ & $68.86$ & $34.47$ & $58.42$ & $40.70$ & $0.00$ & $0.23$
    \\
    DACS \cite{tranheden2021dacs} & $71.78$ & $48.58$ & $36.10$ & $14.34$ & $47.65$ & $60.00$ & $0.00$ & \cellcolor{fly_red!12}$32.97$ & $23.57$ & $69.89$ & $37.76$ & $63.69$ & $43.45$ & \cellcolor{fly_green!12}$6.53$ & $11.60$
    \\
    MIC \cite{hoyer2023mic} & \cellcolor{fly_red!12}$72.46$ & $49.54$ & $36.88$ & $14.42$ & \cellcolor{fly_red!12}$48.15$ & $60.68$ & $0.00$ & $30.87$ & \cellcolor{fly_green!12}$24.95$ & $70.33$ & $39.47$ & \cellcolor{fly_red!12}$65.17$ & $44.51$ & \cellcolor{fly_red!12}$6.36$ & \cellcolor{fly_red!12}$15.21$
    \\
    PLSR \cite{zhao2024plsr} & \cellcolor{fly_red!12}$72.46$ & \cellcolor{fly_red!12}$49.84$ & \cellcolor{fly_red!12}$37.18$ & \cellcolor{fly_red!12}$14.45$ & $44.94$ & \cellcolor{fly_red!12}$62.15$ & \cellcolor{fly_red!12}$2.55$ & \cellcolor{fly_green!12}$35.60$ & \cellcolor{fly_red!12}$23.98$ & \cellcolor{fly_green!12}$72.59$ & \cellcolor{fly_red!12}$41.99$ & $61.18$ & \cellcolor{fly_green!12}$47.92$ & $3.87$ & $12.24$
    \\
    \textbf{\textcolor{fly_green}{Event}\textcolor{fly_red}{Fly}} \textbf{(Ours)} & \cellcolor{fly_green!12}$75.50$ & \cellcolor{fly_green!12}$52.90$ & \cellcolor{fly_green!12}$39.92$ & \cellcolor{fly_green!12}$15.08$ & \cellcolor{fly_green!12}$53.93$ & \cellcolor{fly_green!12}$65.14$ & \cellcolor{fly_green!12}$6.43$ & $31.61$ & $23.93$ & \cellcolor{fly_red!12}$72.18$ & \cellcolor{fly_green!12}$46.22$ & \cellcolor{fly_green!12}$68.68$ & \cellcolor{fly_red!12}$47.90$ & $4.12$ & \cellcolor{fly_green!12}$19.01$
    \\\midrule
    \cellcolor{gray!10}\textcolor{gray}{Target}~\textcolor{gray}{$\bullet$} & \cellcolor{gray!10}\textcolor{gray}{$86.12$} & \cellcolor{gray!10}\textcolor{gray}{$66.02$} & \cellcolor{gray!10}\textcolor{gray}{$55.93$} & \cellcolor{gray!10}\textcolor{gray}{$16.18$} & \cellcolor{gray!10}\textcolor{gray}{$87.07$} & \cellcolor{gray!10}\textcolor{gray}{$75.41$} & \cellcolor{gray!10}\textcolor{gray}{$22.70$} & \cellcolor{gray!10}\textcolor{gray}{$52.59$} & \cellcolor{gray!10}\textcolor{gray}{$39.41$} & \cellcolor{gray!10}\textcolor{gray}{$79.49$} & \cellcolor{gray!10}\textcolor{gray}{$58.82$} & \cellcolor{gray!10}\textcolor{gray}{$77.75$} & \cellcolor{gray!10}\textcolor{gray}{$69.63$} & \cellcolor{gray!10}\textcolor{gray}{$14.79$} & \cellcolor{gray!10}\textcolor{gray}{$37.61$}
    \\\bottomrule
    \end{tabular}}
\label{tab:drone2vehicle}
\end{table*}
\begin{table*}[t]
    \centering
    \caption{
        \textbf{Benchmark results of event camera cross-platform adaptation} from \includegraphics[width=0.021\linewidth]{figures/icons/drone.png}~drone ($\mathcal{P}^{\textcolor{fly_red}{\mathbf{d}}}$) to \includegraphics[width=0.019\linewidth]{figures/icons/quadruped.png}~quadruped ($\mathcal{P}^{\textcolor{fly_blue}{\mathbf{q}}}$). \textcolor{gray}{Target} denotes the model is trained with ground truth from the target domain. All scores are given in percentage (\%). The \textcolor{fly_red}{second best} and \textcolor{fly_green}{best} scores under each evaluation metric are highlighted in \colorbox{fly_red!12}{\textcolor{fly_red}{red}} and \colorbox{fly_green!12}{\textcolor{fly_green}{green}} colors, respectively.}
    \vspace{-0.25cm}
    \resizebox{\linewidth}{!}{
    \begin{tabular}{r|p{26pt}<{\centering}p{26pt}<{\centering}p{26pt}<{\centering}p{26pt}<{\centering}|p{26pt}<{\centering}p{26pt}<{\centering}p{26pt}<{\centering}p{26pt}<{\centering}p{26pt}<{\centering}p{26pt}<{\centering}p{26pt}<{\centering}p{26pt}<{\centering}p{26pt}<{\centering}p{26pt}<{\centering}p{26pt}<{\centering}}
    \toprule
    \textbf{Method} & \textbf{Acc} & \textbf{mAcc} & \textbf{mIoU} & \textbf{fIoU} & \rotatebox{90}{\textcolor{background}{$\blacksquare$}~ground~} & \rotatebox{90}{\textcolor{building}{$\blacksquare$}~build} & \rotatebox{90}{\textcolor{fence}{$\blacksquare$}~fence} & \rotatebox{90}{\textcolor{person}{$\blacksquare$}~person} & \rotatebox{90}{\textcolor{pole}{$\blacksquare$}~pole} & \rotatebox{90}{\textcolor{road}{$\blacksquare$}~road} & \rotatebox{90}{\textcolor{sidewalk}{$\blacksquare$}~walk} & \rotatebox{90}{\textcolor{vegetation}{$\blacksquare$}~veg} & \rotatebox{90}{\textcolor{car}{$\blacksquare$}~car} & \rotatebox{90}{\textcolor{wall}{$\blacksquare$}~wall} & \rotatebox{90}{\textcolor{traffic-sign}{$\blacksquare$}~sign}
    \\\midrule\midrule
    \textcolor{gray}{Source-Only}~\textcolor{gray}{$\circ$} & \textcolor{gray}{$66.83$} & \textcolor{gray}{$34.05$} & \textcolor{gray}{$23.06$} & \textcolor{gray}{$17.24$} & \textcolor{gray}{$59.62$} & \textcolor{gray}{$42.17$} & \textcolor{gray}{$2.76$} & \textcolor{gray}{$0.24$} & \textcolor{gray}{$8.20$} & \textcolor{gray}{$48.56$} & \textcolor{gray}{$8.55$} & \textcolor{gray}{$66.11$} & \textcolor{gray}{$17.12$} & \textcolor{gray}{$0.00$} & \textcolor{gray}{$0.27$}
    \\\midrule
    AdaptSegNet \cite{tsai2018adaptsegnet} & $67.57$ & $49.51$ & $33.99$ & $14.64$ & $42.75$ & $51.73$ & $33.04$ & $33.32$ & $14.33$ & \cellcolor{fly_green!12}$54.05$ & $19.71$ & $73.43$ & $20.56$ & $30.91$ & $0.00$
    \\
    DACS \cite{tranheden2021dacs} & $67.73$ & $51.73$ & $36.11$ & $14.49$ & $42.10$ & \cellcolor{fly_green!12}$55.10$ & \cellcolor{fly_green!12}$36.25$ & $34.55$ & $15.00$ & $50.45$ & \cellcolor{fly_red!12}$21.54$ & \cellcolor{fly_red!12}$75.77$ & $26.54$ & $39.87$ & $0.01$
    \\
    MIC \cite{hoyer2023mic} & $67.29$ & $50.91$ & \cellcolor{fly_red!12}$36.27$ & $14.53$ & \cellcolor{fly_red!12}$44.15$ & $51.15$ & \cellcolor{fly_red!12}$34.40$ & \cellcolor{fly_red!12}$37.99$ & $14.43$ & $45.74$ & \cellcolor{fly_green!12}$23.09$ & $75.38$ & $30.36$ & $41.41$ & \cellcolor{fly_red!12}$0.89$
    \\
    PLSR \cite{zhao2024plsr} & \cellcolor{fly_red!12}$67.83$ & \cellcolor{fly_red!12}$50.57$ & $36.21$ & \cellcolor{fly_red!12}$14.67$ & $42.62$ & \cellcolor{fly_red!12}$53.73$ & $30.80$ & $28.39$ & \cellcolor{fly_red!12}$15.70$ & $50.15$ & $20.94$ & \cellcolor{fly_green!12}$75.82$ & \cellcolor{fly_green!12}$36.48$ & \cellcolor{fly_green!12}$43.70$ & $0.00$
    \\
    \textbf{\textcolor{fly_green}{Event}\textcolor{fly_red}{Fly}} \textbf{(Ours)} & \cellcolor{fly_green!12}$69.68$ & \cellcolor{fly_green!12}$51.03$ & \cellcolor{fly_green!12}$37.37$ & \cellcolor{fly_green!12}$15.30$ & \cellcolor{fly_green!12}$44.92$ & $53.12$ & $34.16$ & \cellcolor{fly_green!12}$39.34$ & \cellcolor{fly_green!12}$16.95$ & \cellcolor{fly_red!12}$53.85$ & $17.59$ & $75.10$ & \cellcolor{fly_red!12}$33.03$ & \cellcolor{fly_red!12}$41.98$ & \cellcolor{fly_green!12}$0.97$
    \\\midrule
    \cellcolor{gray!10}\textcolor{gray}{Target}~\textcolor{gray}{$\bullet$} & \cellcolor{gray!10}\textcolor{gray}{$80.02$} & \cellcolor{gray!10}\textcolor{gray}{$60.55$} & \cellcolor{gray!10}\textcolor{gray}{$49.84$} & \cellcolor{gray!10}\textcolor{gray}{$19.58$} & \cellcolor{gray!10}\textcolor{gray}{$74.80$} & \cellcolor{gray!10}\textcolor{gray}{$56.23$} & \cellcolor{gray!10}\textcolor{gray}{$46.08$} & \cellcolor{gray!10}\textcolor{gray}{$55.28$} & \cellcolor{gray!10}\textcolor{gray}{$21.79$} & \cellcolor{gray!10}\textcolor{gray}{$59.90$} & \cellcolor{gray!10}\textcolor{gray}{$30.31$} & \cellcolor{gray!10}\textcolor{gray}{$77.24$} & \cellcolor{gray!10}\textcolor{gray}{$58.38$} & \cellcolor{gray!10}\textcolor{gray}{$62.47$} & \cellcolor{gray!10}\textcolor{gray}{$5.81$}
    \\\bottomrule
    \end{tabular}}
\label{tab:drone2quadruped}
\end{table*}
\begin{table*}[t]
    \centering
    \caption{
        \textbf{Benchmark results of event camera cross-platform adaptation} from \includegraphics[width=0.019\linewidth]{figures/icons/quadruped.png}~quadruped ($\mathcal{P}^{\textcolor{fly_blue}{\mathbf{q}}}$) to \includegraphics[width=0.018\linewidth]{figures/icons/vehicle.png}~vehicle ($\mathcal{P}^{\textcolor{fly_green}{\mathbf{v}}}$). \textcolor{gray}{Target} denotes the model is trained with ground truth from the target domain. All scores are given in percentage (\%). The \textcolor{fly_red}{second best} and \textcolor{fly_green}{best} scores under each evaluation metric are highlighted in \colorbox{fly_red!12}{\textcolor{fly_red}{red}} and \colorbox{fly_green!12}{\textcolor{fly_green}{green}} colors, respectively.}
    \vspace{-0.25cm}
    \resizebox{\linewidth}{!}{
    \begin{tabular}{r|p{26pt}<{\centering}p{26pt}<{\centering}p{26pt}<{\centering}p{26pt}<{\centering}|p{26pt}<{\centering}p{26pt}<{\centering}p{26pt}<{\centering}p{26pt}<{\centering}p{26pt}<{\centering}p{26pt}<{\centering}p{26pt}<{\centering}p{26pt}<{\centering}p{26pt}<{\centering}p{26pt}<{\centering}p{26pt}<{\centering}}
    \toprule
    \textbf{Method} & \textbf{Acc} & \textbf{mAcc} & \textbf{mIoU} & \textbf{fIoU} & \rotatebox{90}{\textcolor{background}{$\blacksquare$}~ground~} & \rotatebox{90}{\textcolor{building}{$\blacksquare$}~build} & \rotatebox{90}{\textcolor{fence}{$\blacksquare$}~fence} & \rotatebox{90}{\textcolor{person}{$\blacksquare$}~person} & \rotatebox{90}{\textcolor{pole}{$\blacksquare$}~pole} & \rotatebox{90}{\textcolor{road}{$\blacksquare$}~road} & \rotatebox{90}{\textcolor{sidewalk}{$\blacksquare$}~walk} & \rotatebox{90}{\textcolor{vegetation}{$\blacksquare$}~veg} & \rotatebox{90}{\textcolor{car}{$\blacksquare$}~car} & \rotatebox{90}{\textcolor{wall}{$\blacksquare$}~wall} & \rotatebox{90}{\textcolor{traffic-sign}{$\blacksquare$}~sign}
    \\\midrule\midrule
    \textcolor{gray}{Source-Only}~\textcolor{gray}{$\circ$} & \textcolor{gray}{$57.49$} & \textcolor{gray}{$33.39$} & \textcolor{gray}{$21.30$} & \textcolor{gray}{$12.00$} & \textcolor{gray}{$56.09$} & \textcolor{gray}{$30.76$} & \textcolor{gray}{$1.16$} & \textcolor{gray}{$13.67$} & \textcolor{gray}{$8.84$} & \textcolor{gray}{$37.18$} & \textcolor{gray}{$13.08$} & \textcolor{gray}{$56.39$} & \textcolor{gray}{$15.79$} & \textcolor{gray}{$1.08$} & \textcolor{gray}{$0.30$}
    \\\midrule
    AdaptSegNet \cite{tsai2018adaptsegnet} & $66.74$ & $41.78$ & $30.65$ & $13.54$ & $43.30$ & $57.25$ & $2.11$ & $23.74$ & $14.85$ & $66.78$ & $34.40$ & $59.86$ & $33.61$ & $1.12$ & \cellcolor{fly_red!12}$0.11$
    \\
    DACS \cite{tranheden2021dacs} & $71.20$ & $48.04$ & $34.78$ & $14.30$ & $45.05$ & $63.55$ & $3.44$ & $28.44$ & $23.52$ & $67.79$ & $39.25$ & $63.47$ & $43.46$ & \cellcolor{fly_green!12}$4.56$ & $0.00$
    \\
    MIC \cite{hoyer2023mic} & $72.46$ & $47.54$ & $35.22$ & \cellcolor{fly_red!12}$14.59$ & $47.87$ & $64.23$ & \cellcolor{fly_red!12}$4.17$ & \cellcolor{fly_red!12}$30.35$ & $21.61$ & \cellcolor{fly_red!12}$70.35$ & \cellcolor{fly_red!12}$40.20$ & $63.88$ & $42.85$ & $1.91$ & $0.00$
    \\
    PLSR \cite{zhao2024plsr} & \cellcolor{fly_red!12}$72.93$ & \cellcolor{fly_green!12}$49.82$ & \cellcolor{fly_red!12}$36.38$ & $14.48$ & \cellcolor{fly_red!12}$48.51$ & \cellcolor{fly_red!12}$64.69$ & $3.92$ & $30.15$ & \cellcolor{fly_red!12}$23.91$ & \cellcolor{fly_green!12}$71.16$ & \cellcolor{fly_green!12}$43.34$ & \cellcolor{fly_red!12}$65.40$ & \cellcolor{fly_red!12}$46.13$ & \cellcolor{fly_red!12}$2.97$ & $0.00$
    \\
    \textbf{\textcolor{fly_green}{Event}\textcolor{fly_red}{Fly}} \textbf{(Ours)} & \cellcolor{fly_green!12}$73.93$ & \cellcolor{fly_red!12}$49.56$ & \cellcolor{fly_green!12}$37.70$ & \cellcolor{fly_green!12}$14.93$ & \cellcolor{fly_green!12}$50.94$ & \cellcolor{fly_green!12}$66.17$ & \cellcolor{fly_green!12}$4.90$ & \cellcolor{fly_green!12}$35.48$ & \cellcolor{fly_green!12}$26.13$ & $66.73$ & $32.53$ & \cellcolor{fly_green!12}$69.77$ & \cellcolor{fly_green!12}$46.93$ & $2.49$ & \cellcolor{fly_green!12}$12.68$
    \\\midrule
    \cellcolor{gray!10}\textcolor{gray}{Target}~\textcolor{gray}{$\bullet$} & \cellcolor{gray!10}\textcolor{gray}{$86.12$} & \cellcolor{gray!10}\textcolor{gray}{$66.02$} & \cellcolor{gray!10}\textcolor{gray}{$55.93$} & \cellcolor{gray!10}\textcolor{gray}{$16.18$} & \cellcolor{gray!10}\textcolor{gray}{$87.07$} & \cellcolor{gray!10}\textcolor{gray}{$75.41$} & \cellcolor{gray!10}\textcolor{gray}{$22.70$} & \cellcolor{gray!10}\textcolor{gray}{$52.59$} & \cellcolor{gray!10}\textcolor{gray}{$39.41$} & \cellcolor{gray!10}\textcolor{gray}{$79.49$} & \cellcolor{gray!10}\textcolor{gray}{$58.82$} & \cellcolor{gray!10}\textcolor{gray}{$77.75$} & \cellcolor{gray!10}\textcolor{gray}{$69.63$} & \cellcolor{gray!10}\textcolor{gray}{$14.79$} & \cellcolor{gray!10}\textcolor{gray}{$37.61$}
    \\\bottomrule
    \end{tabular}}
\label{tab:quadruped2vehicle}
\end{table*}
\begin{table*}[t]
    \centering
    \caption{
        \textbf{Benchmark results of event camera cross-platform adaptation} from \includegraphics[width=0.019\linewidth]{figures/icons/quadruped.png}~quadruped ($\mathcal{P}^{\textcolor{fly_blue}{\mathbf{q}}}$) to \includegraphics[width=0.021\linewidth]{figures/icons/drone.png}~drone ($\mathcal{P}^{\textcolor{fly_red}{\mathbf{d}}}$). \textcolor{gray}{Target} denotes the model is trained with ground truth from the target domain. All scores are given in percentage (\%). The \textcolor{fly_red}{second best} and \textcolor{fly_green}{best} scores under each evaluation metric are highlighted in \colorbox{fly_red!12}{\textcolor{fly_red}{red}} and \colorbox{fly_green!12}{\textcolor{fly_green}{green}} colors, respectively.}
    \vspace{-0.25cm}
    \resizebox{\linewidth}{!}{
    \begin{tabular}{r|p{26pt}<{\centering}p{26pt}<{\centering}p{26pt}<{\centering}p{26pt}<{\centering}|p{26pt}<{\centering}p{26pt}<{\centering}p{26pt}<{\centering}p{26pt}<{\centering}p{26pt}<{\centering}p{26pt}<{\centering}p{26pt}<{\centering}p{26pt}<{\centering}p{26pt}<{\centering}p{26pt}<{\centering}p{26pt}<{\centering}}
    \toprule
    \textbf{Method} & \textbf{Acc} & \textbf{mAcc} & \textbf{mIoU} & \textbf{fIoU} & \rotatebox{90}{\textcolor{background}{$\blacksquare$}~ground~} & \rotatebox{90}{\textcolor{building}{$\blacksquare$}~build} & \rotatebox{90}{\textcolor{fence}{$\blacksquare$}~fence} & \rotatebox{90}{\textcolor{person}{$\blacksquare$}~person} & \rotatebox{90}{\textcolor{pole}{$\blacksquare$}~pole} & \rotatebox{90}{\textcolor{road}{$\blacksquare$}~road} & \rotatebox{90}{\textcolor{sidewalk}{$\blacksquare$}~walk} & \rotatebox{90}{\textcolor{vegetation}{$\blacksquare$}~veg} & \rotatebox{90}{\textcolor{car}{$\blacksquare$}~car} & \rotatebox{90}{\textcolor{wall}{$\blacksquare$}~wall} & \rotatebox{90}{\textcolor{traffic-sign}{$\blacksquare$}~sign}
    \\\midrule\midrule
    \textcolor{gray}{Source-Only}~\textcolor{gray}{$\circ$} & \textcolor{gray}{$52.62$} & \textcolor{gray}{$29.38$} & \textcolor{gray}{$16.85$} & \textcolor{gray}{$15.45$} & \textcolor{gray}{$50.85$} & \textcolor{gray}{$15.47$} & \textcolor{gray}{$1.65$} & \textcolor{gray}{$2.24$} & \textcolor{gray}{$15.48$} & \textcolor{gray}{$36.88$} & \textcolor{gray}{$9.98$} & \textcolor{gray}{$35.84$} & \textcolor{gray}{$15.20$} & \textcolor{gray}{$1.50$} & \textcolor{gray}{$0.23$}
    \\\midrule
    AdaptSegNet \cite{tsai2018adaptsegnet} & $57.07$ & $33.15$ & $20.96$ & $16.49$ & $31.15$ & $24.78$ & \cellcolor{fly_red!12}$2.71$ & $0.08$ & $19.90$ & $58.22$ & $4.43$ & $53.49$ & $20.42$ & $15.41$ & $0.00$
    \\
    DACS \cite{tranheden2021dacs} & $60.74$ & $38.60$ & $24.50$ & $17.92$ & $32.17$ & $26.42$ & \cellcolor{fly_green!12}$3.56$ & $2.01$ & $23.57$ & $60.32$ & $11.57$ & $56.01$ & $29.39$ & $24.50$ & $0.00$
    \\
    MIC \cite{hoyer2023mic} & \cellcolor{fly_red!12}$64.49$ & $40.02$ & $26.11$ & \cellcolor{fly_red!12}$18.65$ & $40.50$ & \cellcolor{fly_red!12}$29.26$ & $0.70$ & \cellcolor{fly_red!12}$3.02$ & $20.52$ & \cellcolor{fly_red!12}$62.66$ & \cellcolor{fly_green!12}$21.37$ & $57.58$ & \cellcolor{fly_green!12}$36.20$ & $15.36$ & $0.00$
    \\
    PLSR \cite{zhao2024plsr} & $63.57$ & \cellcolor{fly_green!12}$42.62$ & \cellcolor{fly_red!12}$27.34$ & $18.08$ & \cellcolor{fly_red!12}$40.71$ & $26.42$ & $0.42$ & \cellcolor{fly_green!12}$3.39$ & \cellcolor{fly_red!12}$24.07$ & $62.16$ & $18.07$ & \cellcolor{fly_red!12}$57.80$ & $29.16$ & \cellcolor{fly_green!12}$38.50$ & $0.00$
    \\
    \textbf{\textcolor{fly_green}{Event}\textcolor{fly_red}{Fly}} \textbf{(Ours)} & \cellcolor{fly_green!12}$65.78$ & \cellcolor{fly_red!12}$41.91$ & \cellcolor{fly_green!12}$28.79$ & \cellcolor{fly_green!12}$19.01$ & \cellcolor{fly_green!12}$40.74$ & \cellcolor{fly_green!12}$30.90$ & $1.50$ & $2.63$ & \cellcolor{fly_green!12}$24.76$ & \cellcolor{fly_green!12}$64.11$ & \cellcolor{fly_red!12}$18.22$ & \cellcolor{fly_green!12}$61.85$ & \cellcolor{fly_red!12}$33.44$ & \cellcolor{fly_red!12}$38.23$ & \cellcolor{fly_green!12}$0.29$
    \\\midrule
    \cellcolor{gray!10}\textcolor{gray}{Target}~\textcolor{gray}{$\bullet$} & \cellcolor{gray!10}\textcolor{gray}{$79.57$} & \cellcolor{gray!10}\textcolor{gray}{$52.25$} & \cellcolor{gray!10}\textcolor{gray}{$42.90$} & \cellcolor{gray!10}\textcolor{gray}{$23.30$} & \cellcolor{gray!10}\textcolor{gray}{$74.48$} & \cellcolor{gray!10}\textcolor{gray}{$39.40$} & \cellcolor{gray!10}\textcolor{gray}{$7.10$} & \cellcolor{gray!10}\textcolor{gray}{$0.33$} & \cellcolor{gray!10}\textcolor{gray}{$31.67$} & \cellcolor{gray!10}\textcolor{gray}{$71.96$} & \cellcolor{gray!10}\textcolor{gray}{$31.64$} & \cellcolor{gray!10}\textcolor{gray}{$67.87$} & \cellcolor{gray!10}\textcolor{gray}{$57.51$} & \cellcolor{gray!10}\textcolor{gray}{$66.14$} & \cellcolor{gray!10}\textcolor{gray}{$23.79$}
    \\\bottomrule
    \end{tabular}}
\label{tab:quadruped2drone}
\end{table*}

Our approach demonstrates superior performance in static classes, such as \texttt{road} and \texttt{vegetation}, which are critical for general scene understanding. This aligns with the strengths of EAP, which captures spatially consistent patterns. Dynamic classes often pose greater challenges due to motion and variability across domains. However, we observe that our approach achieves competitive results, surpassing existing methods in most cases. For example, in the \includegraphics[width=0.042\linewidth]{figures/icons/quadruped.png}~quadruped ($\mathcal{P}^{\textcolor{fly_blue}{\mathbf{q}}}$) to \includegraphics[width=0.04\linewidth]{figures/icons/vehicle.png}~vehicle ($\mathcal{P}^{\textcolor{fly_green}{\mathbf{v}}}$) scenario, our approach provides notable improvements in \texttt{car} and \texttt{person} classes, highlighting its ability to transfer motion-sensitive information effectively.

Additionally, the adaptation results emphasize the domain discrepancies between platforms. For instance, in the \includegraphics[width=0.045\linewidth]{figures/icons/drone.png}~drone ($\mathcal{P}^{\textcolor{fly_red}{\mathbf{d}}}$) to \includegraphics[width=0.04\linewidth]{figures/icons/vehicle.png}~vehicle ($\mathcal{P}^{\textcolor{fly_green}{\mathbf{v}}}$) setting, static classes such as \texttt{road} and \texttt{building} are better aligned, while smaller, dynamic classes like \texttt{pole} and \texttt{traffic-light} show more variation. This reflects the inherent viewpoint differences between high-altitude drone perspectives and ground-level vehicle data.

Similarly, in the \includegraphics[width=0.042\linewidth]{figures/icons/quadruped.png}~quadruped ($\mathcal{P}^{\textcolor{fly_blue}{\mathbf{q}}}$) to \includegraphics[width=0.045\linewidth]{figures/icons/drone.png}~drone ($\mathcal{P}^{\textcolor{fly_red}{\mathbf{d}}}$) scenario, our framework’s performance in \texttt{vegetation} and \texttt{terrain} highlights its ability to adapt between the low-altitude, close-proximity view of quadrupeds and the expansive aerial coverage of drones.

The additional results reinforce the effectiveness of the \textbf{\textit{\textcolor{fly_green}{Event}\textcolor{fly_red}{Fly}}} framework across diverse cross-platform settings. By addressing both static and dynamic class distributions and leveraging platform-specific activation patterns, our framework demonstrates superior generalization and robust adaptation capabilities. These insights further validate the suitability of our approach for real-world, multi-platform event camera perception applications.

\subsection{Additional Qualitative Assessment}
\begin{figure*}
    \centering
    \includegraphics[width=\linewidth]{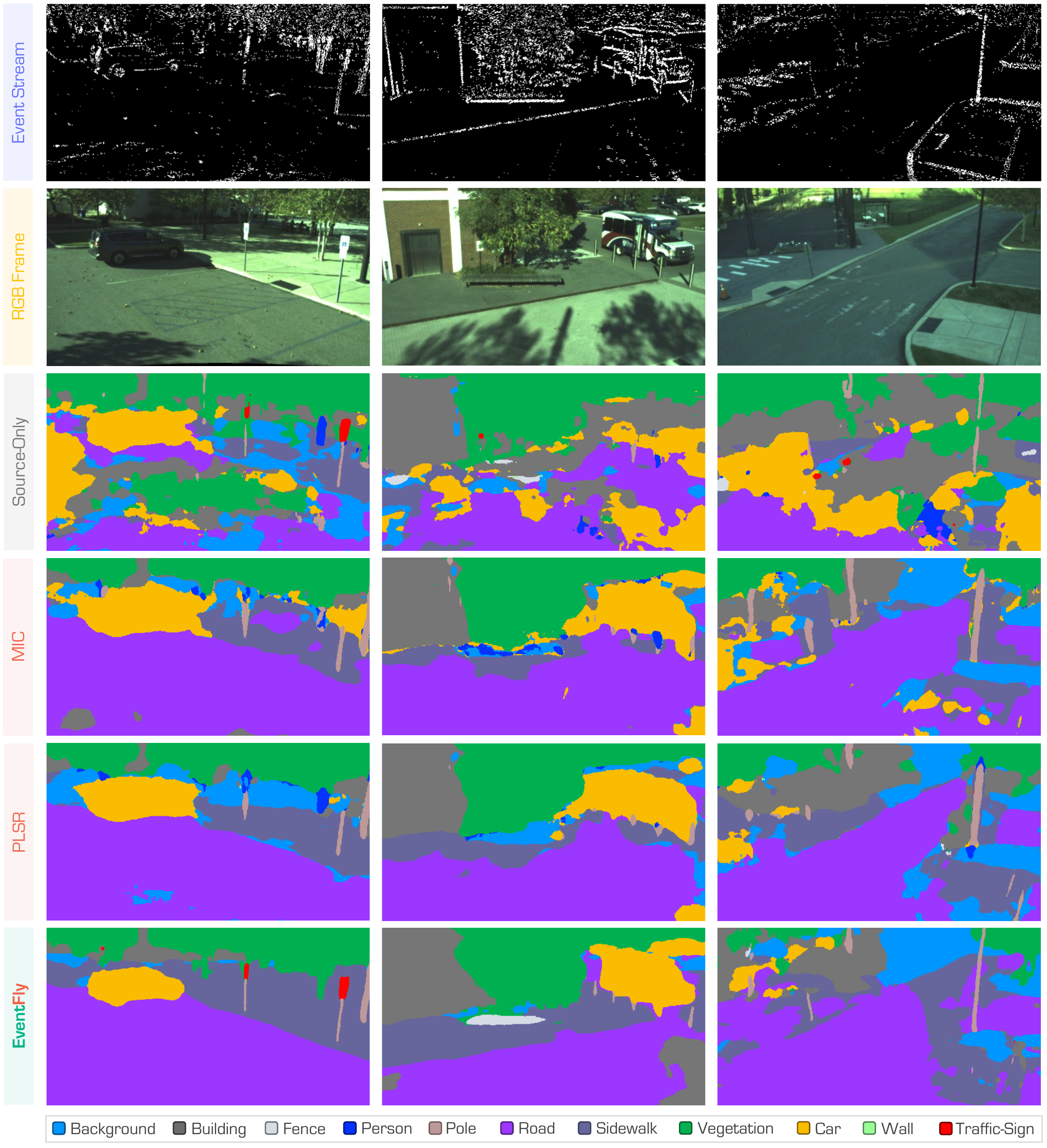}
    \vspace{-0.6cm}
    \caption{\textbf{Additional qualitative assessments} of cross-platform adaptation from the \includegraphics[width=0.018\linewidth]{figures/icons/vehicle.png}~vehicle ($\mathcal{P}^{\textcolor{fly_green}{\mathbf{v}}}$) platform to the \includegraphics[width=0.021\linewidth]{figures/icons/drone.png}~drone ($\mathcal{P}^{\textcolor{fly_red}{\mathbf{d}}}$) platform. We use grayscaled event images for better visibility. The RGB frames are for reference purposes only. Best viewed in colors.}
    \label{fig:qualitative_supp_1}
\end{figure*}

\begin{figure*}
    \centering
    \includegraphics[width=\linewidth]{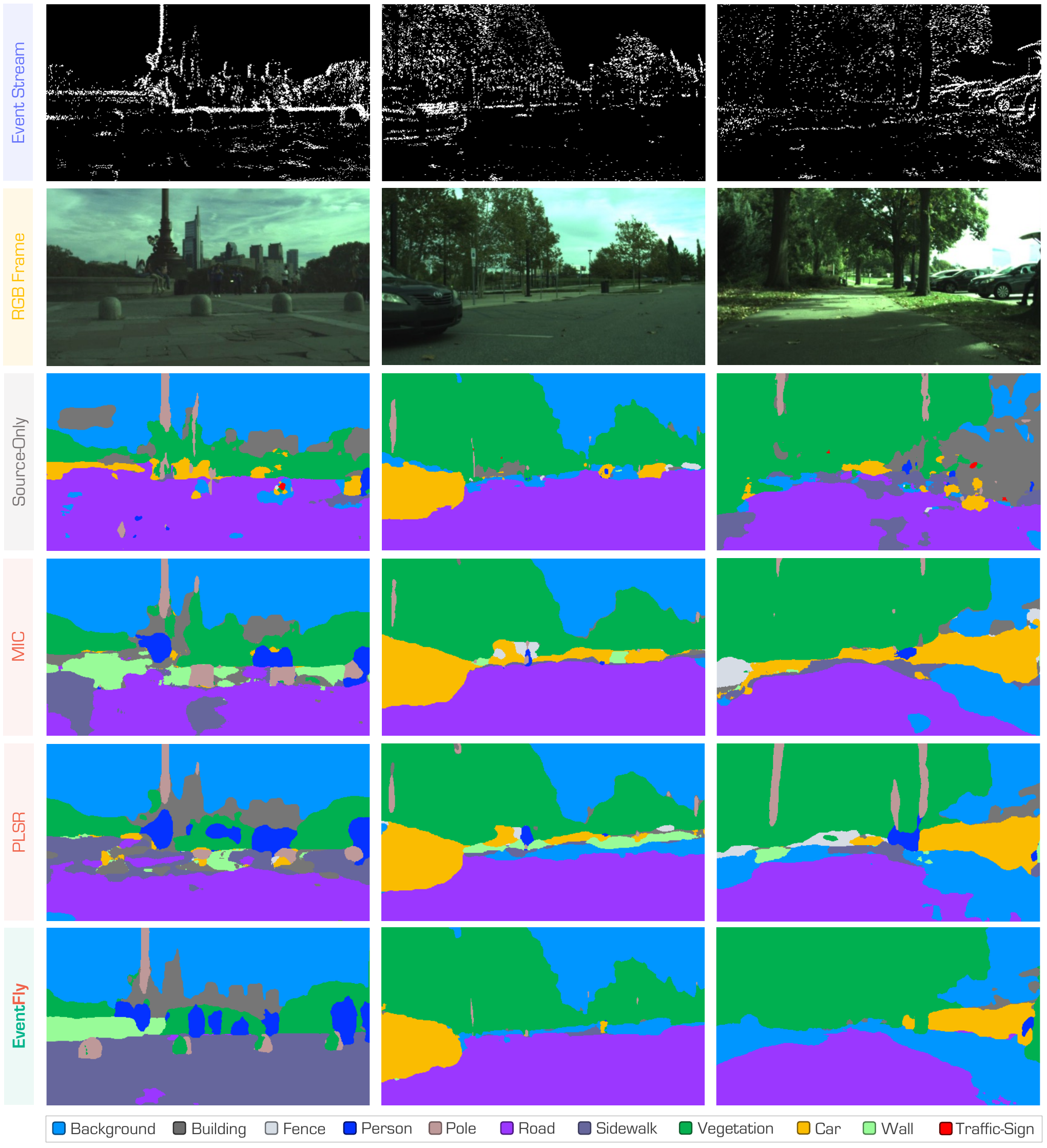}
    \vspace{-0.6cm}
    \caption{\textbf{Additional qualitative assessments} of cross-platform adaptation from the \includegraphics[width=0.018\linewidth]{figures/icons/vehicle.png}~vehicle ($\mathcal{P}^{\textcolor{fly_green}{\mathbf{v}}}$) platform to the \includegraphics[width=0.019\linewidth]{figures/icons/quadruped.png}~quadruped ($\mathcal{P}^{\textcolor{fly_blue}{\mathbf{q}}}$) platform. We use grayscaled event images for better visibility. The RGB frames are for reference purposes only. Best viewed in colors.}
    \label{fig:qualitative_supp_2}
\end{figure*}

\begin{figure*}
    \centering
    \includegraphics[width=\linewidth]{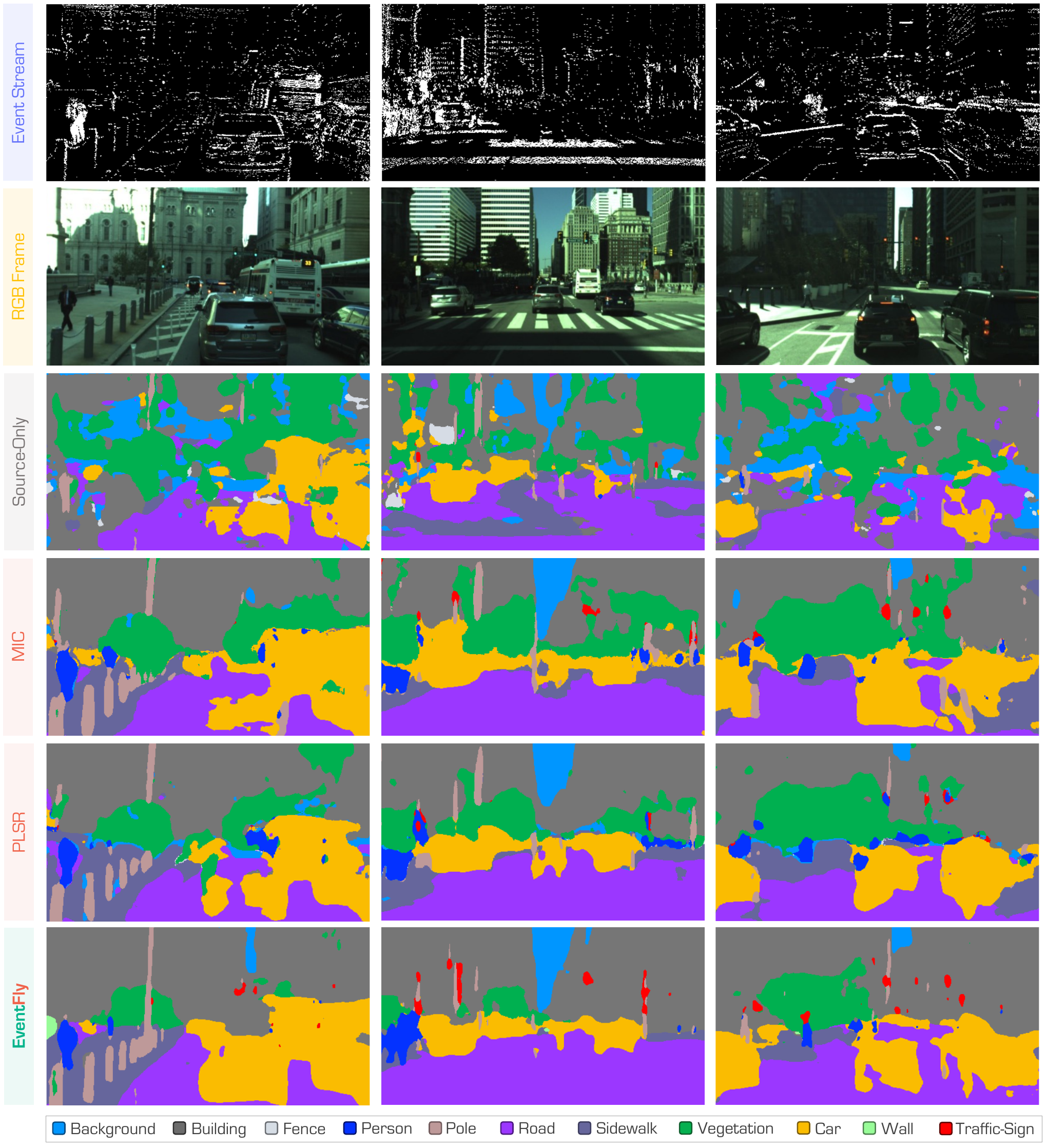}
    \vspace{-0.6cm}
    \caption{\textbf{Additional qualitative assessments} of cross-platform adaptation from the \includegraphics[width=0.021\linewidth]{figures/icons/drone.png}~drone ($\mathcal{P}^{\textcolor{fly_red}{\mathbf{d}}}$) platform to the \includegraphics[width=0.018\linewidth]{figures/icons/vehicle.png}~vehicle ($\mathcal{P}^{\textcolor{fly_green}{\mathbf{v}}}$) platform. We use grayscaled event images for better visibility. The RGB frames are for reference purposes only. Best viewed in colors.}
    \label{fig:qualitative_supp_3}
\end{figure*}

\begin{figure*}
    \centering
    \includegraphics[width=\linewidth]{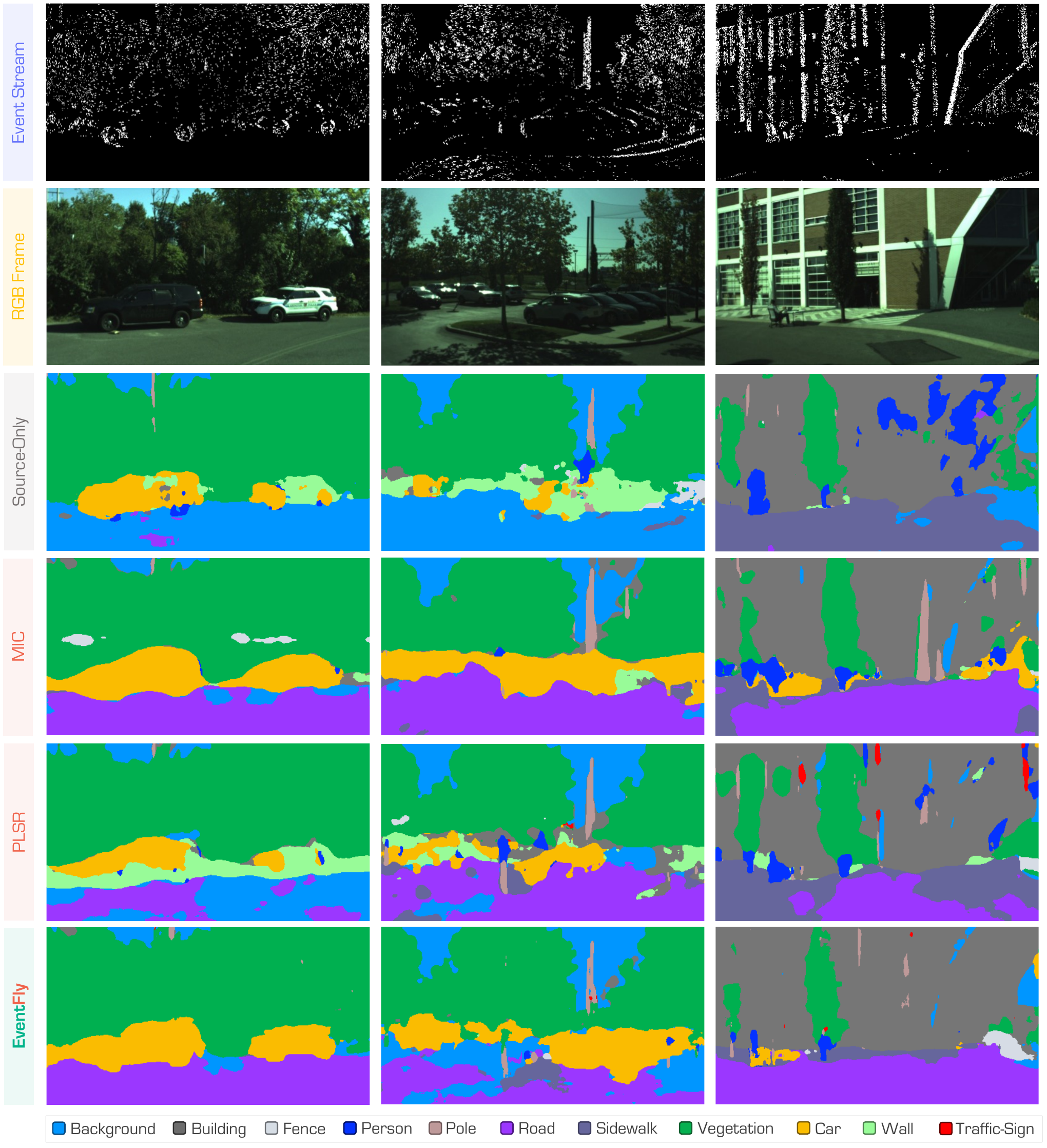}
    \vspace{-0.6cm}
    \caption{\textbf{Additional qualitative assessments} of cross-platform adaptation from the \includegraphics[width=0.019\linewidth]{figures/icons/quadruped.png}~quadruped ($\mathcal{P}^{\textcolor{fly_blue}{\mathbf{q}}}$) platform to the \includegraphics[width=0.018\linewidth]{figures/icons/vehicle.png}~vehicle ($\mathcal{P}^{\textcolor{fly_green}{\mathbf{v}}}$) platform. We use grayscaled event images for better visibility. The RGB frames are for reference purposes only. Best viewed in colors.}
    \label{fig:qualitative_supp_4}
\end{figure*}

In addition to the visual comparisons provided in the main body of this paper, we include more qualitative examples in this supplementary file. Please kindly refer to \cref{fig:qualitative_supp_1}, \cref{fig:qualitative_supp_2}, \cref{fig:qualitative_supp_3}, and \cref{fig:qualitative_supp_4} for the cross-platform adaptation results of the state-of-the-art adaptation methods.

\subsection{Failure Cases}
Although the proposed approach demonstrates promising cross-platform adaptation performance, there are certain failure cases that highlight the limitations and challenges of the approach.

Classes that are inherently dynamic and less frequently represented in the datasets, pose significant challenges.
Classes such as \texttt{traffic-sign}, which occupy small regions in the voxel grid, exhibit higher misclassification rates. This is particularly evident in the adaptation from \includegraphics[width=0.045\linewidth]{figures/icons/drone.png}~drone ($\mathcal{P}^{\textcolor{fly_red}{\mathbf{d}}}$) to \includegraphics[width=0.04\linewidth]{figures/icons/vehicle.png}~vehicle ($\mathcal{P}^{\textcolor{fly_green}{\mathbf{v}}}$), where high-altitude drone perspectives fail to capture the fine details necessary for distinguishing these classes in ground-level data. Additionally, in scenarios involving dense vegetation or crowded urban areas, occlusions lead to reduced prediction confidence.

\subsection{Video Demos}
To provide a more comprehensive illustration of the proposed \textbf{\textit{\textcolor{fly_green}{Event}\textcolor{fly_red}{Fly}}} framework and the \textbf{\textit{\textcolor{fly_green}{E}\textcolor{fly_red}{X}\textcolor{fly_blue}{Po}}} benchmark, we have attached three video demos with this supplementary material. Please kindly find the \texttt{demo1.mp4}, \texttt{demo2.mp4}, and \texttt{demo3.mp4} files on our project page\footnote{Project Page: \url{https://event-fly.github.io}.}.

Specifically, these three video demos contain the following visual content:
\begin{itemize}
    \item \textbf{Demo \#1:} The first demo consists of $813$ frames from the \texttt{penno\_parking\_2} sequence, illustrating the cross-platform adaptation from the \includegraphics[width=0.04\linewidth]{figures/icons/vehicle.png}~vehicle ($\mathcal{P}^{\textcolor{fly_green}{\mathbf{v}}}$) platform to the \includegraphics[width=0.045\linewidth]{figures/icons/drone.png}~drone ($\mathcal{P}^{\textcolor{fly_red}{\mathbf{d}}}$) platform.

    \item \textbf{Demo \#2:} The second demo consists of $1013$ frames from the \texttt{art\_plaza\_loop} sequence, illustrating the cross-platform adaptation from the \includegraphics[width=0.04\linewidth]{figures/icons/vehicle.png}~vehicle ($\mathcal{P}^{\textcolor{fly_green}{\mathbf{v}}}$) platform to the \includegraphics[width=0.042\linewidth]{figures/icons/quadruped.png}~quadruped ($\mathcal{P}^{\textcolor{fly_blue}{\mathbf{q}}}$) platform.

    \item \textbf{Demo \#3:} The third demo consists of $1,000$ frames from the \texttt{city\_hall} sequence, illustrating the cross-platform adaptation from the \includegraphics[width=0.045\linewidth]{figures/icons/drone.png}~drone ($\mathcal{P}^{\textcolor{fly_red}{\mathbf{d}}}$) platform to the \includegraphics[width=0.04\linewidth]{figures/icons/vehicle.png}~vehicle ($\mathcal{P}^{\textcolor{fly_green}{\mathbf{v}}}$) platform.
\end{itemize}

\section{Broader Impact \& Limitations}
In this section, we elaborate on the broader impact, societal influence, and potential limitations of the proposed \textbf{\textit{\textcolor{fly_green}{Event}\textcolor{fly_red}{Fly}}} framework and the \textbf{\textit{\textcolor{fly_green}{E}\textcolor{fly_red}{X}\textcolor{fly_blue}{Po}}} benchmark.

\subsection{Broader Impact}
Our approach and benchmark have the potential to redefine event camera perception across diverse operational platforms, including vehicles, drones, and quadrupeds. By enabling robust cross-platform adaptation, our framework could accelerate advancements in autonomous navigation, disaster response, and robotics, particularly in dynamic and unstructured environments. These contributions could enhance safety, efficiency, and adaptability in real-world applications, such as autonomous driving in dense urban areas, aerial surveillance in remote regions, and robotic assistance in disaster zones.

Moreover, the emphasis on domain-invariant learning for event-based perception addresses a critical gap in current technologies, facilitating the fairer deployment of AI systems across varied socioeconomic and geographical contexts. By creating a benchmark with diverse samples and settings, we aim to foster transparency and reproducibility in the evaluation of event-based systems, contributing to the broader research community’s understanding of event-camera capabilities and limitations.

\subsection{Societal Influence}
The societal influence of our approach and benchmark spans multiple domains:
\begin{itemize}
    \item \textbf{Improved Safety:} Enhanced perception capabilities in dynamic environments can improve safety in autonomous systems, reducing the risk of accidents in transportation and industrial applications.

    \item \textbf{Environmental Monitoring:} The adaptability of our framework to drones and quadrupeds facilitates ecological and environmental monitoring, promoting sustainability and conservation efforts.

    \item \textbf{Accessibility:} The cross-platform design lowers barriers for deploying event camera solutions in resource-constrained settings, democratizing access to advanced vision technologies.
\end{itemize}

Despite its benefits, it is essential to consider potential ethical implications, including misuse in surveillance and privacy-intrusive applications. Researchers and practitioners should adhere to ethical guidelines to mitigate risks associated with deploying these technologies.

\subsection{Potential Limitations}
While our approach and benchmark demonstrate substantial advancements, there are inherent limitations. For example, the reliance on domain-specific activation patterns might struggle in highly heterogeneous environments with atypical dynamics, such as extreme weather or chaotic lighting conditions. Besides, the reliance on pseudo-labels in unsupervised settings may propagate errors, especially when source-to-target domain gaps are substantial.

Additionally, although our benchmark is comprehensive, it might not encompass all possible scenarios, such as multi-agent coordination or environments with severe occlusions, necessitating further expansions. The current version of the benchmark also does not include settings of multi-source or multi-target adaptation.

In future work, we aim to address these challenges by optimizing the framework for real-time applications, expanding the benchmark to include more diverse scenarios, and investigating advanced self-supervised learning techniques to minimize reliance on pseudo-labels. By acknowledging these limitations, we hope to inspire continued innovation and improvement in event-based perception systems.
\section{Public Resource Used}
In this section, we acknowledge the use of the following public resources, during the course of this work:

\begin{itemize}
    \item M3ED\footnote{\url{https://m3ed.io}.}\dotfill CC BY-SA 4.0

    \item ESS\footnote{\url{https://github.com/uzh-rpg/ess}.}\dotfill GNU General Public License v3.0
    
    \item E2VID\footnote{\url{https://github.com/uzh-rpg/rpg_e2vid}.}\dotfill GNU General Public License v3.0

    \item AdaptSegNet\footnote{\url{https://github.com/wasidennis/AdaptSegNet}.}\dotfill Unknown

    \item CBST\footnote{\url{https://github.com/yzou2/CBST}.}\dotfill CC BY-SA 4.0

    \item IntraDA\footnote{\url{https://github.com/feipanir/IntraDA}.}\dotfill MIT License

    \item DACS\footnote{\url{https://github.com/vikolss/DACS}.}\dotfill MIT License

    \item MIC\footnote{\url{https://github.com/lhoyer/MIC}.}\dotfill Unknown

    \item Pytorch \footnote{\url{https://github.com/pytorch/pytorch}.} \dotfill Pytorch License 
    
    \item Pytorch3D \footnote{\url{https://github.com/facebookresearch/pytorch3d}.} \dotfill BSD-Style License 
    
    \item Open3D \footnote{\url{https://github.com/isl-org/Open3D}.} \dotfill MIT license
\end{itemize}

\clearpage

\clearpage

\clearpage

\clearpage

\clearpage\clearpage
{
    \small
    \bibliographystyle{ieeenat_fullname}
    \bibliography{main}
}

\end{document}